\documentclass[runningheads]{llncs}
\usepackage[a4paper,left=3cm,right=3cm]{geometry}

\usepackage[T1]{fontenc}
\usepackage{graphicx}
\usepackage{amsfonts}
\usepackage{amsmath}
\usepackage{algorithm}
\usepackage{algpseudocode}
\usepackage{censor,caption}
\usepackage{multirow}
 
\usepackage[T1]{fontenc}
\usepackage{graphicx,verbatim}
\usepackage{color}

\usepackage{booktabs}
\usepackage{array}
\usepackage{longtable}
\usepackage{tabularx}
\usepackage{threeparttable}
\usepackage{caption}
\usepackage{xurl}
\usepackage{amssymb} 

\usepackage[colorlinks=true,linkcolor=blue,citecolor=blue,urlcolor=blue]{hyperref}

\graphicspath{{figs/}}
\begin{document}
\title{Vision-language models for chest radiography do not always need the image}

\author{
Mahshad Lotfinia\inst{1} \and
Sebastian Ziegelmayer\inst{2} \and
Lisa Adams\inst{2} \and
Tri-Thien Nguyen\inst{1,5} \and
Daniel Truhn\inst{3,4} \and
Andreas Maier\inst{1} \and
Soroosh Tayebi Arasteh\inst{3,4}$^{\ast}$
}

\institute{
Pattern Recognition Lab, Friedrich-Alexander-Universit\"at Erlangen-N\"urnberg, Erlangen, Germany \and
Department of Diagnostic and Interventional Radiology, TUM University Clinic, School of Medicine and Health, Klinikum rechts der Isar, Technical University of Munich, Munich, Germany \and
Lab for AI in Medicine, RWTH Aachen University, Aachen, Germany \and
Department of Diagnostic and Interventional Radiology, University Hospital RWTH Aachen, Aachen, Germany \and
Institute of Radiology, University Hospital Erlangen, Erlangen, Germany
}

\maketitle 
{\footnotesize
\noindent$^{\ast}$Correspondence to: Soroosh Tayebi Arasteh (\email{soroosh.arasteh@rwth-aachen.de})
}

\begin{abstract}
Vision-language models that answer questions about chest radiographs are evaluated by their accuracy on labels derived from radiology reports. High benchmark accuracy is often interpreted as evidence that the model uses the image. A model that answers from the finding named in the question can score as well as a model that uses the radiograph. Keeping the question fixed, we audit eight open-weight systems by swapping in another patient's radiograph with the same or the opposite label, occluding the radiologist-marked region or an equal region elsewhere, and removing the radiograph or replacing it with noise or a photograph. On 2,548 yes-or-no questions from MIMIC-CXR, one multimodal model answers Yes regardless of the image, another multimodal model changes its answers without following the label, and four systems use the image but keep about half of their correct answers when the radiograph is swapped for an opposite-label radiograph. A medical model that receives only the question text scores 55.3\% on the pooled questions, higher than two multimodal systems. It scores 91.8\% where every finding is present, and answering Yes to every question scores 100\% there. Where the image is necessary, the best multimodal system exceeds this model by 10.4\% in balanced accuracy. The categories are unchanged on CheXpert. Confidence is not higher when a correct answer depends on the marked region. In a reader study with three radiologists, the two radiologists who read a balanced set of 200 cases score 86.0\% and 82.0\%, and the systems score 50.0\% to 73.0\%. Accuracy does not establish image use, but an intervention on the image can test it.
\end{abstract}

\section*{Introduction}

Vision-language models (VLMs), which pair a pretrained language model with an image encoder, are entering medical question-answering pipelines. Specialist biomedical variants and general-purpose systems report accuracies that approach expert level on chest radiography \cite{llavamed,googleresearch2026medgemma,achiam2023gpt4,team2024gemini,pal2025rexvqa}. Clinical applications that support, audit, or partially automate radiological workflows require the visual input to affect the answer \cite{thirunavukarasu2023llmclinical,moor2023generalist}. The evaluations behind these accuracies rarely test whether it does.

Accuracy on benchmarks derived from clinical labels and reports cannot distinguish a model that uses the image from a model that infers the answer from the name of the finding or from co-occurrence statistics in its training corpora. On benchmarks built from negated questions \cite{yan2025worse} or from model-confusable image pairs \cite{sepehri2025mediconfusion}, state-of-the-art systems score below chance. On clinical image challenges, the accuracy of multimodal models rises with the amount of informative text in the case \cite{buckley2026exploit}. More than a third of the correct answers of GPT-4V on such challenges came with a flawed rationale, most often in the comprehension of the image \cite{jin2024hidden}. The most capable multimodal models also answer questions about images that were not supplied \cite{asadi2026mirage}. Outside medicine, many questions of multimodal benchmarks can be answered without the image \cite{chen2024mmstar}. Unimodal medical classifiers can exploit shortcuts, recognizing scanner artifacts \cite{zech2018variable}, features correlated with race \cite{gichoya2022reading}, and acquisition signals correlated with disease prevalence \cite{degrave2021ai}.

Image reliance can be tested by removing the image and measuring the change in accuracy. On Italian medical questions that require the image, replacing the image with a blank placeholder lowered the accuracy of GPT-4o by more than 25\% and that of three other models by less than 10\% \cite{felizzi2025grounded}. A change in accuracy under removal depends on what a model answers when it receives no radiograph. It does not show whether an ordinary correct answer depended on the evidence for the queried finding. Saliency and attention maps \cite{selvaraju2017gradcam,adebayo2018sanity,Jain2019Attention} describe where a model attends. A more direct test is to change the image while the question stays fixed and to observe whether the answer changes \cite{pearl2009causality}. Phrase-grounded chest radiograph datasets such as MS-CXR \cite{boecking2022making} provide radiologist-marked regions for such interventions. To our knowledge, no evaluation of medical VLMs has combined label-controlled image swaps and occlusions of these regions with text-only and vision-only controls and with radiologists who read the same altered images.

\begin{figure}[p]
\centering
\includegraphics[width=\textwidth]{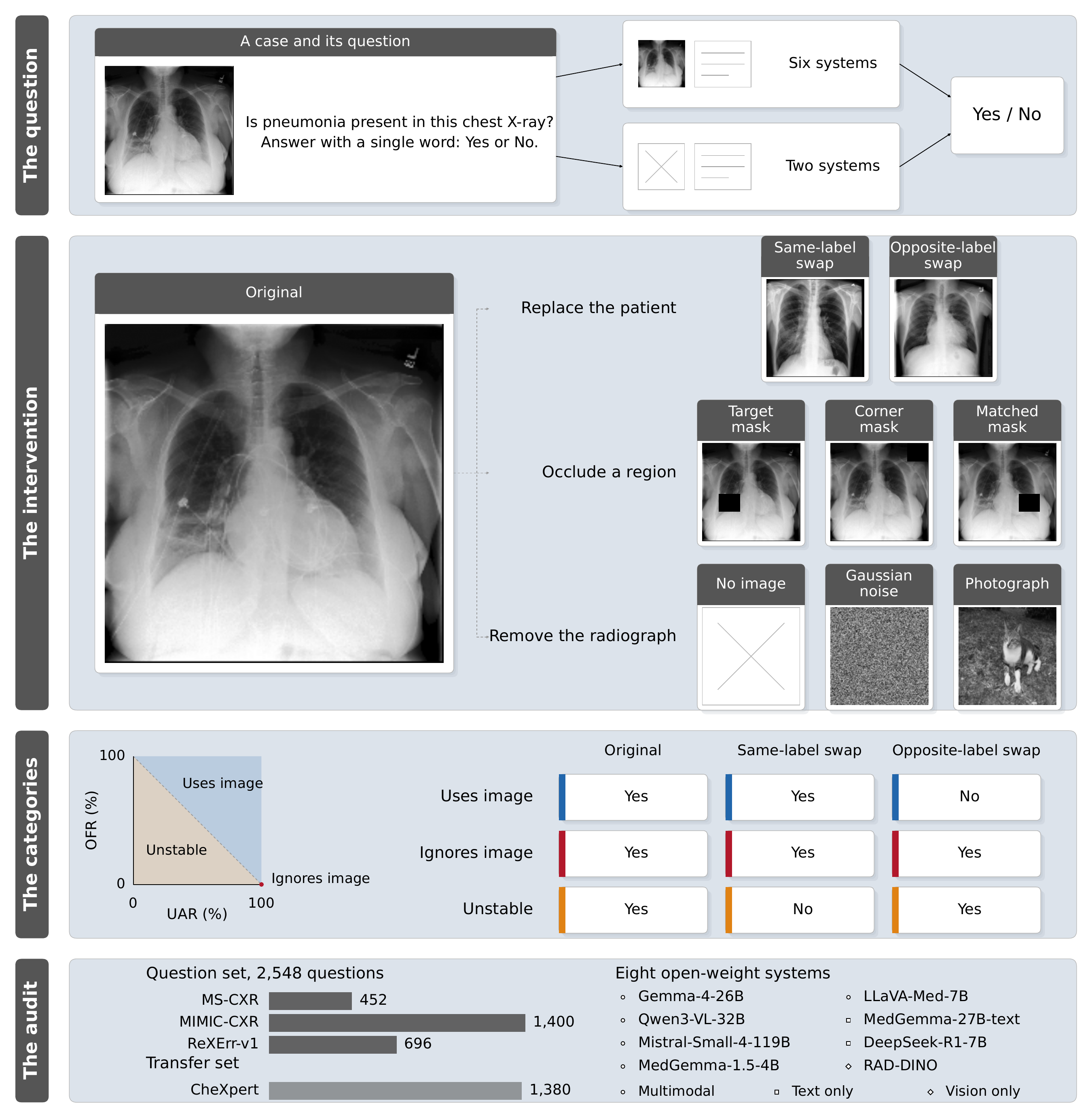}
\caption{Overview of the interventional audit of image use. \textbf{The question.} Each case is a chest radiograph paired with one yes-or-no question about a finding. Six systems receive the radiograph, five of them with the question, and two receive the question alone. \textbf{The intervention.} The question stays fixed while the image takes nine forms on one case: the original, two swaps that replace the patient, three occlusions that hide a region, and three replacements that remove or replace the radiograph. \textbf{The categories.} In the plane, the share of correct answers unchanged under the same-label swap (UAR) is plotted against the share changed under the opposite-label swap (OFR). Above the dashed line, answers follow the label of the swapped image. \textbf{The audit.} The audit uses 2,548 MIMIC questions (452 MS-CXR, 1,400 MIMIC-CXR, and 696 ReXErr) and 1,380 CheXpert questions for transfer. Because the credentialed MIMIC-CXR and MS-CXR images may not be reproduced, every radiograph here is a public example from NIH ChestX-ray14 \cite{wang2017chestxray8}, which is not an audited dataset and on which no result was computed.}
\label{fig:overview}
\end{figure}

We audit eight open-weight systems: three general-purpose multimodal models, two medical multimodal models, two language models that receive the question and no image and serve as text-only controls, and a vision-only reference, a logistic-regression head on frozen RAD-DINO image features \cite{perez2025raddino}. The question set contains 2{,}548 yes-or-no questions drawn from MS-CXR phrase-grounding boxes \cite{boecking2022making}, MIMIC-CXR labels \cite{johnson2019mimic}, and ReXErr-v1 report errors \cite{rajpurkar2024rexerr}. Each question is answered under the original radiograph and under interventions on the image alone (Fig.~\ref{fig:overview}). Two swaps replace the radiograph with another patient's radiograph with the same or the opposite label for the queried finding. Three occlusions black out the radiologist-marked region or a region of the same size at the image corner or at an anatomically matched position. Three replacements remove the radiograph or replace it with Gaussian noise or a photograph. We define the behavioral categories from the swaps, test localization with the occlusions, and measure with the replacements what each system answers without a radiograph. We replicate the categories on CheXpert \cite{irvin2019chexpert} and compare the systems with three board-certified radiologists who read the same displays. Supplementary Table~\ref{stab:glossary} defines every term of the audit.

LLaVA-Med-7B answers Yes regardless of the image, Mistral-Small-4-119B changes its answers under swaps without following the label, and four systems use the image but keep about half of their correct answers when the radiograph is swapped for an opposite-label radiograph. The text-only control MedGemma-27B-text outscores two multimodal systems on the pooled question set. On the MS-CXR questions, in which every finding is present, it scores higher than the multimodal systems that use the image. On the questions that need the image, every system that uses the image scores above the control. Image use varies across findings. These systems are not more confident when a correct answer depends on the marked region. In a reader study with three board-certified radiologists, the two radiologists who read a balanced set of 200 cases are more accurate than every system. Supplementary Table~\ref{stab:findings} summarizes the findings. Accuracy on such a benchmark therefore does not establish that a model uses the radiograph. An intervention on the image can test whether it does.

\section*{Results}

All rates are percentages, and the percent sign is omitted. Unless noted otherwise, every per-system rate is the mean over its cases, reported as that mean $\pm$ the standard deviation (SD) of its bootstrap distribution with the percentile 95\% confidence interval, written mean $\pm$ SD [lower, upper] \cite{efron1993introduction}. A paired between-system difference is instead the difference $\pm$ the SD of its paired bootstrap distribution with the 95\% interval. A per-regime confidence value is a mean of confidence scores $\pm$ their SD. Correlation and agreement coefficients (Spearman $\rho$, Cohen's $\kappa$) and $p$-values are given on their natural scale to three decimals. A comparison is called significant when its $p$-value is below 0.05 after the Benjamini-Hochberg false discovery rate (FDR) correction within its family \cite{benjamini1995controlling}.

\subsection*{The systems fall into three behavioral categories under two image swaps}

We audited eight open-weight systems on the MIMIC question set, 2{,}548 yes-or-no chest radiograph questions assembled from three sources in the MIMIC-CXR family (Supplementary Table~\ref{stab:composition}). It contains 452 MS-CXR cases with a radiologist-marked box for the queried finding, in all of which the finding is present. It also contains 1{,}400 MIMIC-CXR cases, 50 with the finding present and 50 with it absent for each of 13 findings, plus 100 normal studies asked whether any acute abnormality is present. The remaining 696 questions come from ReXErr-v1 and ask whether a report sentence describes the radiograph. In 576 of them, the sentence contains an injected error. The systems are three general-purpose multimodal models (Gemma-4-26B \cite{google2026gemma4}, Qwen3-VL-32B \cite{bai2023qwenvl,qwen2025qwen3vl}, and Mistral-Small-4-119B), two medical multimodal models (MedGemma-1.5-4B \cite{googleresearch2026medgemma} and LLaVA-Med-7B \cite{llavamed}), two text-only controls that receive the question and no image (MedGemma-27B-text \cite{googleresearch2026medgemma} and DeepSeek-R1-7B \cite{deepseekai2025deepseekr1}), and a vision-only reference, one logistic-regression head per finding on frozen RAD-DINO features \cite{perez2025raddino}, labeled RAD-DINO in the figures and tables (Supplementary Table~\ref{stab:registry}). Each question was answered under the original radiograph and under two swaps. The same-label swap replaces the radiograph by another patient's radiograph with the same label for the queried finding. The opposite-label swap replaces it by another patient's radiograph with the opposite label (Fig.~\ref{fig:overview}). On the correct-on-original answers, we summarize the swaps by two rates: the unrelated-image answer rate (UAR), the share of correct answers unchanged under the same-label swap, and the opposite-label flip rate (OFR), the share of correct answers changed under the opposite-label swap. Their difference, OFR minus (100 minus UAR), is the swap specificity premium (SSP), which is positive when answers change with the label of the image and not only with its identity. Each correct-on-original case has one replacement radiograph with each label. SSP therefore equals the sensitivity plus the specificity minus 1 of the answers to the replacement radiographs. It is therefore 0 for any system whose answer is the same for every image, whatever its answer prior.

\begin{figure}[p]
\centering
\includegraphics[width=\textwidth]{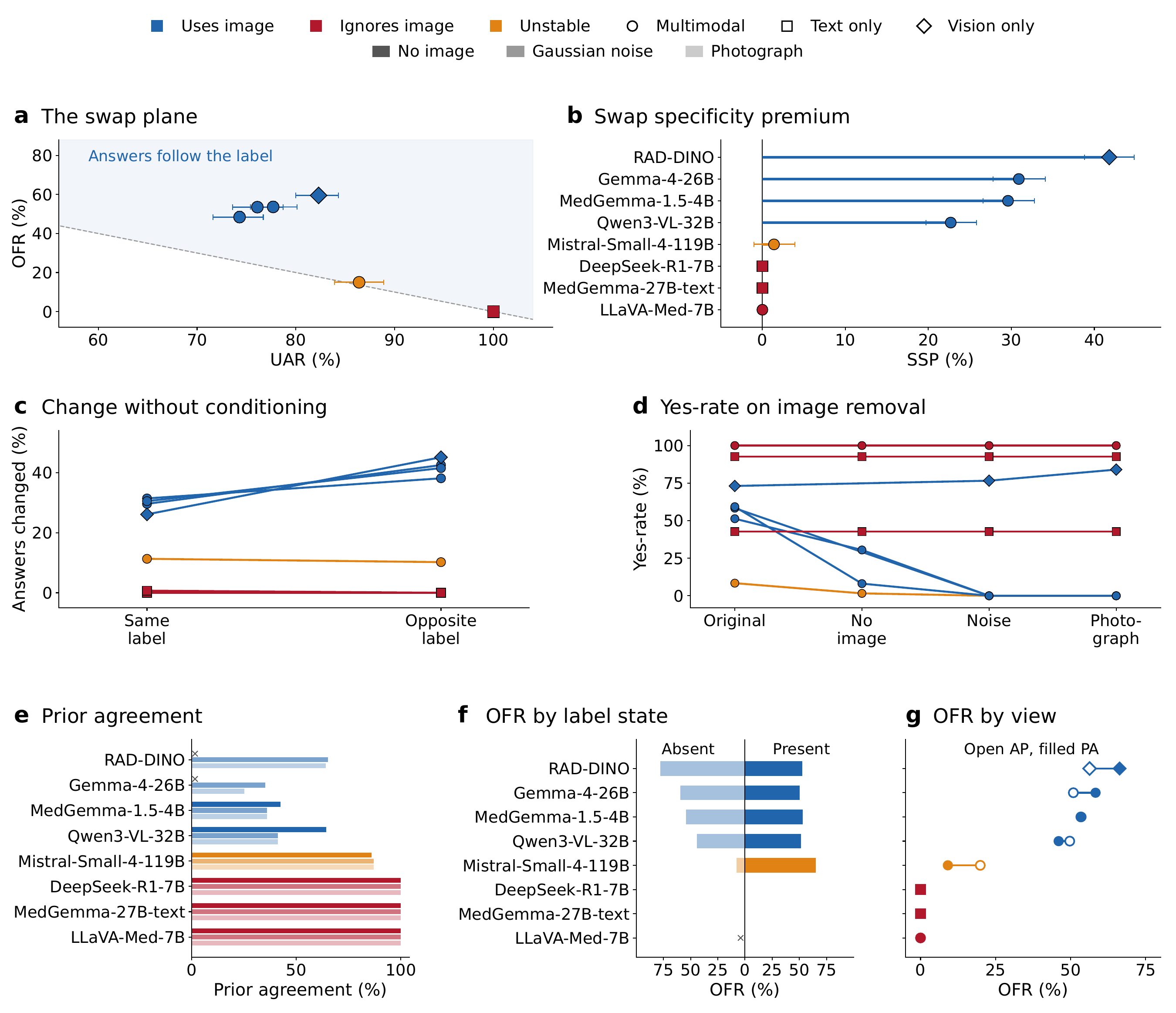}
\caption{The swaps and the image-removal conditions on the finding-presence cases of the MIMIC question set ($n$ = 1{,}852 questions from 1{,}499 patients). Fill color encodes the behavioral category (blue, uses image; red, ignores image; orange, unstable), and marker shape encodes the modality (circle, multimodal; square, text only; diamond, vision only). A marker is the value measured on its cases. Its error bar, drawn in \textbf{a} and \textbf{b} only, is the 95\% confidence interval, the 2.5th to the 97.5th percentile of 1{,}000 patient-cluster bootstrap resamples of that value. Case counts per panel: 731 to 1{,}337 (\textbf{a}, \textbf{b}), 1{,}708 to 1{,}852 (\textbf{c}), 253 to 1{,}852 (\textbf{d}), 191 to 1{,}337 (\textbf{e}), 50 to 1{,}070 (\textbf{f}), 260 to 907 (\textbf{g}). The systems are ordered by SSP in \textbf{b}, and \textbf{e} to \textbf{g} keep that order. The three fills of the second legend row apply to \textbf{e}. \textbf{a}, OFR against UAR, each on that system's correct-on-original answers. The dashed line marks OFR = 100 $-$ UAR, and SSP is positive above it. \textbf{b}, SSP per system. \textbf{c}, Share of all answered questions whose answer changes under each swap. \textbf{d}, Yes-rate under the original radiograph, no image, Gaussian noise, and a photograph. Gemma-4-26B and the vision-only reference have no value for no image. \textbf{e}, Prior agreement, the share of correct answers reproduced under each replacement, with a cross where it is not defined. \textbf{f}, OFR on the finding-absent (pale, left) and the finding-present (solid, right) questions, with a cross where it is not defined. \textbf{g}, OFR by view (AP, open; PA, filled). AP, anteroposterior; OFR, opposite-label flip rate; PA, posteroanterior; SSP, swap specificity premium; UAR, unrelated-image answer rate.}
\label{fig:categories}
\end{figure}

Three systems change no correct answer under either swap (Table~\ref{tab:master}; Fig.~\ref{fig:categories}a,b; Supplementary Fig.~\ref{sfig:swaps}a,b). Their OFR is 0.0 and their UAR is 100.0, on 731 to 1{,}070 informative cases per swap. Informative cases are the correct-on-original answers evaluated under that swap. They are the text-only controls, which receive no image, and LLaVA-Med-7B, which receives the image and answers Yes to every question (Yes-rate 100.0 on 2{,}513 answered questions). For the controls, the same-label swap repeats the original call, and the opposite-label swap reuses the original answer. Their OFR is therefore 0.0 by construction. These systems form the ignores-image category. Four systems flip about half of their correct answers when the label of the radiograph changes. Their OFR ranges from 48.4 $\pm$ 1.5 [45.6, 51.4] for Qwen3-VL-32B to 59.5 $\pm$ 1.4 [56.6, 62.4] for the vision-only reference. Under the same-label swap, they keep 74.3 $\pm$ 1.3 [71.6, 76.7] to 82.3 $\pm$ 1.1 [80.0, 84.3] of their correct answers. Their SSP ranges from 22.7 $\pm$ 1.6 [19.7, 25.8] to 41.8 $\pm$ 1.5 [38.8, 44.8], with every interval above zero. We call these systems the image users. Mistral-Small-4-119B changes 15.0 $\pm$ 1.3 [12.5, 17.5] of its correct answers under the opposite-label swap and 13.6 under the same-label swap (UAR 86.4 $\pm$ 1.3 [83.9, 88.9]). Its SSP of 1.4 $\pm$ 1.3 [$-1.0$, 3.9] has an interval that includes zero. Its answers therefore change about as often when the label changes as when it does not. It answers Yes to 8.4 of the finding-presence questions. We call this system unstable. We fixed the category rule before the opposite-label swap was run. It assigns every system to one category.

\begin{table}[p]
\centering
\caption{Main metrics of all systems and the two constant references on the MIMIC question set. Category is the assignment of the swap rule. Accuracy is computed on the pooled question set ($n$ = 2{,}548 cases) and balanced accuracy on the image-necessary subset ($n$ = 1{,}976 cases). UAR, OFR, and SSP are computed on the finding-presence cases of MS-CXR and MIMIC-CXR. Each value is the mean over its cases $\pm$ the standard deviation of 1{,}000 patient-cluster bootstrap resamples, with the percentile 95\% confidence interval and the case count $n$. The vision-only reference answers finding-presence questions only. Its pooled accuracy is therefore computed on 1{,}852 cases. Its balanced accuracy is computed on the 1{,}400 MIMIC-CXR cases of the subset. For the text-only controls, OFR is 0.0 by construction, because the opposite-label swap reuses their original answer. UAR, unrelated-image answer rate; OFR, opposite-label flip rate; SSP, swap specificity premium. N/A, not applicable.}
\label{tab:master}
\setlength{\tabcolsep}{3pt}
\renewcommand{\arraystretch}{1.05}
\scriptsize
\begin{tabular}{@{}p{0.14\textwidth}p{0.08\textwidth}p{0.13\textwidth}p{0.13\textwidth}p{0.13\textwidth}p{0.13\textwidth}p{0.13\textwidth}@{}}
\toprule
Model & Category & Accuracy (pooled) & Balanced accuracy (image-necessary) & UAR & OFR & SSP \\
\midrule
\multicolumn{7}{@{}l}{\textbf{General-purpose multimodal}} \\
Gemma-4-26B & Uses image & \begin{tabular}[t]{@{}l@{}}68.8 $\pm$ 1.0\\ {\scriptsize [66.8, 70.7]}\\ {\scriptsize $n=2{,}491$}\end{tabular} & \begin{tabular}[t]{@{}l@{}}63.4 $\pm$ 1.2\\ {\scriptsize [61.1, 65.7]}\\ {\scriptsize $n=1{,}947$}\end{tabular} & \begin{tabular}[t]{@{}l@{}}77.7 $\pm$ 1.2\\ {\scriptsize [75.4, 80.1]}\\ {\scriptsize $n=1{,}189$}\end{tabular} & \begin{tabular}[t]{@{}l@{}}53.6 $\pm$ 1.5\\ {\scriptsize [50.9, 56.5]}\\ {\scriptsize $n=1{,}200$}\end{tabular} & \begin{tabular}[t]{@{}l@{}}$+30.9$ $\pm$ 1.7\\ {\scriptsize [+27.8, +34.1]}\\ {\scriptsize $n=1{,}180$}\end{tabular} \\
Qwen3-VL-32B & Uses image & \begin{tabular}[t]{@{}l@{}}65.0 $\pm$ 1.0\\ {\scriptsize [63.0, 66.9]}\\ {\scriptsize $n=2{,}548$}\end{tabular} & \begin{tabular}[t]{@{}l@{}}61.0 $\pm$ 1.1\\ {\scriptsize [58.8, 63.0]}\\ {\scriptsize $n=1{,}976$}\end{tabular} & \begin{tabular}[t]{@{}l@{}}74.3 $\pm$ 1.3\\ {\scriptsize [71.6, 76.7]}\\ {\scriptsize $n=1{,}150$}\end{tabular} & \begin{tabular}[t]{@{}l@{}}48.4 $\pm$ 1.5\\ {\scriptsize [45.6, 51.4]}\\ {\scriptsize $n=1{,}150$}\end{tabular} & \begin{tabular}[t]{@{}l@{}}$+22.7$ $\pm$ 1.6\\ {\scriptsize [+19.7, +25.8]}\\ {\scriptsize $n=1{,}150$}\end{tabular} \\
Mistral-Small-4-119B & Unstable & \begin{tabular}[t]{@{}l@{}}46.6 $\pm$ 1.1\\ {\scriptsize [44.5, 48.7]}\\ {\scriptsize $n=2{,}548$}\end{tabular} & \begin{tabular}[t]{@{}l@{}}48.9 $\pm$ 1.0\\ {\scriptsize [47.0, 50.8]}\\ {\scriptsize $n=1{,}976$}\end{tabular} & \begin{tabular}[t]{@{}l@{}}86.4 $\pm$ 1.3\\ {\scriptsize [83.9, 88.9]}\\ {\scriptsize $n=800$}\end{tabular} & \begin{tabular}[t]{@{}l@{}}15.0 $\pm$ 1.3\\ {\scriptsize [12.5, 17.5]}\\ {\scriptsize $n=800$}\end{tabular} & \begin{tabular}[t]{@{}l@{}}$+1.4$ $\pm$ 1.3\\ {\scriptsize [-1.0, +3.9]}\\ {\scriptsize $n=800$}\end{tabular} \\
\midrule
\multicolumn{7}{@{}l}{\textbf{Medical multimodal}} \\
MedGemma-1.5-4B & Uses image & \begin{tabular}[t]{@{}l@{}}58.2 $\pm$ 1.1\\ {\scriptsize [55.9, 60.5]}\\ {\scriptsize $n=2{,}548$}\end{tabular} & \begin{tabular}[t]{@{}l@{}}57.4 $\pm$ 1.1\\ {\scriptsize [55.1, 59.6]}\\ {\scriptsize $n=1{,}976$}\end{tabular} & \begin{tabular}[t]{@{}l@{}}76.1 $\pm$ 1.3\\ {\scriptsize [73.6, 78.7]}\\ {\scriptsize $n=1{,}249$}\end{tabular} & \begin{tabular}[t]{@{}l@{}}53.5 $\pm$ 1.5\\ {\scriptsize [50.8, 56.3]}\\ {\scriptsize $n=1{,}249$}\end{tabular} & \begin{tabular}[t]{@{}l@{}}$+29.6$ $\pm$ 1.6\\ {\scriptsize [+26.6, +32.8]}\\ {\scriptsize $n=1{,}249$}\end{tabular} \\
LLaVA-Med-7B & Ignores image & \begin{tabular}[t]{@{}l@{}}47.9 $\pm$ 1.2\\ {\scriptsize [45.7, 50.5]}\\ {\scriptsize $n=2{,}513$}\end{tabular} & \begin{tabular}[t]{@{}l@{}}50.0 $\pm$ 0.0\\ {\scriptsize [50.0, 50.0]}\\ {\scriptsize $n=1{,}946$}\end{tabular} & \begin{tabular}[t]{@{}l@{}}100.0 $\pm$ 0.0\\ {\scriptsize [100.0, 100.0]}\\ {\scriptsize $n=1{,}068$}\end{tabular} & \begin{tabular}[t]{@{}l@{}}0.0 $\pm$ 0.0\\ {\scriptsize [0.0, 0.0]}\\ {\scriptsize $n=1{,}070$}\end{tabular} & \begin{tabular}[t]{@{}l@{}}$+0.0$ $\pm$ 0.0\\ {\scriptsize [+0.0, +0.0]}\\ {\scriptsize $n=1{,}055$}\end{tabular} \\
\midrule
\multicolumn{7}{@{}l}{\textbf{Text-only controls}} \\
MedGemma-27B-text & Ignores image & \begin{tabular}[t]{@{}l@{}}55.3 $\pm$ 1.1\\ {\scriptsize [53.1, 57.5]}\\ {\scriptsize $n=2{,}297$}\end{tabular} & \begin{tabular}[t]{@{}l@{}}53.0 $\pm$ 0.8\\ {\scriptsize [51.4, 54.6]}\\ {\scriptsize $n=1{,}744$}\end{tabular} & \begin{tabular}[t]{@{}l@{}}100.0 $\pm$ 0.0\\ {\scriptsize [100.0, 100.0]}\\ {\scriptsize $n=1{,}065$}\end{tabular} & \begin{tabular}[t]{@{}l@{}}0.0 $\pm$ 0.0\\ {\scriptsize [0.0, 0.0]}\\ {\scriptsize $n=1{,}065$}\end{tabular} & \begin{tabular}[t]{@{}l@{}}$+0.0$ $\pm$ 0.0\\ {\scriptsize [+0.0, +0.0]}\\ {\scriptsize $n=1{,}065$}\end{tabular} \\
DeepSeek-R1-7B & Ignores image & \begin{tabular}[t]{@{}l@{}}40.7 $\pm$ 1.0\\ {\scriptsize [38.5, 42.7]}\\ {\scriptsize $n=2{,}392$}\end{tabular} & \begin{tabular}[t]{@{}l@{}}43.3 $\pm$ 1.1\\ {\scriptsize [41.0, 45.5]}\\ {\scriptsize $n=1{,}862$}\end{tabular} & \begin{tabular}[t]{@{}l@{}}100.0 $\pm$ 0.0\\ {\scriptsize [100.0, 100.0]}\\ {\scriptsize $n=731$}\end{tabular} & \begin{tabular}[t]{@{}l@{}}0.0 $\pm$ 0.0\\ {\scriptsize [0.0, 0.0]}\\ {\scriptsize $n=731$}\end{tabular} & \begin{tabular}[t]{@{}l@{}}$+0.0$ $\pm$ 0.0\\ {\scriptsize [+0.0, +0.0]}\\ {\scriptsize $n=731$}\end{tabular} \\
\midrule
\multicolumn{7}{@{}l}{\textbf{Vision-only reference}} \\
RAD-DINO & Uses image & \begin{tabular}[t]{@{}l@{}}72.2 $\pm$ 1.1\\ {\scriptsize [70.0, 74.2]}\\ {\scriptsize $n=1{,}852$}\end{tabular} & \begin{tabular}[t]{@{}l@{}}66.0 $\pm$ 1.2\\ {\scriptsize [63.9, 68.4]}\\ {\scriptsize $n=1{,}400$}\end{tabular} & \begin{tabular}[t]{@{}l@{}}82.3 $\pm$ 1.1\\ {\scriptsize [80.0, 84.3]}\\ {\scriptsize $n=1{,}337$}\end{tabular} & \begin{tabular}[t]{@{}l@{}}59.5 $\pm$ 1.4\\ {\scriptsize [56.6, 62.4]}\\ {\scriptsize $n=1{,}337$}\end{tabular} & \begin{tabular}[t]{@{}l@{}}$+41.8$ $\pm$ 1.5\\ {\scriptsize [+38.8, +44.8]}\\ {\scriptsize $n=1{,}337$}\end{tabular} \\
\midrule
\multicolumn{7}{@{}l}{\textbf{Constant references}} \\
Always Yes & -- & \begin{tabular}[t]{@{}l@{}}48.0 $\pm$ 1.2\\ {\scriptsize [45.7, 50.4]}\\ {\scriptsize $n=2{,}548$}\end{tabular} & \begin{tabular}[t]{@{}l@{}}50.0 $\pm$ 0.0\\ {\scriptsize [50.0, 50.0]}\\ {\scriptsize $n=1{,}976$}\end{tabular} & N/A & N/A & N/A \\
Always No & -- & \begin{tabular}[t]{@{}l@{}}52.0 $\pm$ 1.2\\ {\scriptsize [49.6, 54.3]}\\ {\scriptsize $n=2{,}548$}\end{tabular} & \begin{tabular}[t]{@{}l@{}}50.0 $\pm$ 0.0\\ {\scriptsize [50.0, 50.0]}\\ {\scriptsize $n=1{,}976$}\end{tabular} & N/A & N/A & N/A \\
\bottomrule
\end{tabular}
\end{table}

The same three groups appear without the conditioning on correct answers (Fig.~\ref{fig:categories}c; Supplementary Fig.~\ref{sfig:swaps}c--e). Over all answered finding-presence questions, the image users change 26.1 to 31.4 of their answers under the same-label swap and 38.1 to 45.1 under the opposite-label swap, Mistral-Small-4-119B changes 11.3 and 10.2, and the ignores-image systems change 0.0 to 0.7 and 0.0. DeepSeek-R1-7B changes 0.7 of its answers under the same-label swap, which for a text-only control repeats the call with the same input. None of these answers was correct under the original call. Scored against the label of the swapped radiograph, the answers given under the opposite-label swap are correct on 59.9 to 68.7 of the questions for the image users, on 59.6 for Mistral-Small-4-119B, and on 40.7 to 57.2 for the ignores-image systems. If a correct answer that becomes a non-answer under a swap counts as changed, LLaVA-Med-7B changes 1.5 and 1.3 of its correct answers under the same-label and the opposite-label swap. No other UAR or OFR changes by more than 1.4. The replacement radiographs are matched on the label of the queried finding and not on the view. On the 680 finding-presence questions whose two replacements have the same view as the original radiograph, SSP is 16.1 to 33.5 for the image users and $-1.1 \pm 2.4$ [$-5.6$, 3.7] for Mistral-Small-4-119B. Every category is unchanged on these questions and on the 498 CheXpert questions selected the same way.

To test what each system answers from the question alone, we removed the radiograph or replaced it with Gaussian noise or a photograph (Fig.~\ref{fig:categories}d,e; Supplementary Fig.~\ref{sfig:swaps}f). Shown one fixed noise image or one fixed photograph of a cat in place of every radiograph, the multimodal image users answer No whenever they answer a finding-presence question (Yes-rate 0.0). They therefore keep only their correct No answers, which are 25.1 to 41.3 of their correct answers. Gemma-4-26B declines to answer 840 of the noise displays and 1{,}599 of the photograph displays. To 1{,}562 of the photograph displays, it responds ``I cannot answer this question because there is no chest''. The vision-only reference answers Yes to 76.6 of the noise displays and 84.0 of the photograph displays. It keeps 65.1 and 64.2 of its correct answers. Mistral-Small-4-119B keeps 87.1 of its correct answers under both conditions, and LLaVA-Med-7B keeps all of them. With no image, Gemma-4-26B declines to answer every finding-presence question. Qwen3-VL-32B keeps 64.3 $\pm$ 1.6 [61.2, 67.5] of its correct answers with a Yes-rate of 30.5, MedGemma-1.5-4B keeps 42.4 $\pm$ 1.5 [39.5, 45.4] with a Yes-rate of 8.1, and Mistral-Small-4-119B keeps 86.0 $\pm$ 1.3 [83.6, 88.5] with a Yes-rate of 1.6.

\subsection*{Pooled accuracy includes questions that do not require the image}

\begin{figure}[p]
\centering
\includegraphics[width=\textwidth]{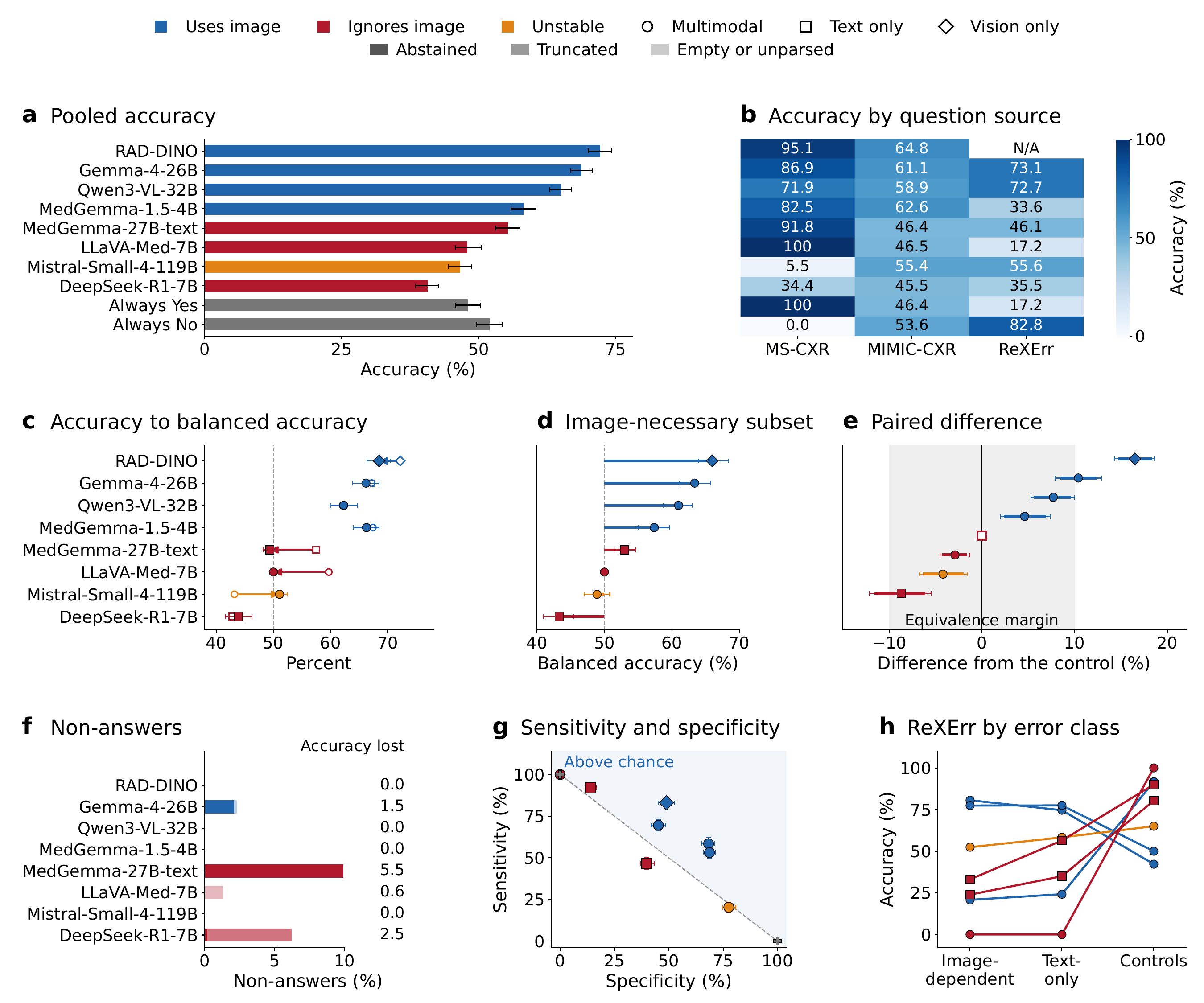}
\caption{The accuracy family of all systems on the MIMIC question set ($n$ = 2{,}548 questions from 1{,}608 patients). Fill color encodes the behavioral category (blue, uses image; red, ignores image; orange, unstable), gray marks the constant answers, and marker shape encodes the modality (circle, multimodal; square, text only; diamond, vision only). A marker is the accuracy measured on its cases, and in \textbf{e} it is the paired difference. An error bar, drawn in \textbf{a}, \textbf{c}, \textbf{d}, \textbf{e}, and \textbf{g} only, spans the 2.5th to the 97.5th percentile of 1{,}000 patient-cluster bootstrap resamples of that value, which is the 95\% confidence interval. Case counts per panel: 1{,}852 to 2{,}548 (\textbf{a}, \textbf{f}), 410 to 1{,}400 (\textbf{b}), 1{,}708 to 1{,}852 (\textbf{c}), 1{,}400 to 1{,}976 (\textbf{d}), 1{,}400 to 1{,}744 (\textbf{e}), 650 to 1{,}206 (\textbf{g}), 60 to 456 (\textbf{h}). N/A marks a cell without a value for the vision-only reference. \textbf{a}, Accuracy on the pooled question set. \textbf{b}, Accuracy on the MS-CXR, MIMIC-CXR, and ReXErr blocks (452, 1{,}400, and 696 questions). \textbf{c}, An arrow per system from accuracy (open) to balanced accuracy (filled), where the dashed line marks chance. \textbf{d}, Balanced accuracy on the image-necessary subset of 1{,}976 questions, where each stem starts at chance. \textbf{e}, Paired difference in balanced accuracy against the text-only control MedGemma-27B-text. The thin bar is the 95\% CI, and the thick bar is the 90\% interval of the two one-sided tests. The shaded band marks the equivalence margin of $\pm$10\%. \textbf{f}, Share of the questions left unanswered by each system, by class, and the pooled accuracy lost when a non-answer counts as incorrect. \textbf{g}, Sensitivity against specificity, where the dashed line marks a balanced accuracy of 50. \textbf{h}, Accuracy on the three ReXErr classes (456, 120, and 120 questions). CI, confidence interval.}
\label{fig:accuracy}
\end{figure}

On the pooled question set, the text-only control MedGemma-27B-text scores 55.3 $\pm$ 1.1 [53.1, 57.5] (Table~\ref{tab:master}; Fig.~\ref{fig:accuracy}a; Supplementary Fig.~\ref{sfig:accuracy}). Three multimodal systems score above it: Gemma-4-26B 68.8, Qwen3-VL-32B 65.0, and MedGemma-1.5-4B 58.2. Three systems score below it: LLaVA-Med-7B 47.9, Mistral-Small-4-119B 46.6, and DeepSeek-R1-7B 40.7. The vision-only reference answers only the 1{,}852 finding-presence questions. On these, it scores 72.2 $\pm$ 1.1 [70.0, 74.2], and the control scores 57.5. Answering Yes to every question scores 48.0, and answering No to every question scores 52.0. On shared questions, the control outscores LLaVA-Med-7B by 4.6 $\pm$ 0.9 [2.7, 6.3] and Mistral-Small-4-119B by 10.1 $\pm$ 1.9 [6.2, 13.7] (each $p_{\mathrm{FDR}} = 0.001$). It is 14.1 $\pm$ 1.3 [11.6, 16.6] below Gemma-4-26B and 14.7 $\pm$ 1.1 [12.7, 16.8] below the vision-only reference (Supplementary Fig.~\ref{sfig:vs_medgemma}a). The lower-scoring control DeepSeek-R1-7B is less accurate than every image user on the pooled question set, the finding-presence questions, the image-necessary subset, and the MIMIC-CXR and MS-CXR blocks (each $p_{\mathrm{FDR}} = 0.001$; Supplementary Fig.~\ref{sfig:vs_deepseek}a--e). On the MS-CXR block, in which every finding is present, the control scores 91.8 $\pm$ 1.6 [88.3, 94.8] and exceeds Gemma-4-26B by 4.5 $\pm$ 1.2 [2.3, 6.9] (Fig.~\ref{fig:accuracy}b). The control answers Yes to 92.6 of all finding-presence questions. On the MIMIC-CXR block, whose correct answer is Yes for 650 and No for 750 of the 1{,}400 questions, it scores 46.4 $\pm$ 1.3 [43.9, 48.9]. The constant No answer scores 53.6 there.

Balanced accuracy is the mean of sensitivity and specificity. A constant answer scores 50 on it (Fig.~\ref{fig:accuracy}c; Supplementary Fig.~\ref{sfig:accuracy}a--e). On the 1{,}852 finding-presence questions, the control's balanced accuracy is 49.4 $\pm$ 0.6 [48.2, 50.6]. Because the control receives the same prompt for the finding-present and the finding-absent cases of a finding, this value is at chance by construction. LLaVA-Med-7B has a balanced accuracy of 50.0 and an accuracy of 59.7. The image users score 62.3 $\pm$ 1.2 [60.0, 64.7] (Qwen3-VL-32B) to 68.5 $\pm$ 1.0 [66.4, 70.5] (the vision-only reference).

We therefore compare the systems on the 1{,}976 questions whose correct answer depends on the radiograph (Fig.~\ref{fig:accuracy}d,g; Supplementary Fig.~\ref{sfig:accuracy}f). They are the MIMIC-CXR block and the ReXErr image-dependent errors and error-free controls. Of these questions, 39.0 have a correct answer of Yes. Balanced accuracy on this subset is 66.0 $\pm$ 1.2 [63.9, 68.4] for the vision-only reference on its 1{,}400 MIMIC-CXR questions, 63.4 $\pm$ 1.2 [61.1, 65.7] for Gemma-4-26B, 61.0 for Qwen3-VL-32B, and 57.4 for MedGemma-1.5-4B. It is 53.0 $\pm$ 0.8 [51.4, 54.6] for the control, 50.0 for LLaVA-Med-7B, 48.9 for Mistral-Small-4-119B, and 43.3 for DeepSeek-R1-7B. Every image user exceeds the control on shared questions, by 16.5 $\pm$ 1.1 [14.3, 18.6] for the vision-only reference, 10.4 $\pm$ 1.3 [7.9, 12.9] for Gemma-4-26B, 7.7 $\pm$ 1.2 [5.3, 10.0] for Qwen3-VL-32B, and 4.6 $\pm$ 1.4 [2.0, 7.4] for MedGemma-1.5-4B (all $p_{\mathrm{FDR}} \leq 0.002$; Fig.~\ref{fig:accuracy}e). Tested against an equivalence margin of 10 with two one-sided tests \cite{schuirmann1987tost}, the 90\% interval of the advantage over the control lies within the margin for Qwen3-VL-32B ([5.6, 9.6]) and for MedGemma-1.5-4B ([2.3, 6.9]). These two systems are therefore equivalent to the control within 10 and also significantly more accurate than it. The interval of Gemma-4-26B includes 10 ([8.5, 12.4]), and the interval of the vision-only reference lies above the margin ([14.7, 18.4]; Fig.~\ref{fig:accuracy}e). On the questions shared with each system other than the control, the vision-only reference has a balanced accuracy higher by 3.2 to 21.3 (Supplementary Fig.~\ref{sfig:vs_medgemma}g). Every pairwise difference among these systems is significant except the difference between LLaVA-Med-7B and Mistral-Small-4-119B ($p_{\mathrm{FDR}} = 0.258$). The control exceeds chance on this subset only through the sentence questions, on which it answers No to 33.1 of the answered image-dependent error sentences (Fig.~\ref{fig:accuracy}h). On the MIMIC-CXR block, its balanced accuracy is 49.5 $\pm$ 0.7 [48.1, 50.8] (Supplementary Fig.~\ref{sfig:accuracy}e). Every image user exceeds it by 9.1 to 16.5 on shared questions (each $p_{\mathrm{FDR}} = 0.001$; Supplementary Fig.~\ref{sfig:vs_medgemma}d). On the ReXErr block, Gemma-4-26B and Qwen3-VL-32B exceed the control's accuracy by 28.8 and 29.0 (Supplementary Fig.~\ref{sfig:vs_medgemma}f). Their balanced accuracy on this block is not significantly different from the control's (differences of 0.9 and 0.6, $p_{\mathrm{FDR}} = 0.883$).

Non-answers are excluded from every rate above and reported separately (Supplementary Fig.~\ref{sfig:nonanswers}a,b). The control declines 251 of the 696 sentence questions with a request for the image. Gemma-4-26B declines 53 questions and returns four unparsed outputs, and LLaVA-Med-7B returns 34 empty outputs and one unparsed output. DeepSeek-R1-7B abstains in four outputs and gives no final answer in 152 further outputs, 150 of which reached its budget of 2{,}048 tokens. Scoring every non-answer as incorrect lowers the control's accuracy from 55.3 to 49.8 on the pooled question set, from 45.8 to 40.4 on the image-necessary subset, and from 46.1 to 29.5 on the ReXErr block. It lowers the pooled accuracy of Gemma-4-26B from 68.8 to 67.3 and of DeepSeek-R1-7B from 40.7 to 38.2 (Fig.~\ref{fig:accuracy}f; Supplementary Fig.~\ref{sfig:nonanswers}c--e). Among the accuracy, UAR, OFR, and CGR of all systems, a permissive parser that accepts the first Yes or No anywhere in an output changes only the accuracy of DeepSeek-R1-7B. Its answered share increases from 93.9 to 100.0, and its accuracy is then 40.6. Rerunning its 152 truncated outputs with a budget of 8{,}192 tokens increases the answered share to 99.8 and the accuracy to 41.5 (Supplementary Fig.~\ref{sfig:nonanswers}f).

\subsection*{Localization is partial and uneven across findings and views}

\begin{figure}[p]
\centering
\includegraphics[width=\textwidth]{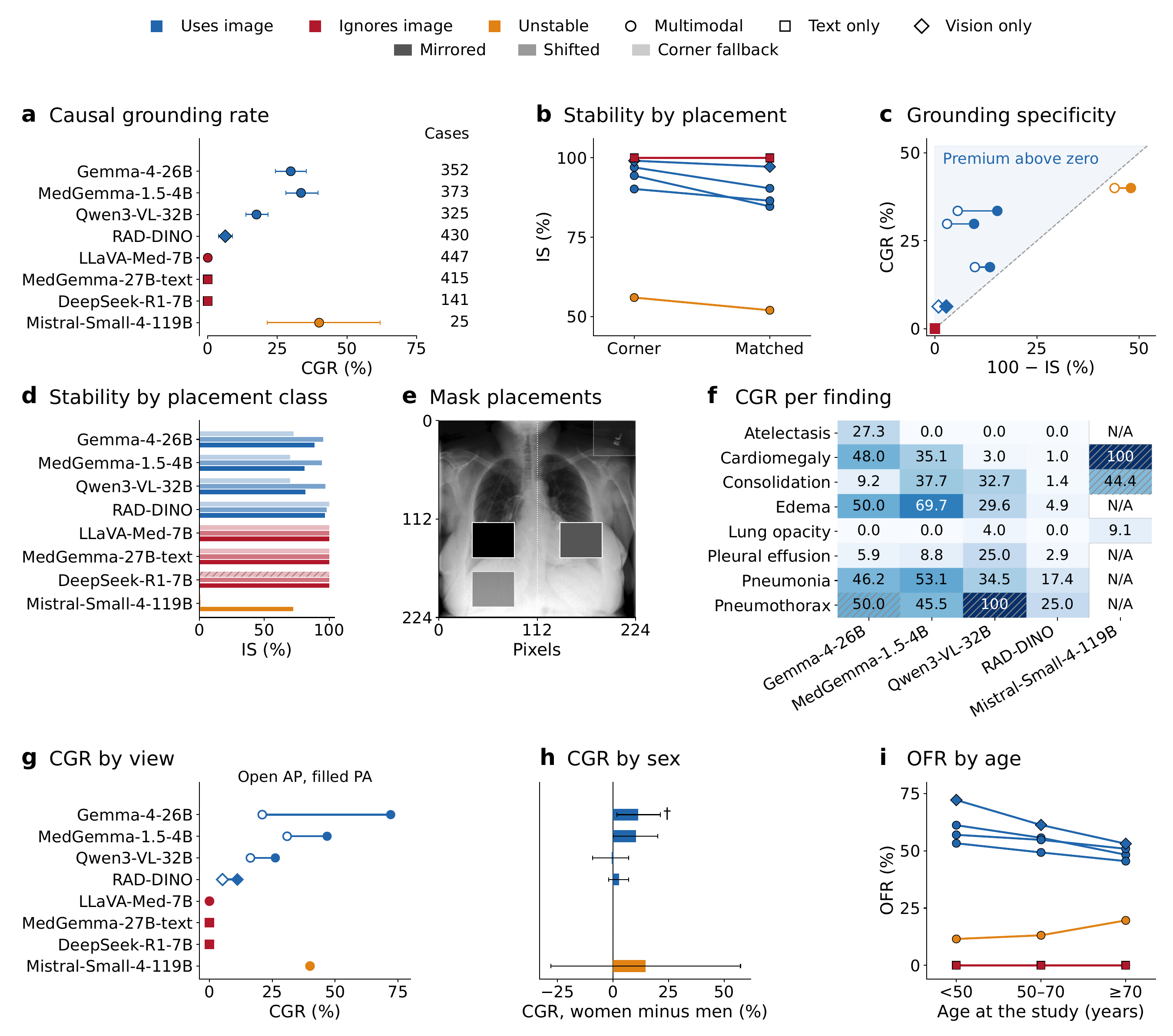}
\caption{Localization on the 452 MS-CXR cases, and subgroups. Fill color encodes the category (blue, uses image; red, ignores image; orange, unstable), and marker shape encodes the modality (circle, multimodal; square, text only; diamond, vision only). A marker is the value measured on its cases. Its error bar, drawn in \textbf{a} and \textbf{h} only, is the 95\% confidence interval, the 2.5th to the 97.5th percentile of 1{,}000 bootstrap resamples of that value, which draw patients in \textbf{a} and the cases of each group separately in \textbf{h}. A hatch marks a value on fewer than 10 cases. Case counts per panel: 25 to 447 (\textbf{a}--\textbf{c}), 1 to 322 (\textbf{d}), 1 to 100 (\textbf{f}), 5 to 366 (\textbf{g}), 8 to 263 (\textbf{h}), 115 to 582 (\textbf{i}). \textbf{a}, CGR under the target mask, with the case count. \textbf{b}, IS under both mask placements. \textbf{c}, CGR against 100 minus IS under the corner (open) and the matched (filled) placement, where the dashed diagonal marks a GSP of zero. \textbf{d}, IS under the matched mask by placement class, with a tick at zero. \textbf{e}, The target mask (black) and the candidate placements. The radiograph is a public NIH ChestX-ray14 example \cite{wang2017chestxray8}, since the MIMIC-CXR and MS-CXR images may not be reproduced. No result was computed on ChestX-ray14. \textbf{f}, CGR per finding for the systems that change any answer. N/A marks a finding with no correct answer. \textbf{g}, CGR by view. \textbf{h}, CGR in women minus men. A dagger marks $p_{\mathrm{FDR}} < 0.05$ (permutation test, nine tests per system). \textbf{i}, OFR across the age bands. AP, anteroposterior; CGR, causal grounding rate; FDR, false discovery rate; GSP, grounding specificity premium; IS, irrelevant-mask stability; OFR, opposite-label flip rate; PA, posteroanterior.}
\label{fig:localization}
\end{figure}

On the 452 MS-CXR cases, we occluded the radiologist-marked region of the queried finding with the target mask and counted the correct answers that changed (Table~\ref{tab:localization}; Fig.~\ref{fig:localization}a). We call the share of changed answers the causal grounding rate (CGR). It is 33.5 $\pm$ 3.0 [28.1, 39.6] for MedGemma-1.5-4B, 29.8 $\pm$ 2.7 [24.3, 35.4] for Gemma-4-26B, 17.5 $\pm$ 2.0 [13.7, 21.7] for Qwen3-VL-32B, and 6.3 $\pm$ 1.2 [3.9, 8.8] for the vision-only reference. CGR is 0.0 for the ignores-image systems and 40.0 $\pm$ 10.4 [21.4, 61.9] for Mistral-Small-4-119B on the 25 cases on which it answers Yes. The target mask of a case covers one box. For the 176 cases that belong to the 88 radiograph-finding pairs with two or more boxes, the other boxes stay visible. On the 276 cases with a single box, CGR is 37.6 for Gemma-4-26B, 35.3 for MedGemma-1.5-4B, 13.3 for Qwen3-VL-32B, and 7.4 for the vision-only reference. Over all answered cases, correct or not, the target mask changes the answer on 17.0 to 28.1 of the cases for the multimodal image users and on 6.9 for the vision-only reference (Supplementary Fig.~\ref{sfig:localization}d).

\begin{table}[p]
\centering
\caption{Localization metrics of all systems on the MS-CXR block of the MIMIC question set ($n$ = 452 cases with a radiologist-marked box, 321 patients). IS is reported under the corner and under the anatomically matched placement, and GSP is reported under each placement. For the text-only controls, IS under the matched placement is 100.0 by construction, because that condition reuses their original answer. Under the matched placement, the box is mirrored across the vertical midline, shifted along the same side where the mirrored box overlaps the target, and placed at the corner for the 11 boxes too large for a mirrored or a shifted position. Each value is the mean over its cases $\pm$ the standard deviation of 1{,}000 patient-cluster bootstrap resamples, with the percentile 95\% confidence interval and the case count $n$. CGR, causal grounding rate; IS, irrelevant-mask stability; GSP, grounding specificity premium.}
\label{tab:localization}
\setlength{\tabcolsep}{3pt}
\renewcommand{\arraystretch}{1.05}
\scriptsize
\begin{tabular}{@{}p{0.15\textwidth}p{0.145\textwidth}p{0.145\textwidth}p{0.145\textwidth}p{0.145\textwidth}p{0.145\textwidth}@{}}
\toprule
Model & CGR & IS (corner) & IS (matched) & GSP (corner) & GSP (matched) \\
\midrule
\multicolumn{6}{@{}l}{\textbf{General-purpose multimodal}} \\
Gemma-4-26B & \begin{tabular}[t]{@{}l@{}}29.8 $\pm$ 2.7\\ {\scriptsize [24.3, 35.4]}\\ {\scriptsize $n=352$}\end{tabular} & \begin{tabular}[t]{@{}l@{}}97.0 $\pm$ 0.9\\ {\scriptsize [95.1, 98.7]}\\ {\scriptsize $n=366$}\end{tabular} & \begin{tabular}[t]{@{}l@{}}90.4 $\pm$ 1.8\\ {\scriptsize [86.8, 93.6]}\\ {\scriptsize $n=363$}\end{tabular} & \begin{tabular}[t]{@{}l@{}}$+27.0$ $\pm$ 2.5\\ {\scriptsize [+22.2, +31.8]}\\ {\scriptsize $n=352$}\end{tabular} & \begin{tabular}[t]{@{}l@{}}$+19.8$ $\pm$ 2.5\\ {\scriptsize [+15.1, +24.8]}\\ {\scriptsize $n=348$}\end{tabular} \\
Qwen3-VL-32B & \begin{tabular}[t]{@{}l@{}}17.5 $\pm$ 2.0\\ {\scriptsize [13.7, 21.7]}\\ {\scriptsize $n=325$}\end{tabular} & \begin{tabular}[t]{@{}l@{}}90.2 $\pm$ 1.8\\ {\scriptsize [86.5, 93.3]}\\ {\scriptsize $n=325$}\end{tabular} & \begin{tabular}[t]{@{}l@{}}86.5 $\pm$ 1.9\\ {\scriptsize [82.6, 90.4]}\\ {\scriptsize $n=325$}\end{tabular} & \begin{tabular}[t]{@{}l@{}}$+7.7$ $\pm$ 2.4\\ {\scriptsize [+3.0, +12.4]}\\ {\scriptsize $n=325$}\end{tabular} & \begin{tabular}[t]{@{}l@{}}$+4.0$ $\pm$ 2.0\\ {\scriptsize [+0.3, +8.0]}\\ {\scriptsize $n=325$}\end{tabular} \\
Mistral-Small-4-119B & \begin{tabular}[t]{@{}l@{}}40.0 $\pm$ 10.4\\ {\scriptsize [21.4, 61.9]}\\ {\scriptsize $n=25$}\end{tabular} & \begin{tabular}[t]{@{}l@{}}56.0 $\pm$ 11.8\\ {\scriptsize [31.8, 76.9]}\\ {\scriptsize $n=25$}\end{tabular} & \begin{tabular}[t]{@{}l@{}}52.0 $\pm$ 11.1\\ {\scriptsize [28.6, 72.0]}\\ {\scriptsize $n=25$}\end{tabular} & \begin{tabular}[t]{@{}l@{}}$-4.0$ $\pm$ 8.6\\ {\scriptsize [-20.9, +13.0]}\\ {\scriptsize $n=25$}\end{tabular} & \begin{tabular}[t]{@{}l@{}}$-8.0$ $\pm$ 11.1\\ {\scriptsize [-29.6, +13.6]}\\ {\scriptsize $n=25$}\end{tabular} \\
\midrule
\multicolumn{6}{@{}l}{\textbf{Medical multimodal}} \\
MedGemma-1.5-4B & \begin{tabular}[t]{@{}l@{}}33.5 $\pm$ 3.0\\ {\scriptsize [28.1, 39.6]}\\ {\scriptsize $n=373$}\end{tabular} & \begin{tabular}[t]{@{}l@{}}94.4 $\pm$ 1.5\\ {\scriptsize [91.1, 97.0]}\\ {\scriptsize $n=373$}\end{tabular} & \begin{tabular}[t]{@{}l@{}}84.7 $\pm$ 2.4\\ {\scriptsize [79.8, 89.1]}\\ {\scriptsize $n=373$}\end{tabular} & \begin{tabular}[t]{@{}l@{}}$+27.9$ $\pm$ 2.6\\ {\scriptsize [+23.2, +33.3]}\\ {\scriptsize $n=373$}\end{tabular} & \begin{tabular}[t]{@{}l@{}}$+18.2$ $\pm$ 2.6\\ {\scriptsize [+13.5, +23.7]}\\ {\scriptsize $n=373$}\end{tabular} \\
LLaVA-Med-7B & \begin{tabular}[t]{@{}l@{}}0.0 $\pm$ 0.0\\ {\scriptsize [0.0, 0.0]}\\ {\scriptsize $n=447$}\end{tabular} & \begin{tabular}[t]{@{}l@{}}100.0 $\pm$ 0.0\\ {\scriptsize [100.0, 100.0]}\\ {\scriptsize $n=444$}\end{tabular} & \begin{tabular}[t]{@{}l@{}}100.0 $\pm$ 0.0\\ {\scriptsize [100.0, 100.0]}\\ {\scriptsize $n=447$}\end{tabular} & \begin{tabular}[t]{@{}l@{}}$+0.0$ $\pm$ 0.0\\ {\scriptsize [+0.0, +0.0]}\\ {\scriptsize $n=444$}\end{tabular} & \begin{tabular}[t]{@{}l@{}}$+0.0$ $\pm$ 0.0\\ {\scriptsize [+0.0, +0.0]}\\ {\scriptsize $n=447$}\end{tabular} \\
\midrule
\multicolumn{6}{@{}l}{\textbf{Text-only controls}} \\
MedGemma-27B-text & \begin{tabular}[t]{@{}l@{}}0.0 $\pm$ 0.0\\ {\scriptsize [0.0, 0.0]}\\ {\scriptsize $n=415$}\end{tabular} & \begin{tabular}[t]{@{}l@{}}100.0 $\pm$ 0.0\\ {\scriptsize [100.0, 100.0]}\\ {\scriptsize $n=415$}\end{tabular} & \begin{tabular}[t]{@{}l@{}}100.0 $\pm$ 0.0\\ {\scriptsize [100.0, 100.0]}\\ {\scriptsize $n=415$}\end{tabular} & \begin{tabular}[t]{@{}l@{}}$+0.0$ $\pm$ 0.0\\ {\scriptsize [+0.0, +0.0]}\\ {\scriptsize $n=415$}\end{tabular} & \begin{tabular}[t]{@{}l@{}}$+0.0$ $\pm$ 0.0\\ {\scriptsize [+0.0, +0.0]}\\ {\scriptsize $n=415$}\end{tabular} \\
DeepSeek-R1-7B & \begin{tabular}[t]{@{}l@{}}0.0 $\pm$ 0.0\\ {\scriptsize [0.0, 0.0]}\\ {\scriptsize $n=141$}\end{tabular} & \begin{tabular}[t]{@{}l@{}}100.0 $\pm$ 0.0\\ {\scriptsize [100.0, 100.0]}\\ {\scriptsize $n=141$}\end{tabular} & \begin{tabular}[t]{@{}l@{}}100.0 $\pm$ 0.0\\ {\scriptsize [100.0, 100.0]}\\ {\scriptsize $n=141$}\end{tabular} & \begin{tabular}[t]{@{}l@{}}$+0.0$ $\pm$ 0.0\\ {\scriptsize [+0.0, +0.0]}\\ {\scriptsize $n=141$}\end{tabular} & \begin{tabular}[t]{@{}l@{}}$+0.0$ $\pm$ 0.0\\ {\scriptsize [+0.0, +0.0]}\\ {\scriptsize $n=141$}\end{tabular} \\
\midrule
\multicolumn{6}{@{}l}{\textbf{Vision-only reference}} \\
RAD-DINO & \begin{tabular}[t]{@{}l@{}}6.3 $\pm$ 1.2\\ {\scriptsize [3.9, 8.8]}\\ {\scriptsize $n=430$}\end{tabular} & \begin{tabular}[t]{@{}l@{}}99.1 $\pm$ 0.5\\ {\scriptsize [98.1, 99.8]}\\ {\scriptsize $n=430$}\end{tabular} & \begin{tabular}[t]{@{}l@{}}97.2 $\pm$ 0.8\\ {\scriptsize [95.7, 98.8]}\\ {\scriptsize $n=430$}\end{tabular} & \begin{tabular}[t]{@{}l@{}}$+5.3$ $\pm$ 1.2\\ {\scriptsize [+3.1, +7.9]}\\ {\scriptsize $n=430$}\end{tabular} & \begin{tabular}[t]{@{}l@{}}$+3.5$ $\pm$ 1.1\\ {\scriptsize [+1.4, +5.7]}\\ {\scriptsize $n=430$}\end{tabular} \\
\bottomrule
\end{tabular}
\end{table}

To test whether any occlusion changes the answers, we also placed a mask of the same size elsewhere in the image. The share of correct answers unchanged under it is the irrelevant-mask stability (IS). IS depends on where the mask is placed (Fig.~\ref{fig:localization}b). With the mask at the image corner farthest from the box, IS is 90.2 to 97.0 for the multimodal image users and 99.1 for the vision-only reference. With the mask at the anatomically matched position, IS is 84.7 to 90.4 for the multimodal image users and 97.2 for the vision-only reference. In 322 of the 452 cases, this position is the box mirrored across the midline. For the multimodal image users, the matched placement lowers IS by 3.7 (Qwen3-VL-32B) to 9.7 (MedGemma-1.5-4B). Under the matched placement, their IS is 81.1 to 88.8 on the cases masked at the mirrored position, 94.5 to 97.2 on those masked at a shifted position, and 70.0 to 72.7 on the 11 cases masked at the corner (Fig.~\ref{fig:localization}d,e; Supplementary Table~\ref{stab:is_placement}). In these 11 cases, the corner rectangle overlaps the target box. The grounding specificity premium (GSP) is CGR minus (100 minus IS). It is positive under both placements for every image user, with every 95\% interval above zero (Fig.~\ref{fig:localization}c). With the corner mask, GSP ranges from 5.3 for the vision-only reference to 27.9 for MedGemma-1.5-4B. With the matched mask, it ranges from 3.5 $\pm$ 1.1 [1.4, 5.7] to 19.8 $\pm$ 2.5 [15.1, 24.8]. Occluding the marked region therefore changes more answers than occluding a region of the same size elsewhere. The matched placement reduces GSP by 1.9 to 9.7. Mistral-Small-4-119B has a low IS only on its 25 affirmative answers, 56.0 under the corner and 52.0 under the matched placement. Of all 452 of its answers, 95.4 stay unchanged under the corner placement and 94.9 under the matched placement (Supplementary Fig.~\ref{sfig:localization}c). On the 290 cases answered correctly by every image user, CGR is 28.6 $\pm$ 2.9 [23.2, 34.1] for MedGemma-1.5-4B, 25.2 for Gemma-4-26B, 15.5 for Qwen3-VL-32B, and 2.4 $\pm$ 0.9 [1.0, 4.3] for the vision-only reference (Supplementary Fig.~\ref{sfig:localization}e). For the image users, IS on these cases is 91.0 to 100.0 under the corner placement and 87.9 to 99.7 under the matched placement (Supplementary Fig.~\ref{sfig:localization}f). If a correct answer that becomes a non-answer under a mask counts as changed, the CGR of Gemma-4-26B increases from 29.8 to 33.2. No IS changes by more than 1.7.

For every image user, CGR is 0.0 to 4.0 on lung opacity. On atelectasis, it is 0.0 for every image user except Gemma-4-26B (27.3 on 33 cases). The findings with the highest CGR differ between the systems (Fig.~\ref{fig:localization}f; Supplementary Table~\ref{stab:cgr_finding}). MedGemma-1.5-4B has a CGR of 69.7 on edema and 53.1 on pneumonia, and Gemma-4-26B has 50.0 on edema, 48.0 on cardiomegaly, and 46.2 on pneumonia. On cardiomegaly, the CGR is 35.1 for MedGemma-1.5-4B, 3.0 for Qwen3-VL-32B, and 1.0 for the vision-only reference. Qwen3-VL-32B has a CGR of 34.5 on pneumonia, 32.7 on consolidation, and 25.0 on pleural effusion. On consolidation and pleural effusion, the CGR of Gemma-4-26B is 9.2 and 5.9. Among the findings with at least 10 correct answers, IS is lowest on edema for every multimodal image user, at 72.7 to 88.1 under the corner and 42.4 to 66.7 under the matched placement (Supplementary Fig.~\ref{sfig:localization}a,b). OFR also differs between findings. The image users flip 66.7 to 91.6 of their correct answers on pneumonia and consolidation, whereas three of the four flip 2.3 to 5.7 on atelectasis (Supplementary Fig.~\ref{sfig:swaps}g).

Image use also varies with the acquisition and the patient (Table~\ref{tab:subgroups}; Fig.~\ref{fig:localization}g--i). CGR is higher on posteroanterior (PA) than on anteroposterior (AP) radiographs for every image user. The difference is significant for Gemma-4-26B (72.1 vs 21.0, $p_{\mathrm{FDR}} = 0.004$). For the other image users, the values are 46.8 vs 30.9 for MedGemma-1.5-4B, 26.2 vs 16.3 for Qwen3-VL-32B, and 11.1 vs 5.2 for the vision-only reference. OFR is higher on PA radiographs for Gemma-4-26B (58.3 vs 50.9, $p_{\mathrm{FDR}} = 0.040$) and the vision-only reference (66.3 vs 56.3, $p_{\mathrm{FDR}} = 0.004$). The views also differ in their mix of findings, since cardiomegaly accounts for 26 of the 86 PA and 74 of the 366 AP cases of the MS-CXR block. These comparisons are not adjusted for this mix. By sex, Gemma-4-26B has a higher CGR in women than in men (36.7 vs 25.4, $p_{\mathrm{FDR}} = 0.040$). The vision-only reference has a higher OFR in women (63.0 vs 56.5, $p_{\mathrm{FDR}} = 0.047$). No other sex difference is significant after correction. The largest difference in accuracy between women and men is 5.9, for DeepSeek-R1-7B on the MIMIC question set (Supplementary Tables~\ref{stab:sex} and~\ref{stab:sex_chexpert}). Across the age bands, Gemma-4-26B's UAR increases from 73.3 to 86.2 and its OFR decreases from 61.2 to 48.3 ($p_{\mathrm{FDR}} = 0.004$ and $0.009$). The vision-only reference's OFR decreases from 72.2 to 53.1 ($p_{\mathrm{FDR}} = 0.004$).

\begin{table}[p]
\centering
\caption{Subgroup comparisons by sex, view, and age on the MIMIC question set ($n$ = 2{,}548 cases). Each cell shows the rate and the case count of each group and the $p$-value of a permutation test with 1{,}000 permutations, FDR-corrected within the nine tests of a model. A \textdagger{} marks $p_{\mathrm{FDR}} < 0.05$. The statistic is the absolute difference in means for sex and view and the one-way $F$ statistic for age. The age bands are under 50, 50 to under 70, and 70 and over. UAR is computed on every correct-on-original answer, the sentence questions included. OFR is computed on the correct-on-original answers of the finding-presence cases. The CGR of Mistral-Small-4-119B is computed on 25 answers and is not interpreted. N/A marks an age band with fewer than two cases. CGR, causal grounding rate; F, female; M, male; PA, posteroanterior; AP, anteroposterior; UAR, unrelated-image answer rate; OFR, opposite-label flip rate; FDR, false discovery rate.}
\label{tab:subgroups}
\setlength{\tabcolsep}{3pt}
\renewcommand{\arraystretch}{1.1}
\scriptsize
\begin{tabular}{@{}p{0.145\textwidth}p{0.085\textwidth}p{0.215\textwidth}p{0.215\textwidth}p{0.215\textwidth}@{}}
\toprule
Model & Attribute & CGR & UAR & OFR \\
\midrule
\multicolumn{5}{@{}l}{\textbf{Uses image}} \\
Gemma-4-26B & \begin{tabular}[t]{@{}l@{}}Sex\\ {\scriptsize F / M}\end{tabular} & \begin{tabular}[t]{@{}l@{}}36.7 / 25.4\\ {\scriptsize $p_{\mathrm{FDR}}$ = 0.040\textdagger{}}\\ {\scriptsize $n$ = 139 / 213}\end{tabular} & \begin{tabular}[t]{@{}l@{}}79.8 / 82.2\\ {\scriptsize $p_{\mathrm{FDR}}$ = 0.284}\\ {\scriptsize $n$ = 783 / 906}\end{tabular} & \begin{tabular}[t]{@{}l@{}}54.0 / 53.3\\ {\scriptsize $p_{\mathrm{FDR}}$ = 0.803}\\ {\scriptsize $n$ = 554 / 646}\end{tabular} \\
 & \begin{tabular}[t]{@{}l@{}}View\\ {\scriptsize PA / AP}\end{tabular} & \begin{tabular}[t]{@{}l@{}}72.1 / 21.0\\ {\scriptsize $p_{\mathrm{FDR}}$ = 0.004\textdagger{}}\\ {\scriptsize $n$ = 61 / 291}\end{tabular} & \begin{tabular}[t]{@{}l@{}}77.2 / 83.1\\ {\scriptsize $p_{\mathrm{FDR}}$ = 0.013\textdagger{}}\\ {\scriptsize $n$ = 580 / 1109}\end{tabular} & \begin{tabular}[t]{@{}l@{}}58.3 / 50.9\\ {\scriptsize $p_{\mathrm{FDR}}$ = 0.040\textdagger{}}\\ {\scriptsize $n$ = 434 / 766}\end{tabular} \\
 & Age band & \begin{tabular}[t]{@{}l@{}}22.2 / 28.5 / 32.6\\ {\scriptsize $p_{\mathrm{FDR}}$ = 0.488}\\ {\scriptsize $n$ = 36 / 144 / 172}\end{tabular} & \begin{tabular}[t]{@{}l@{}}73.3 / 78.4 / 86.2\\ {\scriptsize $p_{\mathrm{FDR}}$ = 0.004\textdagger{}}\\ {\scriptsize $n$ = 236 / 713 / 740}\end{tabular} & \begin{tabular}[t]{@{}l@{}}61.2 / 55.7 / 48.3\\ {\scriptsize $p_{\mathrm{FDR}}$ = 0.009\textdagger{}}\\ {\scriptsize $n$ = 224 / 465 / 511}\end{tabular} \\
\midrule
Qwen3-VL-32B & \begin{tabular}[t]{@{}l@{}}Sex\\ {\scriptsize F / M}\end{tabular} & \begin{tabular}[t]{@{}l@{}}17.2 / 17.8\\ {\scriptsize $p_{\mathrm{FDR}}$ = 0.877}\\ {\scriptsize $n$ = 134 / 191}\end{tabular} & \begin{tabular}[t]{@{}l@{}}79.8 / 77.1\\ {\scriptsize $p_{\mathrm{FDR}}$ = 0.301}\\ {\scriptsize $n$ = 779 / 877}\end{tabular} & \begin{tabular}[t]{@{}l@{}}44.6 / 51.8\\ {\scriptsize $p_{\mathrm{FDR}}$ = 0.180}\\ {\scriptsize $n$ = 534 / 616}\end{tabular} \\
 & \begin{tabular}[t]{@{}l@{}}View\\ {\scriptsize PA / AP}\end{tabular} & \begin{tabular}[t]{@{}l@{}}26.2 / 16.3\\ {\scriptsize $p_{\mathrm{FDR}}$ = 0.286}\\ {\scriptsize $n$ = 42 / 283}\end{tabular} & \begin{tabular}[t]{@{}l@{}}79.4 / 77.9\\ {\scriptsize $p_{\mathrm{FDR}}$ = 0.534}\\ {\scriptsize $n$ = 544 / 1112}\end{tabular} & \begin{tabular}[t]{@{}l@{}}46.0 / 49.7\\ {\scriptsize $p_{\mathrm{FDR}}$ = 0.307}\\ {\scriptsize $n$ = 398 / 752}\end{tabular} \\
 & Age band & \begin{tabular}[t]{@{}l@{}}17.2 / 22.3 / 13.9\\ {\scriptsize $p_{\mathrm{FDR}}$ = 0.286}\\ {\scriptsize $n$ = 29 / 130 / 166}\end{tabular} & \begin{tabular}[t]{@{}l@{}}72.6 / 77.9 / 80.6\\ {\scriptsize $p_{\mathrm{FDR}}$ = 0.180}\\ {\scriptsize $n$ = 226 / 702 / 728}\end{tabular} & \begin{tabular}[t]{@{}l@{}}53.3 / 49.3 / 45.5\\ {\scriptsize $p_{\mathrm{FDR}}$ = 0.286}\\ {\scriptsize $n$ = 210 / 450 / 490}\end{tabular} \\
\midrule
MedGemma-1.5-4B & \begin{tabular}[t]{@{}l@{}}Sex\\ {\scriptsize F / M}\end{tabular} & \begin{tabular}[t]{@{}l@{}}39.5 / 29.2\\ {\scriptsize $p_{\mathrm{FDR}}$ = 0.256}\\ {\scriptsize $n$ = 157 / 216}\end{tabular} & \begin{tabular}[t]{@{}l@{}}75.5 / 78.8\\ {\scriptsize $p_{\mathrm{FDR}}$ = 0.297}\\ {\scriptsize $n$ = 713 / 770}\end{tabular} & \begin{tabular}[t]{@{}l@{}}55.9 / 51.2\\ {\scriptsize $p_{\mathrm{FDR}}$ = 0.297}\\ {\scriptsize $n$ = 599 / 650}\end{tabular} \\
 & \begin{tabular}[t]{@{}l@{}}View\\ {\scriptsize PA / AP}\end{tabular} & \begin{tabular}[t]{@{}l@{}}46.8 / 30.9\\ {\scriptsize $p_{\mathrm{FDR}}$ = 0.198}\\ {\scriptsize $n$ = 62 / 311}\end{tabular} & \begin{tabular}[t]{@{}l@{}}76.5 / 77.6\\ {\scriptsize $p_{\mathrm{FDR}}$ = 0.845}\\ {\scriptsize $n$ = 506 / 977}\end{tabular} & \begin{tabular}[t]{@{}l@{}}53.6 / 53.4\\ {\scriptsize $p_{\mathrm{FDR}}$ = 0.955}\\ {\scriptsize $n$ = 427 / 822}\end{tabular} \\
 & Age band & \begin{tabular}[t]{@{}l@{}}32.4 / 31.6 / 35.4\\ {\scriptsize $p_{\mathrm{FDR}}$ = 0.845}\\ {\scriptsize $n$ = 34 / 158 / 181}\end{tabular} & \begin{tabular}[t]{@{}l@{}}75.6 / 76.6 / 78.4\\ {\scriptsize $p_{\mathrm{FDR}}$ = 0.845}\\ {\scriptsize $n$ = 217 / 619 / 647}\end{tabular} & \begin{tabular}[t]{@{}l@{}}57.0 / 54.8 / 50.9\\ {\scriptsize $p_{\mathrm{FDR}}$ = 0.478}\\ {\scriptsize $n$ = 214 / 491 / 544}\end{tabular} \\
\midrule
RAD-DINO & \begin{tabular}[t]{@{}l@{}}Sex\\ {\scriptsize F / M}\end{tabular} & \begin{tabular}[t]{@{}l@{}}7.9 / 5.2\\ {\scriptsize $p_{\mathrm{FDR}}$ = 0.362}\\ {\scriptsize $n$ = 178 / 252}\end{tabular} & \begin{tabular}[t]{@{}l@{}}80.5 / 83.8\\ {\scriptsize $p_{\mathrm{FDR}}$ = 0.190}\\ {\scriptsize $n$ = 619 / 718}\end{tabular} & \begin{tabular}[t]{@{}l@{}}63.0 / 56.5\\ {\scriptsize $p_{\mathrm{FDR}}$ = 0.047\textdagger{}}\\ {\scriptsize $n$ = 619 / 718}\end{tabular} \\
 & \begin{tabular}[t]{@{}l@{}}View\\ {\scriptsize PA / AP}\end{tabular} & \begin{tabular}[t]{@{}l@{}}11.1 / 5.2\\ {\scriptsize $p_{\mathrm{FDR}}$ = 0.113}\\ {\scriptsize $n$ = 81 / 349}\end{tabular} & \begin{tabular}[t]{@{}l@{}}80.5 / 83.1\\ {\scriptsize $p_{\mathrm{FDR}}$ = 0.346}\\ {\scriptsize $n$ = 430 / 907}\end{tabular} & \begin{tabular}[t]{@{}l@{}}66.3 / 56.3\\ {\scriptsize $p_{\mathrm{FDR}}$ = 0.004\textdagger{}}\\ {\scriptsize $n$ = 430 / 907}\end{tabular} \\
 & Age band & \begin{tabular}[t]{@{}l@{}}4.9 / 5.6 / 7.1\\ {\scriptsize $p_{\mathrm{FDR}}$ = 0.815}\\ {\scriptsize $n$ = 41 / 177 / 212}\end{tabular} & \begin{tabular}[t]{@{}l@{}}80.7 / 79.1 / 85.7\\ {\scriptsize $p_{\mathrm{FDR}}$ = 0.027\textdagger{}}\\ {\scriptsize $n$ = 223 / 532 / 582}\end{tabular} & \begin{tabular}[t]{@{}l@{}}72.2 / 61.3 / 53.1\\ {\scriptsize $p_{\mathrm{FDR}}$ = 0.004\textdagger{}}\\ {\scriptsize $n$ = 223 / 532 / 582}\end{tabular} \\
\midrule
\multicolumn{5}{@{}l}{\textbf{Unstable}} \\
Mistral-Small-4-119B & \begin{tabular}[t]{@{}l@{}}Sex\\ {\scriptsize F / M}\end{tabular} & \begin{tabular}[t]{@{}l@{}}50.0 / 35.3\\ {\scriptsize $p_{\mathrm{FDR}}$ = 0.863}\\ {\scriptsize $n$ = 8 / 17}\end{tabular} & \begin{tabular}[t]{@{}l@{}}87.0 / 84.4\\ {\scriptsize $p_{\mathrm{FDR}}$ = 0.486}\\ {\scriptsize $n$ = 570 / 617}\end{tabular} & \begin{tabular}[t]{@{}l@{}}15.0 / 15.0\\ {\scriptsize $p_{\mathrm{FDR}} > 0.999$}\\ {\scriptsize $n$ = 380 / 420}\end{tabular} \\
 & \begin{tabular}[t]{@{}l@{}}View\\ {\scriptsize PA / AP}\end{tabular} & \begin{tabular}[t]{@{}l@{}}40.0 / 40.0\\ {\scriptsize $p_{\mathrm{FDR}} > 0.999$}\\ {\scriptsize $n$ = 5 / 20}\end{tabular} & \begin{tabular}[t]{@{}l@{}}88.6 / 83.7\\ {\scriptsize $p_{\mathrm{FDR}}$ = 0.084}\\ {\scriptsize $n$ = 481 / 706}\end{tabular} & \begin{tabular}[t]{@{}l@{}}9.1 / 19.9\\ {\scriptsize $p_{\mathrm{FDR}}$ = 0.009\textdagger{}}\\ {\scriptsize $n$ = 363 / 437}\end{tabular} \\
 & Age band & \begin{tabular}[t]{@{}l@{}}N/A / 27.3 / 50.0\\ {\scriptsize $p_{\mathrm{FDR}}$ = 0.568}\\ {\scriptsize $n$ = 0 / 11 / 14}\end{tabular} & \begin{tabular}[t]{@{}l@{}}88.9 / 85.5 / 84.5\\ {\scriptsize $p_{\mathrm{FDR}}$ = 0.568}\\ {\scriptsize $n$ = 190 / 532 / 465}\end{tabular} & \begin{tabular}[t]{@{}l@{}}11.5 / 13.1 / 19.6\\ {\scriptsize $p_{\mathrm{FDR}}$ = 0.084}\\ {\scriptsize $n$ = 183 / 336 / 281}\end{tabular} \\
\bottomrule
\end{tabular}
\end{table}

\subsection*{The categories transfer unchanged to CheXpert}

\begin{figure}[p]
\centering
\includegraphics[width=\textwidth]{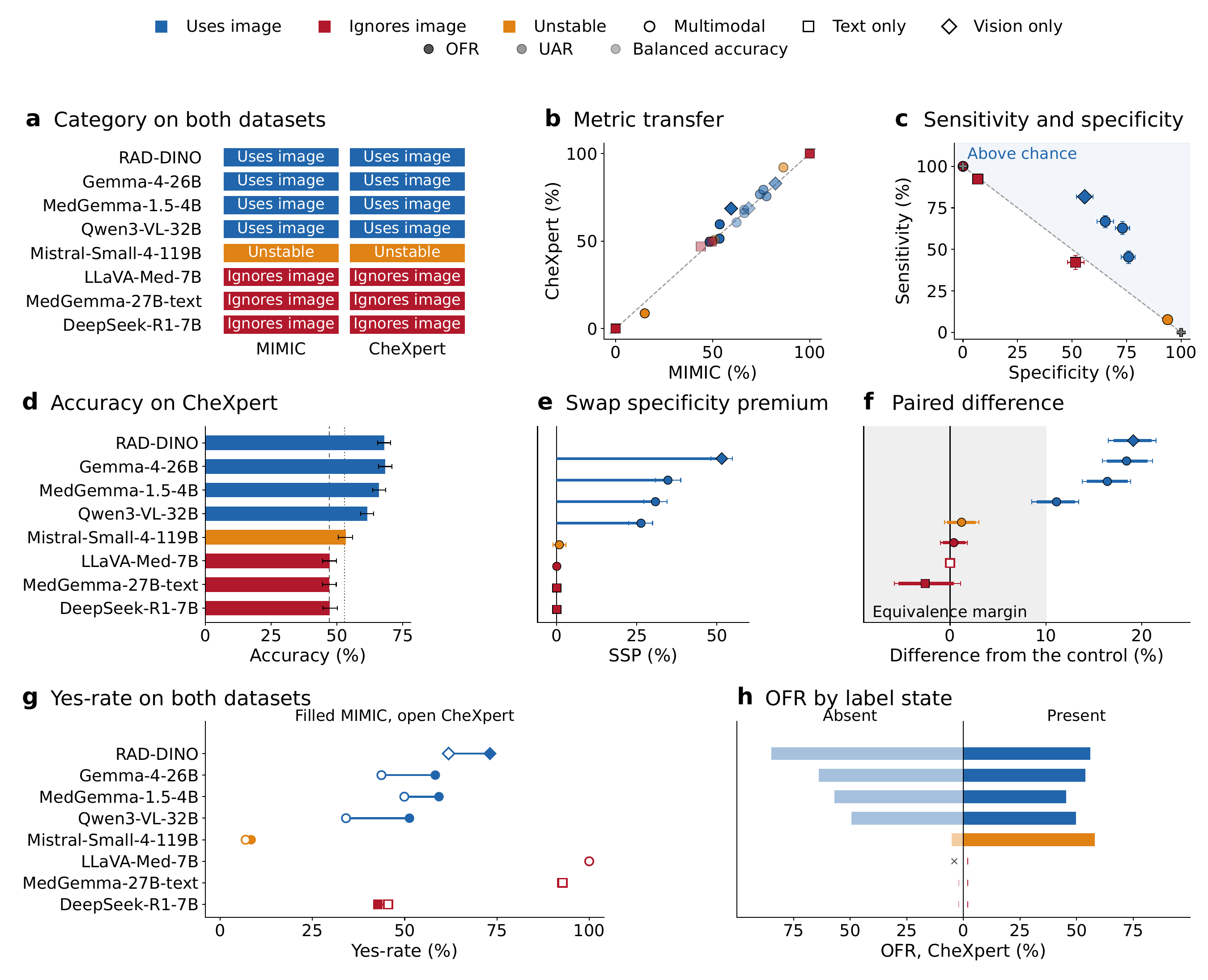}
\caption{Transfer of the categories and the metrics from the MIMIC finding-presence questions ($n$ = 1{,}852) to the CheXpert question set ($n$ = 1{,}380 questions from 1{,}285 patients). Fill color encodes the behavioral category (blue, uses image; red, ignores image; orange, unstable), and marker shape encodes modality (circle, multimodal; square, text only; diamond, vision only). A marker is the value measured on its cases, a rate everywhere except \textbf{f}, where it is a paired difference. An error bar, drawn in \textbf{c} to \textbf{f} only, spans the 2.5th to the 97.5th percentile of 1{,}000 patient-cluster bootstrap resamples of that value, which is the 95\% confidence interval. Case counts per panel: 604 to 1{,}852 (\textbf{b}), 600 to 730 per axis (\textbf{c}), 1{,}280 to 1{,}380 (\textbf{d}, \textbf{f}), 604 to 938 (\textbf{e}), 1{,}280 to 1{,}852 (\textbf{g}), 50 to 685 (\textbf{h}). CheXpert is in distribution for the vision-only reference, whose heads were fit on its training and validation images, and for no other system. \textbf{a}, The category assigned by the rule on each dataset, applied unchanged. \textbf{b}, Each metric on MIMIC against CheXpert, one point per system and metric, with the identity line. \textbf{c}, Sensitivity against specificity on CheXpert, where the dashed line marks a balanced accuracy of 50. \textbf{d}, Accuracy on CheXpert, where the dashed and dotted lines mark the constant answers. \textbf{e}, SSP on CheXpert. \textbf{f}, Paired difference in balanced accuracy against the text-only control MedGemma-27B-text, on the questions answered by both systems. The thin bar is the 95\% CI, and the thick bar is the 90\% interval of the two one-sided tests. The shaded band marks the equivalence margin of $\pm$10\%. \textbf{g}, Yes-rate on both datasets. \textbf{h}, OFR on CheXpert by label state, with a tick at zero and a cross where it is not defined. CI, confidence interval; OFR, opposite-label flip rate; SSP, swap specificity premium; UAR, unrelated-image answer rate.}
\label{fig:transfer}
\end{figure}

Applied unchanged to 1{,}380 CheXpert finding-presence questions built with the same stratification, the category rule assigns every system the same category as on the MIMIC question set (Fig.~\ref{fig:transfer}a; Supplementary Table~\ref{stab:chexpert}). For the text-only controls, which receive no image, this agreement is fixed by construction. On CheXpert, the OFR of the image users is 49.6 $\pm$ 1.7 [46.4, 52.9] (Qwen3-VL-32B) to 68.6 $\pm$ 1.5 [65.6, 71.3] (the vision-only reference), with an SSP of 26.3 to 51.5. Mistral-Small-4-119B flips 8.7 $\pm$ 1.0 [6.8, 10.7] of its correct answers, with an SSP of 0.8 $\pm$ 1.0 [$-1.2$, 2.9]. The ignores-image systems have an OFR of 0.0 and a UAR of 100.0 on both datasets (Fig.~\ref{fig:transfer}e,h). The other five systems keep the same order in OFR (Fig.~\ref{fig:transfer}b). Their order in UAR changes, and Gemma-4-26B moves from the third to the last place among them. Across all systems, the Spearman rank correlation of balanced accuracy on the finding-presence questions between the datasets is 0.976 ($p < 0.001$). On CheXpert, the control scores 47.1 $\pm$ 1.4 [44.5, 49.7], and the constant No answer scores 52.9. Its balanced accuracy is 49.6 $\pm$ 0.7 [48.3, 51.0] (Fig.~\ref{fig:transfer}d,g). In balanced accuracy, every image user exceeds it, by 11.1 for Qwen3-VL-32B to 19.1 for the vision-only reference (all $p_{\mathrm{FDR}} = 0.002$; Fig.~\ref{fig:transfer}c,f). The heads of the vision-only reference were fit on the training and validation images of the CheXpert list. As on the MIMIC question set, OFR is higher on PA than on AP radiographs for Gemma-4-26B (64.9 vs 54.8, $p_{\mathrm{FDR}} = 0.030$) and the vision-only reference (74.2 vs 63.8, $p_{\mathrm{FDR}} = 0.012$). No other difference by sex, view, or age band is significant on CheXpert.

\subsection*{The grounding rates keep their order under another phrasing and a higher resolution}

Under a radiologist-framed phrasing of the question, CGR on all 452 MS-CXR cases is 28.2, 22.4, 36.9, and 38.0 for Gemma-4-26B, Qwen3-VL-32B, MedGemma-1.5-4B, and Mistral-Small-4-119B. Under the default phrasing, the values are 29.8, 17.5, 33.5, and 40.0. IS is 89.4 to 97.1 for the multimodal image users under the radiologist-framed phrasing and 90.2 to 97.0 under the default phrasing. Under the radiologist-framed phrasing, LLaVA-Med-7B keeps a CGR of 0.0 and an IS of 100.0 (Supplementary Fig.~\ref{sfig:robustness}a--d). The text-only controls have these values by construction, because they answered the phrasings under the original image only. A terse phrasing that drops the single-word instruction changes how many of the 452 cases a system answers. With a budget of 128 tokens, LLaVA-Med-7B, Mistral-Small-4-119B, and the control answer 1.5, 0.2, and 0.0 of them. Mistral-Small-4-119B mostly writes an explanation that is cut off at the budget, LLaVA-Med-7B mostly returns an empty output, and the control mostly declines to answer without more information. Under this phrasing, every answer of LLaVA-Med-7B is a No taken from a sentence such as ``The chest No. is a reference number''. Gemma-4-26B and MedGemma-1.5-4B answer 54.2 and 55.3, and Qwen3-VL-32B and DeepSeek-R1-7B answer 100.0 and 98.2. On a 200-case subsample of the MIMIC-CXR block, the language systems score 46.0 to 61.8 in balanced accuracy under the three phrasings (Supplementary Fig.~\ref{sfig:robustness}e). Because neither swap was rerun under either phrasing, the categories are not recomputed there. Every system has at least 731 informative cases under each swap. A minimum informative count of 50, 100, or 200 cases therefore changes no category (Supplementary Fig.~\ref{sfig:robustness}h). The alternative rule built on the localization metrics assigns every system the same category as the swap rule under the corner placement. Under the matched placement, Qwen3-VL-32B and MedGemma-1.5-4B remain unassigned under this rule, because their IS is between 70 and 90. On 100 MS-CXR cases rerun at $512 \times 512$ pixels, CGR is 28.2 $\pm$ 5.2 [18.4, 38.5] for Gemma-4-26B, 16.7 for Qwen3-VL-32B, and 30.6 for MedGemma-1.5-4B. At $224 \times 224$ pixels, the values on the same cases are 34.3 $\pm$ 5.4 [24.3, 44.8], 14.7, and 37.3. On the 68, 62, and 75 cases answered correctly at both resolutions, the paired differences are $-7.4$, 0.0, and $-13.3$ (uncorrected $p = 0.126$, $p > 0.999$, and $p = 0.009$). At both resolutions, MedGemma-1.5-4B has the highest CGR of the three and Qwen3-VL-32B has the lowest (Supplementary Fig.~\ref{sfig:robustness}f). The systems that change no answer keep a CGR of 0.0 at $512 \times 512$ pixels. Excluding the 78 finding-absent cases with an uncertain label changes accuracy, balanced accuracy, UAR, OFR, and SSP by at most 2.8 on the finding-presence questions and the MIMIC-CXR block (Supplementary Fig.~\ref{sfig:robustness}g).

\subsection*{Confidence is not higher on grounded than on ungrounded answers}

\begin{figure}[p]
\centering
\includegraphics[width=\textwidth]{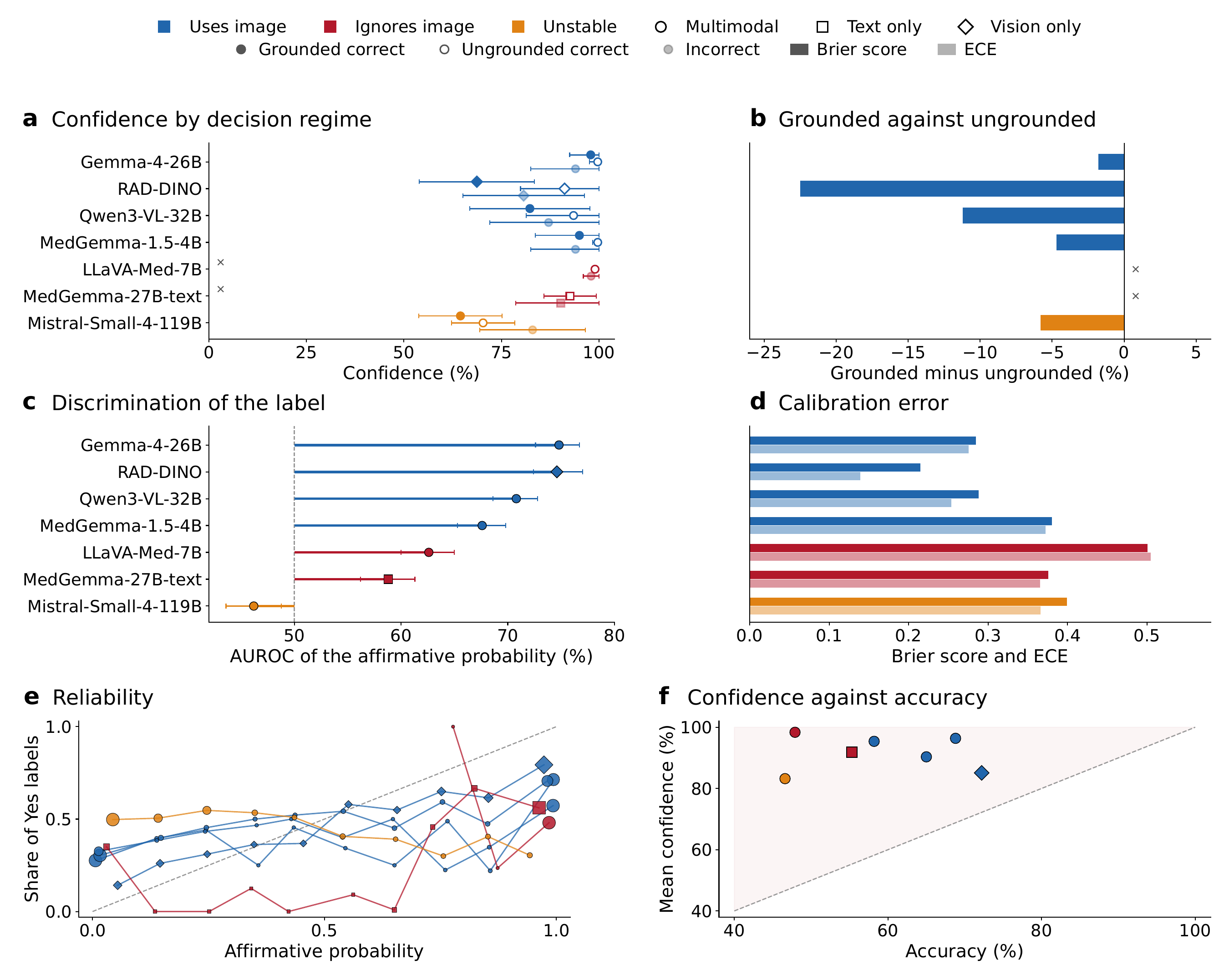}
\caption{Confidence and calibration on the MIMIC question set ($n$ = 2{,}548 questions). Confidence is the probability assigned to the chosen answer, in percent. Fill color encodes the behavioral category (blue, uses image; red, ignores image; orange, unstable), marker shape encodes modality (circle, multimodal; square, text only; diamond, vision only), and a cross marks a system or a regime with no value. Only \textbf{a} and \textbf{c} have error bars. Case counts per panel: 10 to 1{,}361 answers per regime (\textbf{a}, \textbf{b}), 1{,}852 to 2{,}548 answered questions (\textbf{c}--\textbf{f}). DeepSeek-R1-7B has no confidence and is left out. \textbf{a}, Confidence on the grounded correct MS-CXR answers (filled), the ungrounded correct MS-CXR answers (open), and the incorrect pooled answers (pale). The marker is the mean confidence of that regime and the error bar spans one standard deviation on each side of it. The count of every regime is in Supplementary Table~\ref{stab:confidence}. \textbf{b}, The difference between the first two means of \textbf{a}. \textbf{c}, AUROC of the affirmative probability as a detector of the Yes label. The marker is the AUROC measured on those questions and the error bar spans the 2.5th to the 97.5th percentile of 1{,}000 patient-cluster bootstrap resamples of it, which is the 95\% confidence interval. The dashed line is chance. \textbf{d}, Brier score (solid) and expected calibration error (pale), one value per system. \textbf{e}, Share of Yes labels against the mean affirmative probability in 10 equal-width bins, with the marker area proportional to the number of questions in the bin. The per-system diagrams are in Supplementary Fig.~\ref{sfig:reliability}. \textbf{f}, Mean confidence against accuracy on the pooled question set, where the shaded region above the diagonal marks confidence above accuracy. AUROC, area under the receiver operating characteristic curve; CI, confidence interval; ECE, expected calibration error.}
\label{fig:confidence}
\end{figure}

For every system except DeepSeek-R1-7B, which has no confidence, we split the correct MS-CXR answers into grounded answers, which change under the target mask, and ungrounded answers, which stay unchanged. We then compared the probability assigned by each system to its answer between grounded and ungrounded answers descriptively, without a statistical test (Fig.~\ref{fig:confidence}a,b; Supplementary Table~\ref{stab:confidence}). Mean confidence is lower on grounded than on ungrounded correct answers for every image user. The values are 97.9 $\pm$ 5.4 and 99.7 $\pm$ 2.1 for Gemma-4-26B (105 and 247 answers), 82.3 $\pm$ 15.4 and 93.5 $\pm$ 12.1 for Qwen3-VL-32B (57 and 268), 95.0 $\pm$ 11.3 and 99.7 $\pm$ 1.3 for MedGemma-1.5-4B (125 and 248), and 68.7 $\pm$ 14.8 and 91.2 $\pm$ 11.3 for the vision-only reference (27 and 403). Used to detect the label on the pooled questions, the affirmative probability has an area under the receiver operating characteristic curve (AUROC) of 74.8 $\pm$ 1.1 [72.6, 76.7] for Gemma-4-26B, 74.6 $\pm$ 1.2 [72.4, 77.0] for the vision-only reference on its 1{,}852 finding-presence questions, and 58.8 $\pm$ 1.3 [56.2, 61.3] for the control. For Mistral-Small-4-119B, it is below chance at 46.2 $\pm$ 1.3 [43.6, 48.8]. Within each finding of the MIMIC-CXR block, the finding-present and the finding-absent cases share one question. Computed within the findings, the AUROC is 50.0 for the control, 52.7 for LLaVA-Med-7B, 56.9 for Mistral-Small-4-119B, and 66.1 to 76.3 for the image users. The pooled AUROC of the control and of Mistral-Small-4-119B therefore reflects differences between the findings. The mean confidence of the incorrect answers is 80.7 to 98.0 (Fig.~\ref{fig:confidence}a). The expected calibration error is 0.139 for the vision-only reference and 0.254 to 0.505 for the language systems (Fig.~\ref{fig:confidence}c--f; Supplementary Figs.~\ref{sfig:confidence_histograms} and~\ref{sfig:reliability}).

\subsection*{Radiologists are more accurate than every system on a balanced set}

\begin{figure}[p]
\centering
\includegraphics[width=\textwidth]{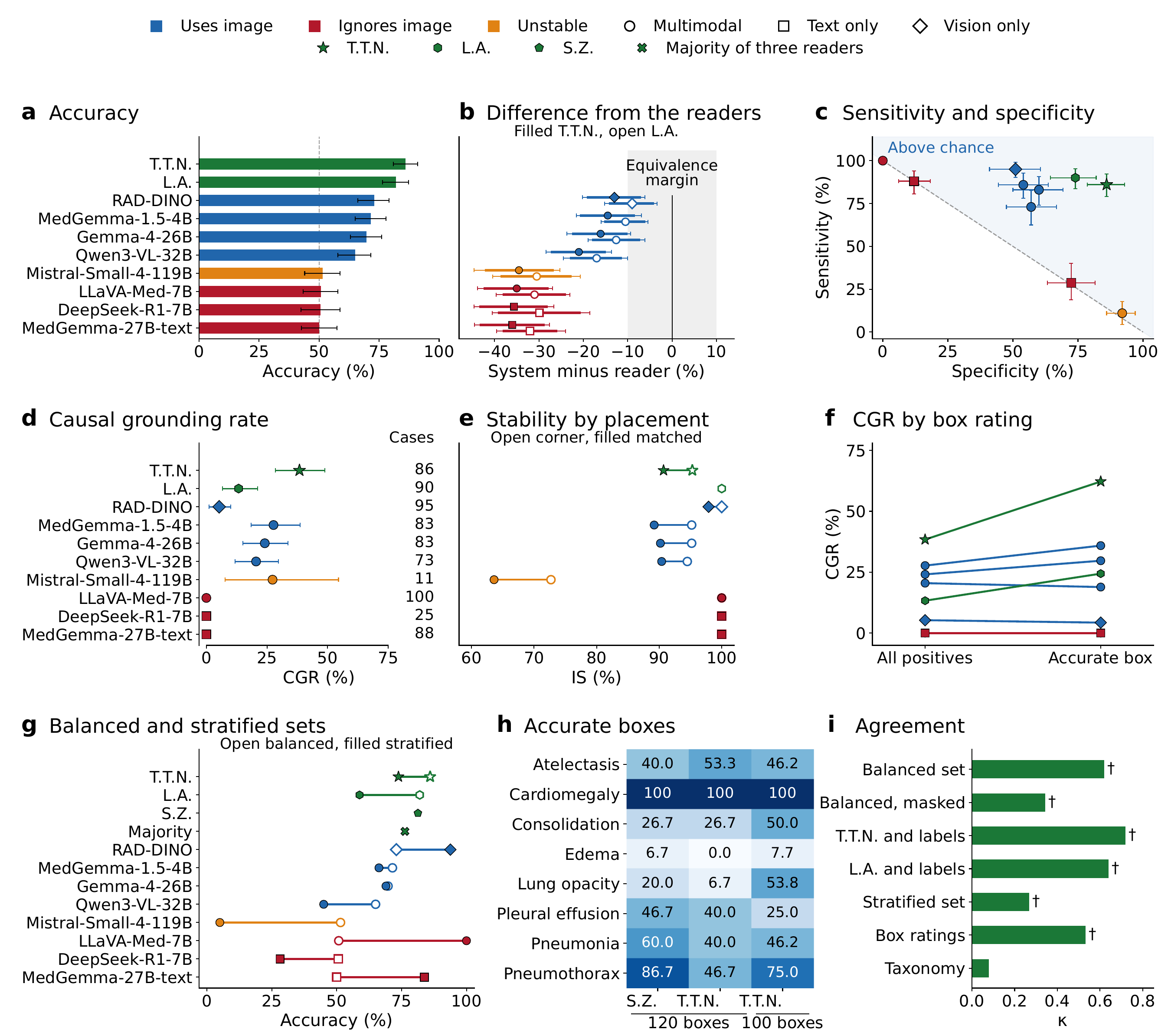}
\caption{Systems against three board-certified radiologists, two on the balanced set ($n$ = 200 cases) and all three on the difficulty-stratified set ($n$ = 80 finding-present cases). Fill color encodes the category of a system (blue, uses image; red, ignores image; orange, unstable), and marker shape encodes its modality (circle, multimodal; square, text only; diamond, vision only). Green marks a radiologist. A bar or marker is the value on its cases, and in \textbf{b} it is a paired difference. Its error bar, drawn in \textbf{a} to \textbf{d} only, is the 95\% confidence interval, the 2.5th to the 97.5th percentile of 1{,}000 patient-cluster bootstrap resamples of that value. Case counts per panel: 174 to 200 (\textbf{a}, \textbf{b}), 87 to 100 per axis (\textbf{c}), 11 to 100 (\textbf{d}, \textbf{e}), 21 to 100 (\textbf{f}), 71 to 200 (\textbf{g}), 12 to 15 per cell (\textbf{h}), 22 to 200 (\textbf{i}). \textbf{a}, Accuracy on the balanced set, with a dashed line at 50. \textbf{b}, System minus reader in accuracy, with the 90\% interval of the two one-sided tests as a thick bar and the equivalence margin of $\pm$10\% as a band. \textbf{c}, Sensitivity against specificity, where the dashed diagonal marks a balanced accuracy of 50. \textbf{d}, CGR on the 100 positives, with the case count. \textbf{e}, IS under both placements. \textbf{f}, CGR on all positives and on those with a box rated accurate by T.T.N., where each line has at least 10 cases per group. \textbf{g}, Accuracy on both sets. \textbf{h}, Share of boxes rated accurate, in the 120-box sample and the balanced set. \textbf{i}, Cohen's $\kappa$, or Fleiss' $\kappa$ for the three readers of the stratified set. A dagger marks $p < 0.05$ in a one-sided permutation test. CGR, causal grounding rate; IS, irrelevant-mask stability.}
\label{fig:readers}
\end{figure}

Three board-certified radiologists took part in the reader study, S.Z. (6 years of experience), L.A. (10 years), and T.T.N. (8 years of experience in diagnostic and interventional radiology). L.A. and T.T.N. read a balanced set of 200 cases with the same question as the systems, and all three read a second, difficulty-stratified set of 80 cases. The balanced set contains 100 MS-CXR cases with a box, in which the finding is present, and 100 MIMIC-CXR cases of the same eight findings, in which it is absent. The two readers give the same answer on 81.0 $\pm$ 2.8 [75.4, 86.1] of the 200 original displays, with a Cohen's $\kappa$ of 0.620 ($p < 0.001$; Fig.~\ref{fig:readers}i; Supplementary Table~\ref{stab:reader_agreement}). T.T.N. scores 86.0 $\pm$ 2.5 [81.0, 91.0], and L.A. scores 82.0 $\pm$ 2.7 [76.5, 87.2]. The systems score 50.0 to 73.0 (Fig.~\ref{fig:readers}a). On the cases answered by both the system and the reader, every system is less accurate than both readers, by 13.0 to 36.0 against T.T.N. and by 9.0 to 32.0 against L.A. ($p_{\mathrm{FDR}}$ = 0.001 to 0.005; Fig.~\ref{fig:readers}b; Supplementary Table~\ref{stab:reader_balanced}). For every system, the 90\% interval of the difference extends below $-10$. The two one-sided tests therefore establish equivalence with neither reader. With balanced accuracy, every test has the same outcome. The image users are less specific than both readers, by 14.0 to 35.0 ($p_{\mathrm{FDR}}$ = 0.001 to 0.003; Fig.~\ref{fig:readers}c; Supplementary Table~\ref{stab:reader_senspec}). Their sensitivity of 73.0 to 95.0 differs from that of the readers by $-17.0$ to $+9.0$. The text-only control answers Yes to 176 of the 200 questions, and LLaVA-Med-7B answers 197 questions, all with Yes.

On the 100 positives, 38.4 $\pm$ 5.2 [28.4, 48.8] of T.T.N.'s correct answers (86 cases) and 13.3 $\pm$ 3.6 [6.7, 21.2] of L.A.'s (90 cases) change under the target mask (Fig.~\ref{fig:readers}d). The multimodal image users have a CGR of 20.5 to 27.7, which lies between the values of the readers. On the positives answered correctly by both the system and the reader, these systems are 4.2 to 11.3 below T.T.N. ($p_{\mathrm{FDR}} \geq 0.104$) and 15.4 to 20.3 above L.A. ($p_{\mathrm{FDR}}$ = 0.004 to 0.011; Supplementary Table~\ref{stab:reader_masks}). T.T.N. rated 50 of the 100 boxes as accurate, meaning that the box covers the primary location of the finding. On these positives, the CGR of T.T.N. is 62.2 $\pm$ 7.1 [48.9, 75.6] and that of L.A. is 24.4 $\pm$ 6.5 [13.3, 37.8] (45 cases each). These systems have a CGR of 18.9 to 35.9 on the same positives (Fig.~\ref{fig:readers}f). Under the corner mask, T.T.N. keeps 95.3 $\pm$ 2.3 [90.4, 98.9] of the correct answers, L.A. keeps all of them, and these systems keep 94.5 to 95.2. Only T.T.N. read the displays with the matched mask. Under it, T.T.N. keeps 90.7 $\pm$ 3.2 [84.1, 96.5] of the correct answers, and these systems keep 89.2 to 90.4 (Fig.~\ref{fig:readers}e). No difference in IS between a reader and these systems is significant ($p_{\mathrm{FDR}} \geq 0.192$).

The difficulty-stratified set contains 80 MS-CXR cases in which the finding is present. We selected its cases by the answers of four multimodal systems (Supplementary Table~\ref{stab:reader_submitted}). S.Z. scores 81.3 $\pm$ 4.7 [71.2, 89.7], L.A. scores 58.8 $\pm$ 6.2 [46.7, 70.1], and T.T.N. scores 73.8 $\pm$ 5.6 [62.6, 84.2] (Fig.~\ref{fig:readers}g). Their majority answer scores 76.3 $\pm$ 5.3 [65.4, 85.7]. S.Z. scores 87.1 $\pm$ 6.2 [74.2, 96.8] on the 31 cases of this set with a box rated by S.Z. before the reading and 77.6 $\pm$ 6.1 [65.2, 89.1] on the other 49. L.A., who rated no box, scores 67.7 $\pm$ 8.2 [51.6, 83.9] and 53.1 $\pm$ 7.5 [38.0, 67.3] on the same two groups. The readers agree with a Fleiss' $\kappa$ of 0.268 ($p < 0.001$). Every correct answer on this set is Yes. So accuracy equals sensitivity. LLaVA-Med-7B scores 100.0, the vision-only reference scores 93.8, and the text-only control scores 83.8. The control differs from the majority answer by $+7.5 \pm 7.1$ [$-5.9$, 21.1] ($p_{\mathrm{FDR}} = 0.316$). The two one-sided tests do not establish equivalence between the control and the majority answer (90\% interval [$-3.6$, 18.8]). S.Z. and T.T.N. also rated a sample of 120 MS-CXR boxes, 15 per finding. S.Z. rated 58 of them as accurate, and T.T.N. rated 47 (Fig.~\ref{fig:readers}h; Supplementary Table~\ref{stab:box_validity}). Their Cohen's $\kappa$ on the four rating levels is 0.533 ($p < 0.001$). Both raters rated 40 of the boxes as accurate. Every rating of a cardiomegaly box is accurate, whereas at most one edema box per rater and sample is rated accurate. L.A. and T.T.N. also classified 22 cases, 21 of them wrong answers of Gemma-4-26B and the vision-only reference, by the likely cause of the answer. They assigned the same category to seven of the 22 cases (Cohen's $\kappa$ = 0.078, $p = 0.342$; Supplementary Table~\ref{stab:taxonomy}).

\section*{Discussion}

Benchmark accuracy and image use are separable. On chest radiograph questions, each occurs without the other. LLaVA-Med-7B receives the image and answers Yes whatever image it is shown. The text-only control, which receives none, outscores two multimodal systems on the pooled question set and scores 91.8 on the block in which every finding is present. The systems that use the image keep about half of their correct answers when the radiograph is swapped for another patient's radiograph with the opposite label. They change 6.3 to 33.5 of them when the region marked by a radiologist as the evidence is occluded. These results do not imply that medical VLMs are inaccurate. They imply that benchmark accuracy and image use have to be measured separately.

The control answers Yes to 92.6 of the finding-presence questions. With this prior, it scores 91.8 where every finding is present, and its balanced accuracy is at chance on the MIMIC-CXR block. On CheXpert, where this prior does not match the label distribution, the control scores below the constant No answer. Its behavior is consistent with shortcut learning reported across machine learning \cite{geirhos2020shortcut,lapuschkin2019cleverhans}. Visual question answering models also rely on answer priors when a benchmark permits it \cite{goyal2017vqa}. In a recent study of image reliance in medicine, replacing the image with a blank placeholder lowered the accuracy of GPT-4o by 27.9\% and that of three other models by 2.4\% to 8.5\% \cite{felizzi2025grounded}. In our removal conditions, every answer of the multimodal image users to a finding-presence question was No when they were shown noise or a photograph. Their accuracy then equaled the share of answered questions whose correct answer is No (Supplementary Fig.~\ref{sfig:swaps}f). Accuracy under removal therefore reflects the default answer of a system and the label distribution of the question set, and not how the system uses a radiograph. Where the image is necessary, every image user exceeds the control. On the report sentences, however, the balanced accuracy of Gemma-4-26B and Qwen3-VL-32B is not significantly different from that of the control. A high accuracy on a pooled benchmark can reflect how well a model's priors fit the label distribution as much as how well it uses the radiograph.

Unlike the removal of the image, the opposite-label swap still presents a radiograph, which has the opposite label for the queried finding. It is the primary test of image use in this audit. A same-label swap changes the identity of the radiograph and keeps the label. An answer that is unchanged under it may therefore still depend on the image, because the swapped radiograph has the same label. Under the opposite-label swap, an answer that changes with the label depends on image content that differs with the label. Because the replacement radiographs are matched on this label only, that content can be the finding itself or the findings, devices, and acquisition features that accompany it. SSP distinguishes a system whose answers follow the label from a system whose answers change under any swap. It is the Youden index of the answers to the replacement radiographs, which include one radiograph of each label for every question. A system that answers from the question alone therefore has an SSP of 0 on any benchmark, including a benchmark whose questions are not balanced by design. Because it needs no boxes, it applies to any dataset with image-level labels for the queried findings. Removing or swapping the supplied evidence has also been used to test whether a trained verifier of radiology claims depends on it \cite{tayebi2026evidence}. The occlusions add a test of localization. With the target mask, we test the dependence on a radiologist-marked region directly. Compared with the corner placement, the matched placement of the irrelevant mask reduces GSP by 1.9 to 9.7. GSP remains positive for every image user under both placements, with every 95\% interval above zero.

We modify only the input and measure only the answer, whereas mechanistic interpretability studies how an answer is formed inside the model. Removing the image tokens of an object in LLaVA lowers the accuracy of identifying the object by more than 70\% \cite{neo2025interpreting}. In LLaVA-1.5 and InternVL-3.5, token ablation, attention knockout, and causal mediation analysis show that image tokens aligned with an object define the extent of the object in the predicted box, and that a small set of attention heads mediates both localization and classification \cite{schaumloeffel2026localization}. If the image users localize a finding in the same way, the target mask removes the tokens that cover the finding. That mechanism would explain the correct answers changed by the mask. On visually realistic counterfactuals that contradict a world-knowledge prior, such as a blue strawberry, the predictions of multimodal models first follow the prior and shift toward the visual evidence in middle-to-late layers \cite{golovanevsky2025pixels}. The opposite-label swap poses the same conflict on radiographs, where the base rate of a finding takes the place of the world-knowledge prior. Under this swap, the question and therefore the prior stay fixed. An answer that follows the label of the swapped radiograph therefore depends on the image. The NOTICE pipeline corrupts images with semantic minimal pairs, which differ in one semantic element, in place of Gaussian noise. Its causal mediation analysis identified cross-attention heads that segment the image and suppress irrelevant objects \cite{golovanevsky2024notice}. Our swaps pair radiographs of different patients and are therefore not minimal pairs. Radiographs edited to differ only in the queried finding \cite{perezgarcia2024radedit} would allow the swap and a mediation analysis on the same inputs. The no-image condition is the input-level counterpart of removing the image tokens. Applying these tools to the multimodal image users is a natural next step.

The target mask changed fewer correct answers on AP radiographs for every image user, significantly for Gemma-4-26B. AP radiographs are mostly portable acquisitions of more acutely ill patients \cite{asrani2011urgent}. Because the views also differ in their mix of findings and the comparison is not adjusted for it, the difference does not show that image use is lower in sicker patients. For every image user, the mean confidence on grounded correct answers was lower than on ungrounded correct answers. On the MIMIC question set, the affirmative probability detected the label with an AUROC of at most 74.8. The expected calibration error was 0.139 to 0.505. A high confidence therefore does not indicate that a correct answer depends on the marked region. Monitoring the chain-of-thought captures only part of a model's reliance on each modality \cite{sanchezvillegas2026reasoning}. Accuracy and confidence together are an insufficient basis for a deployment claim \cite{wind2026safescale}. A test of image use can be reported beside them \cite{varoquaux2022machine,moor2023generalist}.

On the balanced set, both radiologists were more accurate than every system. Equivalence with either of them within the margin of 10\% was established for no system. The image users answered Yes on 40.0 to 49.0 of the finding-absent cases and were less specific than both readers. The grounding rates of the multimodal image users were between those of the readers, whose rates differed by a factor of almost three. Because the readers were asked to judge from the rest of the image when a rectangle covered the region needed for the answer, a reader's grounding rate also measures how much evidence remains outside the box. The grounding rate of each reader was higher on the positives with a box rated accurate by T.T.N. than on all positives (62.2 and 38.4 for T.T.N., 24.4 and 13.3 for L.A.). The multimodal image users changed 18.9 to 35.9 of their correct answers on these positives and 20.5 to 27.7 on all positives. Under the matched mask, T.T.N. kept 90.7 of the correct answers, these systems kept 89.2 to 90.4, and no paired difference was significant. On the difficulty-stratified set, in which every correct answer is Yes, the system that answered Yes to every question scored above every reader. A comparison of accuracy with radiologists therefore needs finding-absent cases, because on finding-present cases alone, a system that always answers Yes can score higher than a radiologist.

Several limitations qualify these conclusions. First, CGR measures the effect of occluding one marked rectangle. Evidence outside the rectangle lowers it, whether that evidence comes from global image features, from a diffuse or bilateral finding, or from the other boxes of a finding marked with several. The occlusion itself can also raise it, because a black rectangle is an unusual image feature. GSP subtracts this effect as measured with the irrelevant masks. For a bilateral finding, the mirrored mask can cover evidence of the finding. Two radiologists rated 58 and 47 of 120 marked boxes as covering the primary location of the finding (Cohen's $\kappa$ = 0.580). The per-finding CGR is least reliable for edema, since each radiologist rated at most one edema box per sample as accurate. We therefore define the categories from the swaps alone. Masks merged over every box of a finding, attribution methods that target global cues \cite{kim2018tcav}, and reference regions drawn by several radiologists would measure localization more completely. Second, every intervention takes the input off the training distribution. Some answer changes may therefore reflect sensitivity to an unfamiliar image and not the loss of evidence. Image artifacts alone lower the accuracy of VLMs on chest radiographs \cite{cheng2025artefacts}. For the occlusions, we estimate this sensitivity with the matched placement of the irrelevant mask. Under the noise and photograph conditions, a system that finds no evidence cannot be told apart from a system that recognizes a non-radiograph. Counterfactual radiographs that remove the finding and preserve the image statistics \cite{jeanneret2023adversarial,perezgarcia2024radedit} would be a more faithful intervention. Third, the human reference comes from three radiologists at two institutions. Two of them read the balanced set and agreed with a Cohen's $\kappa$ of 0.620 on its original displays. The masks appeared only on finding-present displays, and each positive was read under several conditions in separate sessions. The readers saw each radiograph at $512 \times 512$ pixels, whereas the systems saw it at $224 \times 224$ pixels. The positives were also more often AP radiographs than the negatives (83 against 64 of 100). A reading by more radiologists from more institutions would narrow the range of the human reference, which is widest for the grounding rate. Masked finding-absent displays at the resolution of the systems would make the readers' grounding rates directly comparable with those of the systems. Fourth, the audit poses yes-or-no finding-presence and sentence-verification questions, on which a change of answer under an intervention is unambiguous. The primary clinical use of these systems is report generation, in which image use may differ. The same interventions could be applied to a generated report. Fifth, we evaluate fixed model snapshots under one zero-shot, single-turn protocol. We ran Mistral-Small-4-119B as its 4-bit release and LLaVA-Med-7B as a conversion of its release to the Hugging Face format. Neither was compared with its reference implementation. Under the terse prompt, the wording alone changes how many questions a system answers. Few-shot prompting, retrieval-augmented frameworks \cite{tayebi2025radiorag,wind2025rar}, agentic tool use, or finetuning could also change how much a model uses the image. The category assignments therefore apply to the evaluated versions and protocol. A rerun of the swaps would update them. Sixth, the labels come from an automated labeler applied to the reports \cite{irvin2019chexpert} and not from pixel-level verification. Three questions name a finding more narrowly or more broadly than its label: the fracture label is asked as rib fracture, pleural other as pleural abnormality, and the normal studies as any acute abnormality. On the balanced set, T.T.N. and L.A. agreed with these labels with a Cohen's $\kappa$ of 0.720 and 0.640. For edema and pleural other, the test split contained too few certain negatives. We therefore drew 78 finding-absent cases from studies whose label for the finding is uncertain. Accuracy, balanced accuracy, and the swap metrics are also reported without these 78 cases. Because finding-presence questions have exploitable base rates, the absolute accuracies are relative to this benchmark. Labels verified by radiologists on the images would remove the dependence on the labeler. Seventh, the MedGemma models were trained on MIMIC-CXR radiographs and reports, the RAD-DINO encoder was pretrained on MIMIC-CXR and CheXpert images, and the heads of the vision-only reference were fit on images that include part of the question set. These systems may therefore have seen some question radiographs in training. A question set built from radiographs outside the training data of every system would remove this overlap. The panel contains eight open-weight systems and no closed system, because the confidence analysis needs token probabilities and every run needs a fixed decoding configuration. Because the swaps need only the answers, a closed system can be audited in the same way. The reasoning model has no first-token confidence, and the unstable system answers Yes so rarely that its localization metrics are computed on 25 cases.

The systems that use the radiograph change about half of their correct answers when it is swapped for a radiograph with the opposite label. A system that receives no radiograph scores above two multimodal systems. Confidence is not higher on the correct answers that depend on the image, and image use varies by model, finding, and view. Accuracy remains necessary, but it does not establish image use. With two image swaps and three occlusions, image dependence can be tested without access to model internals, at the cost of a few thousand additional inferences per system. The swaps need no bounding boxes and no access to model weights. They therefore apply to any dataset with image-level labels and to any system that can be queried. Reported beside accuracy before deployment, they can show whether a system's correct answers depend on the radiograph or on an answer prior.

\section*{Methods}
\subsection*{Ethics statement}

This study was conducted in accordance with the relevant national and international guidelines and regulations. It used previously collected chest radiographs, reports, and labels, each released for research in de-identified form, some openly and some under credentialed access. MIMIC-CXR \cite{johnson2019mimic} was approved by the institutional review board of Beth Israel Deaconess Medical Center, which waived individual patient consent. The same board approved MIMIC-IV \cite{johnson2023mimiciv}, the source of the age and sex fields, with the same waiver. MS-CXR \cite{boecking2022making} and ReXErr-v1 \cite{rajpurkar2024rexerr} annotate MIMIC-CXR studies and contain no new patient data. CheXpert \cite{irvin2019chexpert} was approved by the institutional review board of Stanford Hospital, which waived individual patient consent for the de-identified data. NIH ChestX-ray14 \cite{wang2017chestxray8}, the source of the example radiograph in Figs.~\ref{fig:overview} and~\ref{fig:localization}, was released publicly by the NIH Clinical Center after all personally identifiable information had been removed. No approving committee and no consent procedure are reported in its source publication. Each dataset was used under its license or data use agreement. No new patient data were generated and no participants were recruited. No further ethics approval or informed consent was therefore required for this secondary analysis. No re-identification was attempted, and no image or report is redistributed. Every model ran on institutional hardware, and no image, report, or record was passed to an external service.

\subsection*{Datasets}

The MIMIC question set is the primary question set of this study. It contains $n = 2{,}548$ yes-or-no questions on frontal chest radiographs from 1{,}608 patients, drawn from three corpora of the MIMIC-CXR family \cite{johnson2019mimic}: MS-CXR ($n = 452$), MIMIC-CXR ($n = 1{,}400$), and ReXErr-v1 ($n = 696$). An independent CheXpert question set ($n = 1{,}380$) from a second institution is used to test whether the audit transfers. Only frontal PA and AP radiographs were eligible, and cases lacking age or sex were excluded before sampling. Both sets were assembled once with a fixed seed of 42 and used unchanged for every system and condition. Supplementary Tables~\ref{stab:composition} and~\ref{stab:composition_chexpert} report their composition.

\paragraph{MS-CXR.} MS-CXR \cite{boecking2022making} provides radiologist-marked boxes that localize the visual evidence for eight findings on MIMIC-CXR radiographs. Every phrase-grounded annotation whose box measured at least $50 \times 50$ pixels at the $224 \times 224$ working resolution was retained, with at most 100 per finding drawn at random. Each box was scaled from the released image to the working resolution by the ratio of the widths and the ratio of the heights and rounded to the nearest pixel. Each box is a separate case. A finding marked with several boxes on one radiograph therefore contributes one case per box. The block contains $n = 452$ finding-present cases from 321 patients, for atelectasis ($n = 35$), cardiomegaly ($n = 100$), consolidation ($n = 76$), edema ($n = 43$), lung opacity ($n = 30$), pleural effusion ($n = 34$), pneumonia ($n = 97$), and pneumothorax ($n = 37$). The view is AP for $n = 366$ and PA for $n = 86$. The 452 cases form 364 radiograph-finding pairs, and 176 cases belong to the 88 pairs with two or more boxes.

\paragraph{MIMIC-CXR.} The MIMIC-CXR block was drawn from the held-out test split of the MIMIC-CXR label list used for sampling. This split contains about a fifth of the patients and differs from the split released with MIMIC-CXR. Its studies are labeled over the 14 findings of the CheXpert vocabulary by the labeler released with MIMIC-CXR, which marks each finding as present, absent, uncertain, or not mentioned. The radiographs of the MS-CXR block were removed, and for each finding and label, each patient contributes only the first of their radiographs with that label in the list. From this pool, 50 finding-present and 50 finding-absent cases were drawn for each of the 12 pathology findings and for support devices, together with 100 normal studies marked as having no finding. For edema and pleural other, fewer than 50 patients of the test split have the finding marked absent. The finding-absent pool of these findings was therefore extended with the studies whose label for the finding is uncertain. All 50 edema and 28 of the 50 pleural-other finding-absent cases came from that extension. The normal studies are asked whether any acute abnormality is present, and their correct answer is No. The block therefore contains $n = 1{,}400$ cases from 1{,}235 patients, $n = 650$ with the finding present and $n = 750$ with it absent. The view is AP for $n = 847$ and PA for $n = 553$.

\paragraph{ReXErr-v1.} ReXErr-v1 \cite{rajpurkar2024rexerr} provides MIMIC-CXR report sentences with one injected error each and the type of each error, together with error-free sentences. An item was eligible when it had an original sentence, a frontal radiograph, and an error flag that marks it as erroneous or as error-free. The eligibility rule excludes the added sentences of the repetition type and all but three of the added-device type. The measurement-change type belongs to neither group of error types and was not sampled. A budget of 560 image-dependent and 120 text-only error sentences was split evenly across the eligible error types of the test split, and 120 error-free sentences were added. In the test split, 476 of the 1{,}398 false-negation items have no error sentence. Of the 80 false-negation items sampled, the 27 without an error sentence were excluded. The block contains $n = 696$ sentence questions on 611 radiographs from 189 patients. They are $n = 456$ image-dependent errors, $n = 120$ text-only errors (60 typos and 60 homophones), and $n = 120$ error-free control sentences. The image-dependent errors are 80 changes of location, 80 changes of the name of a device, 80 changes of the position of a device, 80 changes of severity, 80 false predictions, 53 false negations, and three added devices. The correct answer is Yes for a control sentence and No for an error sentence, including a sentence whose only error is a typo or a homophone.

\paragraph{CheXpert.} The CheXpert question set was drawn from a held-out fifth of the CheXpert training set \cite{irvin2019chexpert}, whose labels come from the automatic labeler applied to the reports. The 39{,}323 images with an uncertain label for cardiomegaly, lung opacity, lung lesion, edema, or pneumonia had been removed before this fifth was held out. It uses the stratification of the MIMIC-CXR block, with 50 finding-present and 50 finding-absent cases per finding, 30 finding-absent cases for pleural other because no more existed, and 100 normal studies. The set contains $n = 1{,}380$ cases from 1{,}285 patients. The view is AP for $n = 761$ and PA for $n = 619$. No CheXpert report text is used. CheXpert has no phrase-grounding boxes.

\paragraph{Analysis sets and patient attributes.} The finding-presence cases of the MIMIC question set are its $n = 1{,}852$ MS-CXR and MIMIC-CXR questions, and every CheXpert question is a finding-presence question. The image-necessary subset contains the $n = 1{,}976$ questions whose correct answer depends on the radiograph. They are the 1{,}400 MIMIC-CXR cases, the 456 image-dependent ReXErr errors, and the 120 error-free control sentences. Of the 1{,}976 questions, 770 have a correct answer of Yes. Sex is the patient's sex as recorded in the hospital record, documented in MIMIC-IV \cite{johnson2023mimiciv} under the field name gender. It is female for $n = 1{,}201$ and male for $n = 1{,}347$ of the MIMIC questions, and female for $n = 582$ and male for $n = 798$ of the CheXpert questions. Age is the age at the study, grouped into three bands, under 50, 50 to under 70, and 70 and over. The median age is 68.2 years (interquartile range 56.8--78.5) on the MIMIC question set and 58.0 years (45.0--71.0) on the CheXpert question set.

\subsection*{Systems and inference}

Eight open-weight systems were evaluated: three general-purpose multimodal models (Gemma-4-26B \cite{google2026gemma4}, Qwen3-VL-32B \cite{bai2023qwenvl,qwen2025qwen3vl}, and Mistral-Small-4-119B), two medical multimodal models (MedGemma-1.5-4B \cite{googleresearch2026medgemma} and LLaVA-Med-7B \cite{llavamed}), two language models that receive the prompt and no image and serve as text-only controls (MedGemma-27B-text, the multimodal MedGemma 27B release called without an image \cite{googleresearch2026medgemma}, and the text-only DeepSeek-R1-7B \cite{deepseekai2025deepseekr1}), and one vision-only reference, a logistic-regression head over frozen RAD-DINO features \cite{perez2025raddino}. Parameter counts, modalities, developers, and release dates are listed in Supplementary Table~\ref{stab:registry}. The text-only controls answer from the question alone, and the vision-only reference answers from the image alone. Their scores serve as reference levels for the multimodal systems. The audit itself is about the multimodal systems.

Every language model was called through a chat-completion interface with greedy decoding at temperature 0. The generation budget was 10 new tokens for the non-reasoning models and 2{,}048 tokens for the reasoning model DeepSeek-R1-7B. Unless stated otherwise, every image was presented at $224 \times 224$ pixels. The radiographs were taken from copies of the released images resized to $224 \times 224$ and $512 \times 512$ pixels. Images sent to a language model were encoded as JPEG at quality 75, whereas the vision-only reference received the decoded image. Mistral-Small-4-119B was served from its NVFP4 release, which is quantized after training to a 4-bit floating-point format. Our calls did not request its optional reasoning mode. Under the default prompt, it answered with a single word. LLaVA-Med-7B was served from a conversion of its release to the LLaVA format of the Hugging Face Transformers library. In every call with a stored prompt length, a prompt with an image was longer than the same prompt without an image by the same number of tokens, which was 66 for Qwen3-VL-32B, 72 for Mistral-Small-4-119B, 258 for Gemma-4-26B, 259 for MedGemma-1.5-4B, and 578 for LLaVA-Med-7B. Greedy decoding through a batching server is deterministic up to the composition of the batch. Under the original image, each of the 88 MS-CXR radiograph-finding pairs with two or more boxes is asked two or more times with the same input. The answers to these repeated inputs agree for all 88 pairs for every system except Gemma-4-26B (87 pairs) and DeepSeek-R1-7B (84 pairs). Requests were issued 16 at a time. A call that failed at the transport level was retried with exponential backoff. A call that still failed was left without a record and was repeated when the run was restarted. Every case therefore has a recorded output under every condition. An empty output of a non-reasoning model was rerun once. The reasoning model returned no empty output.

For finding-presence cases, the default prompt was ``Is [display] present in this chest X-ray? Answer with a single word: Yes or No.'', where [display] is the human-readable finding name, for example pulmonary edema for the edema label and any acute abnormality for the normal studies (Supplementary Note~\ref{snote:prompts}). For ReXErr cases, the prompt presented the candidate sentence and asked ``Does the following sentence accurately describe the findings visible in this chest X-ray? Sentence: ``[sentence]'' Answer with a single word: Yes or No.'' DeepSeek-R1-7B additionally received a system message asking it to end its response with a single word on its own line.

A fixed parser mapped each output to Yes, No, or neither. It first decoded the byte-level space and line-break markers of the tokenizer. For the reasoning model, it then compared the last non-empty line, ignoring case and surrounding punctuation, with the affirmative words yes, yeah, correct, true, present, and positive and the negative words no, not, absent, negative, false, and incorrect. For every model, it next compared the first word of the output with the same words. An output of a non-reasoning model whose first word did not match was classified as an abstention when it contained one of a fixed list of phrases declining to answer (for example, ``I need to see the image'' or ``cannot determine''; Supplementary Note~\ref{snote:parser}). Otherwise, the parser looked in the first 60 characters of the output for a whole-word yes or no and accepted it when the other word did not appear. An output that was neither Yes nor No was classified as empty when it contained no text. For the reasoning model, any other such output was classified as an abstention when its last line contained one of the phrases and the output had not ended at the token budget, and as truncated otherwise. For every other model, it was classified as truncated when the server reported that it had ended at the token budget, and as unparsed otherwise. Under the default prompt, every answer of a non-reasoning system is the first word of its output. Under this prompt, no answered output of a non-reasoning system contains a phrase of the list. The outputs of DeepSeek-R1-7B contain no opening tag of the reasoning trace. Each of its 2{,}392 answered outputs under the original image contains the closing tag. Under the default prompt, every answer of this model comes from the last line of its output. The two exceptions are No answers found by the 60-character rule in a report sentence quoted at the start of the trace. Non-answers are excluded from the denominators of every rate. Their share is reported per system and class (Fig.~\ref{fig:accuracy}f) and per block and condition (Supplementary Fig.~\ref{sfig:nonanswers}a,b). As a sensitivity analysis, we also apply a permissive parser. Where the fixed parser returns neither Yes nor No, this parser accepts the first standalone Yes or No anywhere in the output.

The vision-only reference uses a frozen RAD-DINO encoder, a DINOv2 vision transformer with $14 \times 14$ patches and an input of $518 \times 518$ pixels, 37 patches per side. Each $224 \times 224$ image is preprocessed with the processor released with the encoder, which resizes the shortest edge to 518 pixels by bicubic interpolation, center-crops to $518 \times 518$, and normalizes with the released channel mean and standard deviation. The encoder runs in half precision. The 768-dimensional class-token embedding of the final block is standardized per dimension over the images used to fit the heads. One $L_2$-regularized logistic-regression head ($C = 1.0$, L-BFGS solver, at most 1{,}000 iterations, seed 42) is fit per finding on the training and validation frontal images of the source dataset, with the images that have the finding marked present as positives and those that have it marked absent as negatives. The source dataset is MIMIC-CXR for the MIMIC question set and CheXpert for the CheXpert question set. The MIMIC-CXR labels used here mark no image as edema absent. The negatives of the edema head are therefore the images whose edema label is uncertain. The head that answers the normal-study question is trained with the images of normal studies as negatives and the images with at least one of the 12 pathology findings present as positives. At inference, the head's probability is thresholded at 0.5. Because its CheXpert heads are fit on CheXpert images, the reference is in distribution on the CheXpert question set. No other system was trained or tuned on either question set. According to their model cards, the MedGemma models were trained on MIMIC-CXR radiographs and reports, and the RAD-DINO encoder was pretrained without labels on 368{,}960 MIMIC-CXR and 223{,}648 CheXpert images. The MIMIC-CXR heads were fit on the training and validation images of the current version of the MIMIC-CXR label list. These images include the radiographs of 252 of the 452 MS-CXR cases and of 289 of the 1{,}400 MIMIC-CXR cases. They also include the replacement radiographs of 1{,}502 of the 1{,}852 finding-presence cases under the same-label swap and of 1{,}513 under the opposite-label swap. The radiographs of the 289 MIMIC-CXR cases belong to the test split of the version used for sampling. On the MS-CXR block, the accuracy of the reference is 96.4 on the cases whose radiograph is among these images and 93.5 on the others. The multimodal image users differ by 9.0 to 11.9 between the same two groups, although their training does not depend on this split. On the MIMIC-CXR block, its balanced accuracy is 61.9 on the cases whose radiograph is among these images and 66.9 on the others. The reference ignores the prompt and answers finding-presence questions only. So its accuracy is computed on the 1{,}852 finding-presence questions of the MIMIC question set.

\subsection*{Image swaps, image removal, and behavioral categories}

Each condition changes only the image and keeps the question and the system fixed. The replacement radiographs of each swap were drawn with seed 42 before the systems were run under that swap, and every system saw the same replacement radiograph for a given case. The text-only controls receive no image under any condition. They were called again under the same-label swap, the target mask, and the corner mask. Under every other condition, their answer to the original call is used, since their input does not change. The same-label swap replaces the radiograph by a frontal radiograph of a different patient with the same label for the queried finding, drawn from the MIMIC-CXR radiographs of every split or, for CheXpert, from the CheXpert radiographs of every split. A pneumothorax-present case therefore swaps to another patient's pneumothorax-present radiograph, and a pneumothorax-absent case swaps to another patient's pneumothorax-absent radiograph. A normal study swaps to another normal study. The replacement radiographs are matched on the label of the queried finding only. The patient, the view, the other findings, and the devices can differ. As a sensitivity analysis, the swap metrics and the categories are also computed on the finding-presence questions whose two replacement radiographs have the same view as the original radiograph. Every finding-absent radiograph of a swap has the finding marked absent, except for edema on MIMIC-CXR, where the replacement radiographs have an uncertain edema label. For a ReXErr case, the same-label swap uses the alphabetically first finding marked present in its study, among the 12 pathology findings and support devices. The replacement radiograph was drawn among those with that finding present for an error sentence and absent for an error-free sentence. A ReXErr case whose study has no finding marked present swaps to any different-patient frontal radiograph. The opposite-label swap replaces the radiograph by a frontal radiograph of a different patient with the opposite label for the queried finding. A finding-present case swaps to a finding-absent radiograph, a finding-absent case swaps to a finding-present radiograph, and a normal study swaps to a radiograph with at least one of the 12 pathology findings present. The radiographs of the MS-CXR block are excluded from the pool of the opposite-label swap. The same-label swap is applied to every question. The opposite-label swap is applied to every finding-presence question, because a ReXErr sentence has no opposite label.

Let $a^o_j$ be a system's parsed answer to case $j$ under the original image, $y_j$ the label, and $a^c_j$ its answer under condition $c$. Every metric is computed on the cases answered under each of its conditions. The swap metrics are computed on the finding-presence cases of each question set. With $\mathcal{C} = \{j : a^o_j = y_j\}$ the correct-on-original cases, UAR under the same-label swap $s$ and OFR under the opposite-label swap $\bar{s}$ are
\begin{equation}\label{eq:swaps}
\mathrm{UAR} = \frac{1}{|\mathcal{C}|}\sum_{j \in \mathcal{C}} \mathbf{1}\{a^{s}_j = a^o_j\}, \qquad \mathrm{OFR} = \frac{1}{|\mathcal{C}|}\sum_{j \in \mathcal{C}} \mathbf{1}\{a^{\bar{s}}_j \neq a^o_j\},
\end{equation}
and SSP is $\mathrm{SSP} = \mathrm{OFR} - (1 - \mathrm{UAR})$, the excess of answer changes under a label-changing swap over answer changes under a label-preserving swap. SSP is computed as a paired difference on the cases answered under the original image and under both swaps. Because UAR and OFR are computed on separate sets of answered cases, SSP can differ slightly from the value obtained by combining the reported UAR and OFR. Because the question is binary and the swapped radiograph has the opposite label, with an uncertain edema label counted as absent, a changed answer under the opposite-label swap is the correct answer for the swapped radiograph. OFR is therefore also the counterfactual accuracy on those cases. In the same way, an unchanged answer under the same-label swap is correct for its replacement radiograph. Because each case has one replacement radiograph with each label, SSP equals the sensitivity plus the specificity minus 1 of the answers to the replacement radiographs. This Youden index is taken on a set that is balanced for every question. It is therefore 0 for any system whose answer is the same for every image. Without the conditioning on correctness, we also report two rates for each swap: the change rate, the share of all answered cases whose answer changes, and the swapped accuracy, the accuracy of the answer given under the swap against the label of the swapped radiograph.

The behavioral categories are defined from the swap metrics on the finding-presence cases of the MIMIC question set. A system uses the image when its SSP (Eq.~\ref{eq:swaps}) is positive with a bootstrap 95\% interval that excludes zero. It ignores the image when its OFR is 0 and its UAR is 100, each on at least 100 informative cases. It is unstable otherwise, when its answers change under swaps without following the label. The rule is deterministic, assigns every system to one category, and was fixed before the opposite-label swap was run.

The image-removal conditions replace the radiograph with no image, with Gaussian noise, or with a natural photograph, on every question of the MIMIC question set. Under no image, the prompt is sent without an image. The non-reasoning models have a budget of 64 tokens in this condition. Their outputs were first generated with 10 tokens, and every output that reached 10 tokens without an answer was generated again with 64 (2{,}103 outputs of Gemma-4-26B and 101 of LLaVA-Med-7B). Under Gaussian noise, every case is shown one fixed $224 \times 224$ grayscale image whose pixels were drawn independently from a normal distribution with the mean and SD of the question-set radiographs (120.9 and 77.3 on the 8-bit scale). The pixel values were clipped to the 8-bit range, and the image was generated once with seed 0. Under the natural-image condition, every case is shown one fixed public-domain photograph of a cat \cite{asabae2019cat}, stored at $1024 \times 768$ pixels and resampled to $224 \times 224$ by bilinear interpolation. The vision-only reference, which has no input without an image, has no no-image condition. For each condition, the accuracy, the Yes-rate, and the prior agreement are reported on the finding-presence cases. The Yes-rate is the share of answered questions answered Yes, and the prior agreement is the share of correct-on-original answers reproduced under the condition.

\subsection*{Accuracy and non-answers}

The accuracy family comprises accuracy, sensitivity (the share of Yes questions answered Yes), specificity (the share of No questions answered No), balanced accuracy (their mean), F1 of the Yes answer, and the Yes-rate. Each is reported on the pooled question set, on the finding-presence cases, on the image-necessary subset, per source, and per ReXErr class, beside two constant references, the answer Yes to every question and the answer No to every question. On the MIMIC question set, every system is compared with each text-only control in accuracy and balanced accuracy on the pooled questions, the finding-presence cases, the image-necessary subset, and each source. The same comparison is made on the CheXpert question set. Balanced accuracy is undefined on the MS-CXR block, in which every finding is present. The primary comparison between systems is balanced accuracy on the image-necessary subset, on which every pair of systems is compared.

As sensitivity analyses for the non-answers, we score every abstention, truncation, empty, and unparsed output as incorrect and recompute accuracy on the pooled question set, the image-necessary subset, and the ReXErr block. We also rescore every output with the permissive parser and recompute the answered share, the accuracy, UAR, OFR, and CGR. Finally, we rerun every truncated output of DeepSeek-R1-7B once with a budget of 8{,}192 tokens and recompute its answered share and accuracy on the pooled question set.

\subsection*{Occlusion masks and localization}

The masks are defined on the 452 MS-CXR cases, which are the only cases with boxes. Each placement was computed once per case before the systems were run under it. The target mask sets the pixels of the radiologist box to black. Two irrelevant masks black out a rectangle of the same width and height elsewhere. Every mask is drawn with both edges of its rectangle included. The blacked-out area is therefore one pixel wider and one pixel taller than the box. Under the corner placement, the rectangle lies in the image corner at which its center is farthest from the center of the box. In 11 cases with a box wider than 100 pixels, this rectangle covers 0.7 to 25.3 of the area of the target box. The matched placement is at the position of the target box mirrored across the vertical midline of the image. Where the mirrored rectangle overlaps the target box, it is moved vertically along the same side of the image to the nearest position with no overlap. Where no such position exists, it is moved horizontally to the position immediately beside the target box. Because the drawn rectangles include their edges, a mask moved next to the target box can share one pixel row or column with it. A box too large for either move is masked at the corner instead. Of the 452 cases, 322 are masked at the mirrored position, 119 at a shifted position, and 11 at the corner. The corner rectangle overlaps the target box in these 11 cases only. Fig.~\ref{fig:localization}e draws the placements on a public NIH ChestX-ray14 radiograph \cite{wang2017chestxray8} with that release's annotation, since no image of the question set may be reproduced.

CGR is the share of correct affirmative answers that change under the target mask $t$, and IS is the share of correct-on-original answers unchanged under an irrelevant mask $i$,
\begin{equation}\label{eq:masks}
\mathrm{CGR} = \frac{\sum_{j \in \mathcal{C}} \mathbf{1}\{a^{t}_j \neq a^o_j\}}{|\mathcal{C}|}, \qquad \mathrm{IS} = \frac{\sum_{j \in \mathcal{C}} \mathbf{1}\{a^{i}_j = a^o_j\}}{|\mathcal{C}|},
\end{equation}
and GSP is $\mathrm{GSP} = \mathrm{CGR} - (1 - \mathrm{IS})$, which is positive when answers change more under the occlusion of the marked region than under the occlusion of a region of the same size elsewhere. GSP is paired on the cases answered under the original image and under both masks. IS and GSP are reported under the corner and under the matched placement, and IS also by placement class. Without the conditioning on correctness, we also report the share of all answered cases whose answer changes under the target mask and the share unchanged under each irrelevant mask. CGR and IS are also computed on the 290 MS-CXR cases answered correctly by every system that uses the image.

CGR, IS, and OFR are also reported per finding. On the MIMIC question set, CGR, UAR, and OFR are compared between the sexes, between PA and AP radiographs, and across the age bands within each system. In these subgroup analyses, UAR and accuracy are computed on every question of the MIMIC question set, the ReXErr questions included. Accuracy, CGR, UAR, and OFR are also reported per sex on both question sets. The comparisons of UAR and OFR by sex, view, and age band are repeated on CheXpert. The sex analyses are reported whatever their outcome.

\subsection*{Transfer and robustness analyses}

Because the category rule does not use boxes, we apply it unchanged to the CheXpert question set. Transfer is the agreement between the category assignments on the question sets. The rank agreement of balanced accuracy on the finding-presence questions between the question sets is computed across the systems. For OFR and UAR, whose values are 0 and 100 for the ignores-image systems on both question sets, the order of the other systems is compared.

For the prompt-sensitivity analysis, we added two phrasings of the finding-presence question to the default phrasing (Supplementary Note~\ref{snote:prompts}). The terse variant, ``Is [display] present? Yes or No.'', drops the chest X-ray framing and the single-word instruction. The non-reasoning models answer it with a generation budget of 128 tokens. The radiologist-framed variant, ``You are a radiologist reviewing a chest X-ray. Is [display] present? Answer with a single word: Yes or No.'', prepends a role. The multimodal language systems answered both variants on all 452 MS-CXR cases under the original image, the target mask, and the corner mask, and the text-only controls under the original image. Every language system also answered both variants on a 200-case subsample of the MIMIC-CXR block under the original image, drawn with seed 42 and stratified by finding and label state. For the resolution analysis, the language systems answered 100 MS-CXR cases, drawn at random with seed 42, at $512 \times 512$ pixels under the original image and the target mask, with the box coordinates scaled by 512/224 and rounded down. CGR is compared between the resolutions by the paired difference per system.

In the category sensitivity analysis, the minimum informative count of the swap rule was varied over 50, 100, and 200 cases. We also evaluate an alternative rule built on the localization metrics of Eq.~\ref{eq:masks}. It assigns the ignores-image category at a CGR of 0 with a UAR and an IS of 100, the unstable category at an IS below 70, and the uses-image category at a CGR interval excluding zero and an IS of at least 90. Any other system, including a system with an IS between 70 and 90, is left unassigned. The thresholds of this rule are applied to rates rounded to one decimal. In a threshold sensitivity, its two IS thresholds are set to a common value from 50 to 90 in steps of 10, under each placement of the irrelevant mask and also with the unconditioned IS (Supplementary Fig.~\ref{sfig:robustness}h). We also recomputed accuracy, balanced accuracy, and the swap metrics without the 78 finding-absent cases with an uncertain label (Supplementary Fig.~\ref{sfig:robustness}g). This analysis keeps the edema-present cases, whose opposite-label replacements have an uncertain edema label.

\subsection*{Confidence and calibration}

The affirmative probability is the first-token probability of Yes renormalized over the affirmative and negative token sets, from the five most probable first tokens returned by the serving system,
\begin{equation}\label{eq:confidence}
P(\text{Yes} \mid \text{prompt}, \text{image}) = \frac{\sum_{t \in \mathcal{T}_{\text{yes}}} p(t)}{\sum_{t \in \mathcal{T}_{\text{yes}}} p(t) + \sum_{t \in \mathcal{T}_{\text{no}}} p(t)},
\end{equation}
where $p(t)$ is the first-token probability and $\mathcal{T}_{\text{yes}}$ and $\mathcal{T}_{\text{no}}$ are the affirmative and negative token sets (\{Yes, yes, YES, \texttt{\textvisiblespace}Yes, \texttt{\textvisiblespace}yes, \texttt{\textvisiblespace}YES\} and the analogous No set). The confidence of an answer is $P(\text{Yes})$ of Eq.~\ref{eq:confidence} for a Yes answer and $1 - P(\text{Yes})$ for a No answer. Every answer used in the confidence analyses is the first word of its output, and its confidence is at least 49.9. For the vision-only reference, $P(\text{Yes})$ is the sigmoid output of the per-finding head. DeepSeek-R1-7B's first token belongs to its reasoning trace. So it has no confidence and is excluded from every confidence and calibration analysis. For every other language system, each answered question had a Yes or a No token among its five most probable first tokens. A system would also be excluded if at least 90\% of its confidence values were 0 or 1. No system in the panel met this condition.

For the systems with a confidence, the correct MS-CXR answers are split into grounded correct answers, which change under the target mask, and ungrounded correct answers, which stay unchanged. Every incorrect answer on the pooled question set forms a third regime. The mean confidence, its SD, and the count are reported per regime and compared descriptively. Discrimination and calibration are summarized on the MIMIC and on the CheXpert question set by the AUROC of $P(\text{Yes})$ as a detector of the Yes label, the Brier score \cite{brier1950verification}, and the expected calibration error \cite{naeini2015obtaining,guo2017calibration}. On the MIMIC question set, the vision-only reference is scored on its 1{,}852 finding-presence questions. The AUROC is computed by trapezoidal integration of the empirical curve, with its bootstrap interval. The AUROC is also computed within each finding of the MIMIC-CXR block, over the pairs of a finding-present and a finding-absent case of the same finding. The Brier score is the mean squared difference between the probability of the correct answer and 1. The expected calibration error compares the confidence with the accuracy in 10 equal-width bins of confidence.

\subsection*{Reader study}

Three board-certified radiologists took part: S.Z. and L.A., with 6 and 10 years of experience, and T.T.N., with 8 years of experience in diagnostic and interventional radiology. L.A. and T.T.N. read the balanced set, and all three read the difficulty-stratified set. Each display showed one radiograph at $512 \times 512$ pixels with the finding-presence question of the systems, for example ``Is pneumonia present in this chest X-ray? Answer with a single word: Yes or No.''. On the difficulty-stratified set, the question was shown without its answer-format sentence. Each reader entered Yes or No, a confidence from 1 to 5, and an optional comment. The reading sheet of each packet listed only the display identifier, the session, and the question. The readers therefore saw no model output, no label, and no answer of another reader while reading. They were told that some images have a black rectangle, and they were told nothing else about a display. They were asked to answer from what is visible, to give their best judgment from the rest of the image when a rectangle covered the region needed for the answer, to answer every display, and to treat every display as independent.

The balanced set consists of 200 cases. The 100 positives are MS-CXR finding-present cases with a box, 12 or 13 per finding, drawn at random within finding with seed 42 from the 372 MS-CXR cases outside the difficulty-stratified set, with no selection by model behavior. The 100 negatives are MIMIC-CXR finding-absent cases of the same eight findings, 12 or 13 per finding, drawn at random within finding with seed 43 from the question set, so that every system's answers on them exist. The set contains 83 AP and 17 PA positives, 64 AP and 36 PA negatives, 87 women and 113 men, and 187 patients. Thirteen negatives, all edema, are finding absent by the uncertain-label rule. The accuracy of each reader is also reported without them. Every positive was shown under four conditions, the original image, the target mask, the corner mask, and the matched mask. Every negative was shown under the original image only. T.T.N. read all 500 displays in four sessions of 100 positives and 25 negatives each. The displays of one positive fell in different sessions, with the condition of each case rotated across the sessions in a Latin square and the order within a session random. L.A. read the 400 displays without the matched mask in three sessions of 100 positives and 33 or 34 negatives, with the same rotation.

The difficulty-stratified set consists of 80 MS-CXR finding-present cases, at most 12 per finding, drawn with seed 20260526. The cases were selected by the answers of four multimodal systems on the original image, three of them in the panel (Gemma-4-26B, Qwen3-VL-32B, and MedGemma-1.5-4B). The set contains 40 cases answered correctly by at least three of the four, 28 answered correctly by at most one system, and 12 in between. Of its radiographs, 61 are AP and 19 are PA. Every correct answer on it is Yes. Accuracy on it therefore equals sensitivity. Because the selection used model correctness, the set is enriched for cases answered incorrectly by the selecting systems. The three readers read its 240 displays, the 80 cases under the original image, the target mask, and the corner mask, in three sessions. It is reported as a secondary analysis.

S.Z. and T.T.N. rated a sample of 120 MS-CXR boxes, 15 per finding, drawn at random with seed 20260526, and T.T.N. also rated the 100 boxes of the balanced positives. Each box was drawn in red on the radiograph at $512 \times 512$ pixels with the name of the finding and rated accurate (the rectangle covers the primary location of the finding), partial (it covers part of the finding but misses important regions or includes substantial unrelated anatomy), inaccurate (it does not cover the finding, or the finding is not present where indicated), or cannot tell. T.T.N. rated both samples after the last reading session. S.Z. was asked to rate the 120 boxes before reading the difficulty-stratified set, and 31 of its 80 cases are in the sample. The failure taxonomy contains 22 finding-presence cases of the MIMIC-CXR block, drawn with seed 20260526 from the most confident quarter of the wrong answers of Gemma-4-26B (14 cases) and of the vision-only reference (eight cases). Because the cases were drawn with an earlier fit of its heads, the vision-only reference analyzed here answers one of its eight cases correctly. L.A. and T.T.N. classified each case as an ambiguous case, poor image quality, a plausible image confounder, a clear model failure, or other. Each case was shown with its question, the correct answer, and the answer and confidence of the system, which was named only by a letter. Each rater added a 1-to-5 rating of how confidently a typical radiologist would answer the case.

Every analysis is computed per reader. On the difficulty-stratified set, which all three read, it is also computed for the majority answer of the three readers. On the balanced set, diagnostic performance is accuracy, sensitivity, specificity, and balanced accuracy on the 200 original displays. For each of these metrics, a system is compared with each reader by the paired difference, system minus reader, on the cases answered by both the system and the reader. Localization is measured by CGR and IS on the positives, with the same conditioning on correct original answers as for the systems. Each system is compared with each reader by the paired difference in CGR and IS. CGR is also computed on the positives with a box rated accurate by T.T.N., for the readers and for the systems on the same cases. On the difficulty-stratified set, every system is compared with the majority answer in accuracy. Agreement is computed between the readers on the original and on the masked displays, between each reader and the report-derived label, and between the raters of the boxes and of the taxonomy. Box validity is the share of boxes rated accurate.

\subsection*{Statistical analysis}

Every rate is a percentage with one decimal, rounded half up. Every per-system rate is the mean over its cases, reported with the SD and the 2.5th and 97.5th percentiles of its bootstrap distribution, from 1{,}000 resamples at seed 0 that draw patients with replacement \cite{efron1993introduction}. Balanced accuracy, F1, and AUROC are recomputed on each resample as functions of the resampled cases. Paired comparisons between two systems and between a system and a reader use the paired bootstrap over their shared answered cases under the conditions involved, resampling patients 1{,}000 times. Each reports the observed difference, the SD of the bootstrap distribution, its 2.5th and 97.5th percentiles, and the shared case count. The two-sided $p$-value is computed by the shift-and-reflect method. The bootstrap difference distribution is recentered at zero, and the $p$-value is the share of recentered draws whose magnitude is at least the magnitude of the observed difference, bounded below by 1 over the number of resamples (Supplementary Algorithm~\ref{alg:paired_bootstrap}). Every comparison between a system and a text-only control, every pairwise comparison on the image-necessary subset, and every accuracy and balanced-accuracy comparison between a system and a reader is accompanied by two one-sided tests of equivalence at a margin of 10\% \cite{schuirmann1987tost}, fixed before the balanced set was read. Equivalence is established when the 90\% percentile interval of the paired bootstrap difference lies within $\pm$10\%. A rate computed on fewer than 10 cases is reported and not interpreted. A paired comparison is computed only when at least 10 cases are shared. With 1{,}000 resamples, a $p$-value near 0.05 has a Monte Carlo standard error of about 0.007.

Within each family, $p$-values are corrected by the FDR at 5\% \cite{benjamini1995controlling}, and a comparison is called significant only when its corrected value is below 0.05. The families are the comparisons of every system with each text-only control, per question set, metric, scope, and control; the pairwise balanced-accuracy comparisons among all systems on the image-necessary subset; the subgroup tests within each system and question set; and the comparisons between the systems and a reader within each metric, reader, and case set. The $p$-values of the resolution differences, the rank correlations, and the agreement coefficients are not corrected and are reported descriptively.

Subgroup differences are tested by permutation, with 1{,}000 permutations at seed 0 \cite{good2005permutation} that reshuffle the group labels across cases. The interval of a subgroup difference comes from a bootstrap that resamples the cases of each group separately. Neither procedure keeps the cases of a patient together. The test statistic is the absolute difference in means for sex and for view, and the one-way $F$ statistic for the age bands. A band with fewer than two cases is dropped, and the $p$-value is bounded below by 1 over 1{,}001. Rank agreement is Spearman's $\rho$ across the systems, with its two-sided $p$-value from the $t$ approximation. Agreement coefficients are reported as their value with a one-sided permutation $p$-value against no agreement beyond chance, from 1{,}000 permutations at seed 0 in which the other raters' labels are permuted across the displays, and without an interval. We use Cohen's $\kappa$ \cite{cohen1960coefficient} for the agreement between two readers or raters on the Yes or No answer, the four-level box rating, and the taxonomy category, and between each reader and the report-derived label. We use quadratic-weighted $\kappa$ \cite{cohen1968weighted} for the 1-to-5 confidence ratings and Fleiss' $\kappa$ \cite{fleiss1971measuring} for the three readers. Percent agreement, box validity, and every reader accuracy are rates and are reported with the bootstrap convention above. The box validity of a single finding is reported with a Wilson interval \cite{wilson1927probable} instead, because of its small counts.

\section*{Data availability}
\sloppy

Every dataset in this study comes from an existing, publicly released source. No images or derived records are redistributed here. MIMIC-CXR~\cite{johnson2019mimic} is available from PhysioNet under credentialed access at \url{https://physionet.org/content/mimic-cxr-jpg/2.0.0/}. Access requires a signed data use agreement and completion of the required human-subjects training. The MS-CXR boxes~\cite{boecking2022making} (\url{https://physionet.org/content/ms-cxr/1.1.0/}) and the age and sex fields of MIMIC-IV~\cite{johnson2023mimiciv} (\url{https://physionet.org/content/mimiciv/3.1/}) are available under the same terms. ReXErr-v1~\cite{rajpurkar2024rexerr} is openly available from PhysioNet at \url{https://physionet.org/content/rexerr-v1/1.0.0/}. CheXpert~\cite{irvin2019chexpert} is available from the Stanford Machine Learning Group at \url{https://stanfordmlgroup.github.io/competitions/chexpert/} on request. NIH ChestX-ray14~\cite{wang2017chestxray8}, the source of the example radiograph in Figs.~\ref{fig:overview} and~\ref{fig:localization}, is openly available from the US National Institutes of Health Clinical Center at \url{https://nihcc.app.box.com/v/ChestXray-NIHCC}. The photograph of the image-removal condition is openly available on Wikimedia Commons~\cite{asabae2019cat}. The source data for Figs.~\ref{fig:categories} to~\ref{fig:readers} are available as Supplementary Data 1 and 2.

\section*{Code availability}
The analysis code is publicly available at \url{https://github.com/mahshadlotfinia/causal}. The repository contains the code for building the question sets, applying the image interventions, running and parsing the systems, fitting the vision-only reference, computing the metrics and statistical tests behind the tables and figures, and analyzing the returned reader sheets. The additional checks reported only in the text were computed from the stored outputs and records of the study. Its configuration and its question-set builder fix the seeds of the question sets and of the replacement radiographs, and its statistics module fixes the bootstrap and permutation seeds. It also contains the fixed noise image and the stored copy of the photograph of the image-removal conditions. It does not redistribute the model weights or the underlying datasets.

All systems were run on institutional infrastructure, without any cloud service or third-party application programming interface. Every evaluated system is an open-weight model. The language models were served from their Hugging Face releases through a chat-completion interface. LLaVA-Med-7B was served from a conversion of its release to the LLaVA format of the Hugging Face Transformers library. The vision-only reference used the frozen RAD-DINO encoder with one logistic-regression head per finding. All models were accessed and all experiments were run between May and September 2026. The URLs of the evaluated checkpoints are:

\begingroup
\footnotesize
\sloppy
\vspace{0.4em}
\textbf{General-purpose multimodal models:}
\begin{itemize}
    \item Gemma-4-26B: \url{https://huggingface.co/google/gemma-4-26B-A4B-it}
    \item Qwen3-VL-32B: \url{https://huggingface.co/Qwen/Qwen3-VL-32B-Instruct}
    \item Mistral-Small-4-119B: \url{https://huggingface.co/mistralai/Mistral-Small-4-119B-2603-NVFP4}
\end{itemize}
\textbf{Medical multimodal models:}
\begin{itemize}
    \item MedGemma-1.5-4B: \url{https://huggingface.co/google/medgemma-1.5-4b-it}
    \item LLaVA-Med-7B: \url{https://huggingface.co/microsoft/llava-med-v1.5-mistral-7b}
\end{itemize}
\textbf{Text-only controls:}
\begin{itemize}
    \item MedGemma-27B-text: \url{https://huggingface.co/google/medgemma-27b-it}
    \item DeepSeek-R1-7B: \url{https://huggingface.co/deepseek-ai/DeepSeek-R1-Distill-Qwen-7B}
\end{itemize}
\textbf{Vision-only reference:}
\begin{itemize}
    \item RAD-DINO: \url{https://huggingface.co/microsoft/rad-dino}
\end{itemize}
\endgroup
The model runs and the analyses used separate Python environments. Model inference and the vision-only reference ran under Python~3.11 with PyTorch~2.9 and transformers~5.0, together with huggingface-hub~1.3, tokenizers~0.22, accelerate~1.12, safetensors~0.7, scikit-learn~1.8, NumPy~1.26, pandas~3.0, and Pillow~12.2. The serving system was accessed through the OpenAI Python client~2.26 over httpx~0.28. The metrics, the statistical procedures, and the figures were produced under Python~3.13 with NumPy~2.1, SciPy~1.15, pandas~2.2, statsmodels~0.14, Matplotlib~3.10, Pillow~11.1, PyYAML~6.0, and tqdm~4.67. Model inference and head fitting ran on NVIDIA RTX PRO 6000 GPUs, and the other analyses ran on CPUs.

\section*{Acknowledgements}

STA is supported by the German Federal Ministry of Research, Technology and Space (ARISTOTLE, 01ZU2602B) and the Excellence Strategy of the German Federal Government, the L\"ander, and RWTH ERS (START\_526-26). DT is supported by the German Federal Ministry of Research, Technology and Space (TRANSFORM LIVER - 031L0312C, DECIPHER-M - 01KD2420B), DFG (515639690), and the European Union (Horizon Europe, ODELIA - GA 101057091, ERC Starting Grant SAGMA - GA 101222556).

\section*{Author contributions}

The formal analysis was conducted by ML, AM, and STA. The original draft was written by ML and STA and edited by STA. ML developed the code. The experiments were performed by ML. The statistical analyses were performed by ML and STA. SZ, LA, and TTN performed the reader studies. SZ, LA, TTN, and DT provided clinical expertise. ML, DT, AM, and STA provided technical expertise. The study was defined by STA. All authors read the manuscript and agreed to the submission of this paper.

\section*{Competing interests}

ML is employed by Generali Deutschland Services GmbH, Germany, and is on the editorial board of European Radiology Experimental. LA is on the trainee editorial board of Radiology: Artificial Intelligence. DT received honoraria for lectures by Bayer, GE, Roche, AstraZeneca, and Philips and holds shares in StratifAI GmbH, Germany, and in Synagen GmbH, Germany. AM is an associate editor at IEEE Transactions on Medical Imaging. STA is on the editorial board of Communications Medicine and of European Radiology Experimental, and on the trainee editorial board of Radiology: Artificial Intelligence. The other authors do not have any competing interests to disclose.

\bibliographystyle{splncs04}
\bibliography{bibliography}

\setcounter{table}{0}
\setcounter{figure}{0}
\setcounter{equation}{0}
\renewcommand{\tablename}{Supplementary Table}
\renewcommand{\figurename}{Supplementary Fig.}
\floatname{algorithm}{Supplementary Algorithm}
\renewcommand{\thealgorithm}{\arabic{algorithm}}
\renewcommand{\theequation}{S\arabic{equation}}
\newcounter{suppnote}

\section*{Supplementary information}

\begin{figure}[p]
\centering
\includegraphics[width=\textwidth]{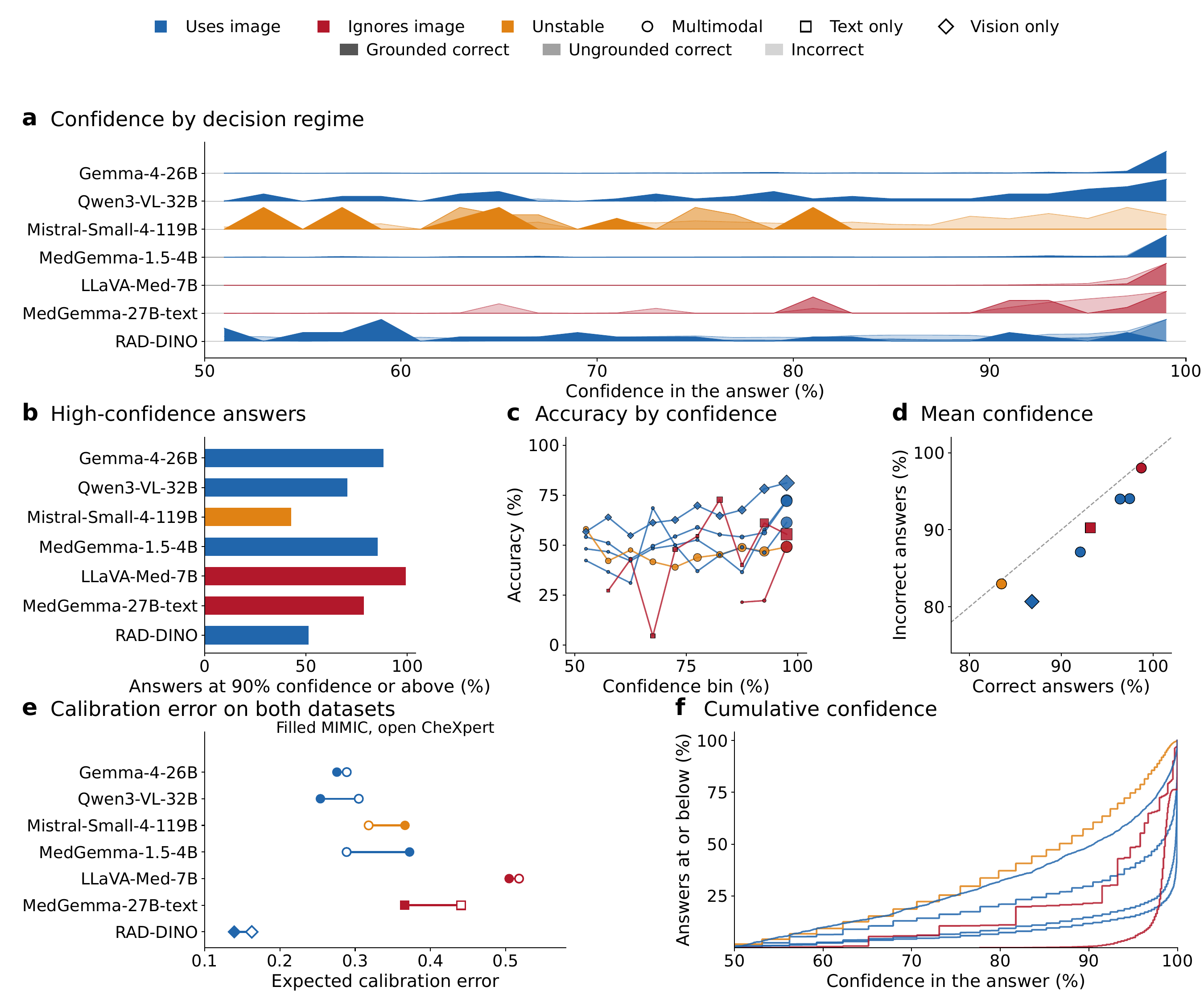}
\caption{Distribution of each system's confidence in its answer, on the MIMIC question set ($n$ = 2{,}548 questions). The confidence in a chosen answer is at least 49.9, and every confidence axis starts at 50. Fill color encodes the behavioral category (blue, uses image; red, ignores image; orange, unstable) and marker shape encodes modality (circle, multimodal; square, text only; diamond, vision only). No panel has an error bar. Panels \textbf{b} and \textbf{e} keep the order of \textbf{a}. \textbf{a}, Density of the confidence in each decision regime on 25 equal-width bins ($n$ = 10 to 1{,}361 answers per regime), each ridge scaled to a common height. The grounded and the ungrounded correct answers are the correct affirmative MS-CXR answers that change and that stay the same under the target mask. The incorrect answers are those of the pooled question set. The systems that change no answer have no grounded regime. \textbf{b}, Share of the answered pooled questions answered with at least 90\% confidence ($n$ = 1{,}852 to 2{,}548 per system). \textbf{c}, Accuracy within each 5\% bin of the confidence ($n$ = 1{,}852 to 2{,}548 per system), with the marker area proportional to the number of answers in the bin. A bin of fewer than five answers is omitted. \textbf{d}, Mean confidence on the correct against the incorrect answers ($n$ = 1{,}852 to 2{,}548), with the identity line. \textbf{e}, Expected calibration error on the MIMIC question set and on the CheXpert question set ($n$ = 1{,}852 to 2{,}548 and 1{,}355 to 1{,}380). \textbf{f}, Cumulative share of the answered questions at or below each confidence ($n$ = 1{,}852 to 2{,}548). DeepSeek-R1-7B has no confidence and is left out.}
\label{sfig:confidence_histograms}
\end{figure}

\begin{figure}[p]
\centering
\includegraphics[width=\textwidth]{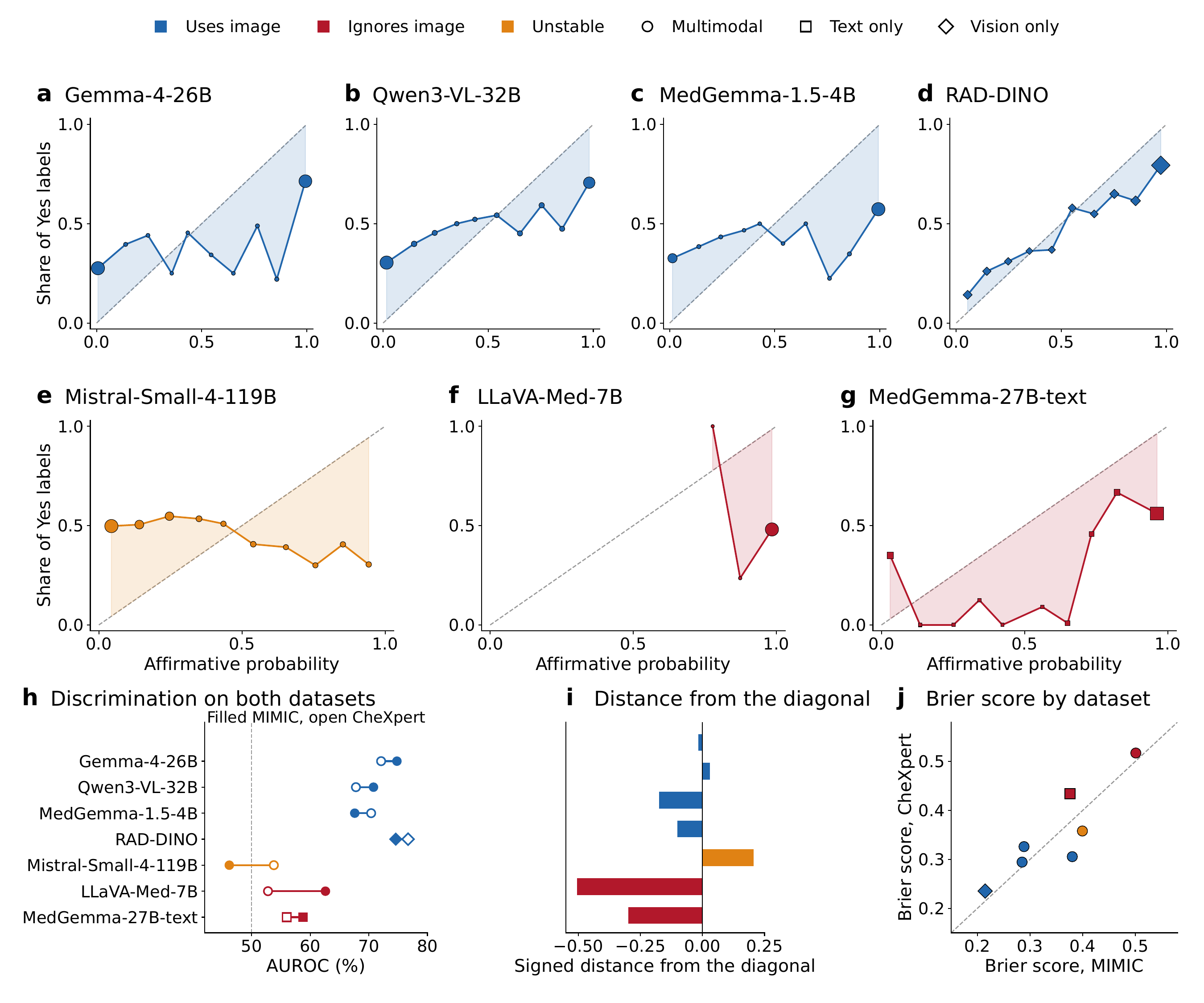}
\caption{Reliability of the affirmative probability on the MIMIC question set ($n$ = 2{,}548 questions), with the discrimination and the Brier score on both question sets. Fill color encodes the behavioral category (blue, uses image; red, ignores image; orange, unstable) and marker shape encodes modality (circle, multimodal; square, text only; diamond, vision only). No panel has an error bar. \textbf{a}--\textbf{g}, The answered pooled questions of each system ($n$ = 1{,}852 to 2{,}548), binned by the affirmative probability into 10 equal-width bins, with each bin's share of Yes labels against its mean probability. The marker area is proportional to the number of questions in the bin, the dashed diagonal marks perfect calibration, and the shaded area marks the distance from it. Panels \textbf{a} to \textbf{d} show the systems that change an answer under the image interventions, and panels \textbf{e} to \textbf{g} show the systems that change none. \textbf{h}, AUROC of the affirmative probability as a detector of the Yes label on each question set ($n$ = 1{,}852 to 2{,}548 on MIMIC and 1{,}355 to 1{,}380 on CheXpert), with the dashed line at chance. \textbf{i}, Mean signed distance of the reliability curve from the diagonal, weighted by the number of answers in each bin ($n$ = 1{,}852 to 2{,}548). A negative value means the affirmative probability is above the share of Yes labels. \textbf{j}, Brier score on one question set against the other, with the identity line ($n$ = 1{,}852 to 2{,}548 and 1{,}355 to 1{,}380). Fig.~\ref{fig:confidence}e draws all curves in one panel. DeepSeek-R1-7B begins its output with a reasoning trace. So it has no affirmative probability and is left out. AUROC, area under the receiver operating characteristic curve.}
\label{sfig:reliability}
\end{figure}

The supplementary information contains a glossary of every term and a summary of every finding (Supplementary Tables~\ref{stab:glossary} and~\ref{stab:findings}), the prompt texts and the parser's phrase list (Supplementary Notes~\ref{snote:prompts} and~\ref{snote:parser}), the composition of both question sets and the registry of the systems (Supplementary Tables~\ref{stab:composition}, \ref{stab:registry}, and~\ref{stab:composition_chexpert}), the swap and mask metrics (Supplementary Figs.~\ref{sfig:swaps} and~\ref{sfig:localization}; Supplementary Tables~\ref{stab:is_placement} and~\ref{stab:cgr_finding}), the accuracy family with the paired comparisons and the equivalence tests (Supplementary Figs.~\ref{sfig:accuracy} to~\ref{sfig:vs_deepseek}), the non-answer audit and the sensitivity analyses (Supplementary Figs.~\ref{sfig:nonanswers} and~\ref{sfig:robustness}), the sex-disaggregated metrics (Supplementary Tables~\ref{stab:sex} and~\ref{stab:sex_chexpert}), the CheXpert metrics (Supplementary Table~\ref{stab:chexpert}), the confidence and calibration values (Supplementary Figs.~\ref{sfig:confidence_histograms} and~\ref{sfig:reliability}; Supplementary Table~\ref{stab:confidence}), the reader study (Supplementary Tables~\ref{stab:reader_agreement} to~\ref{stab:taxonomy}), and the paired bootstrap algorithm (Supplementary Algorithm~\ref{alg:paired_bootstrap}).

\refstepcounter{suppnote}
\section*{Supplementary Note \thesuppnote: Prompt texts}
\label{snote:prompts}

The placeholder [display] is replaced by the human-readable finding name (atelectasis, cardiomegaly, consolidation, pulmonary edema, enlarged cardiomediastinum, rib fracture, lung lesion, lung opacity, pleural effusion, pleural abnormality, pneumonia, pneumothorax, support device, and, for the normal studies, any acute abnormality), and the placeholder [sentence] by the sentence of the ReXErr manifest. The same parser is applied to every phrasing.

\medskip
\noindent Default phrasing, finding-presence questions (MS-CXR, MIMIC-CXR, and CheXpert):
\begin{quote}
Is [display] present in this chest X-ray? Answer with a single word: Yes or No.
\end{quote}

\noindent Default phrasing, ReXErr sentence questions:
\begin{quote}
Does the following sentence accurately describe the findings visible in this chest X-ray?\\
Sentence: ``[sentence]''\\
Answer with a single word: Yes or No.
\end{quote}

\noindent Terse phrasing, finding-presence questions:
\begin{quote}
Is [display] present? Yes or No.
\end{quote}

\noindent Terse phrasing, ReXErr sentence questions:
\begin{quote}
Is this sentence accurate for this X-ray? ``[sentence]'' Yes or No.
\end{quote}

\noindent Radiologist-framed phrasing, finding-presence questions:
\begin{quote}
You are a radiologist reviewing a chest X-ray. Is [display] present? Answer with a single word: Yes or No.
\end{quote}

\noindent Radiologist-framed phrasing, ReXErr sentence questions:
\begin{quote}
You are a radiologist. Does the following sentence accurately describe findings in this chest X-ray?\\
Sentence: ``[sentence]''\\
Answer with a single word: Yes or No.
\end{quote}

\noindent System message given to the reasoning model DeepSeek-R1-7B with every phrasing:
\begin{quote}
You are a clinical decision support tool. After your reasoning, you MUST end your response with a single word on its own line: either Yes or No. No other text after that word.
\end{quote}

\refstepcounter{suppnote}
\section*{Supplementary Note \thesuppnote: Answer classes of the parser}
\label{snote:parser}

The fixed parser first decodes the byte-level space and line-break markers of the tokenizer. For the reasoning model, it then compares the last non-empty line, ignoring case and surrounding punctuation, with the affirmative words yes, yeah, correct, true, present, and positive and the negative words no, not, absent, negative, false, and incorrect. For every model, it next compares the first word of the output with the same words. For a non-reasoning model, an output whose first word does not match is an abstention when it contains one of the phrases below. Otherwise, the parser looks in the first 60 characters for a whole-word yes or no and accepts it when the other word does not appear. An output that yields neither Yes nor No is classified in the following order. It is empty when it contains no text. For the reasoning model, it is an abstention when its last non-empty line contains one of the phrases and the output had not ended at the token budget, and truncated otherwise. For every other model, it is truncated when the server reported that the output had ended at the token budget, and unparsed otherwise. The phrases are matched case-insensitively as substrings: need to see; cannot see; can't see; unable to see; cannot determine; can't determine; cannot assess; unable to assess; cannot evaluate; unable to evaluate; no image; without the image; without an image; image is not; not provided; not able to; i cannot; i can't; i am unable; i'm unable; insufficient information; not possible to determine; please provide. The permissive parser of the sensitivity analysis is applied where the fixed parser returns neither Yes nor No, and it accepts the first standalone Yes or No anywhere in the decoded output.

\clearpage

\begin{figure}[p]
\centering
\includegraphics[width=\textwidth]{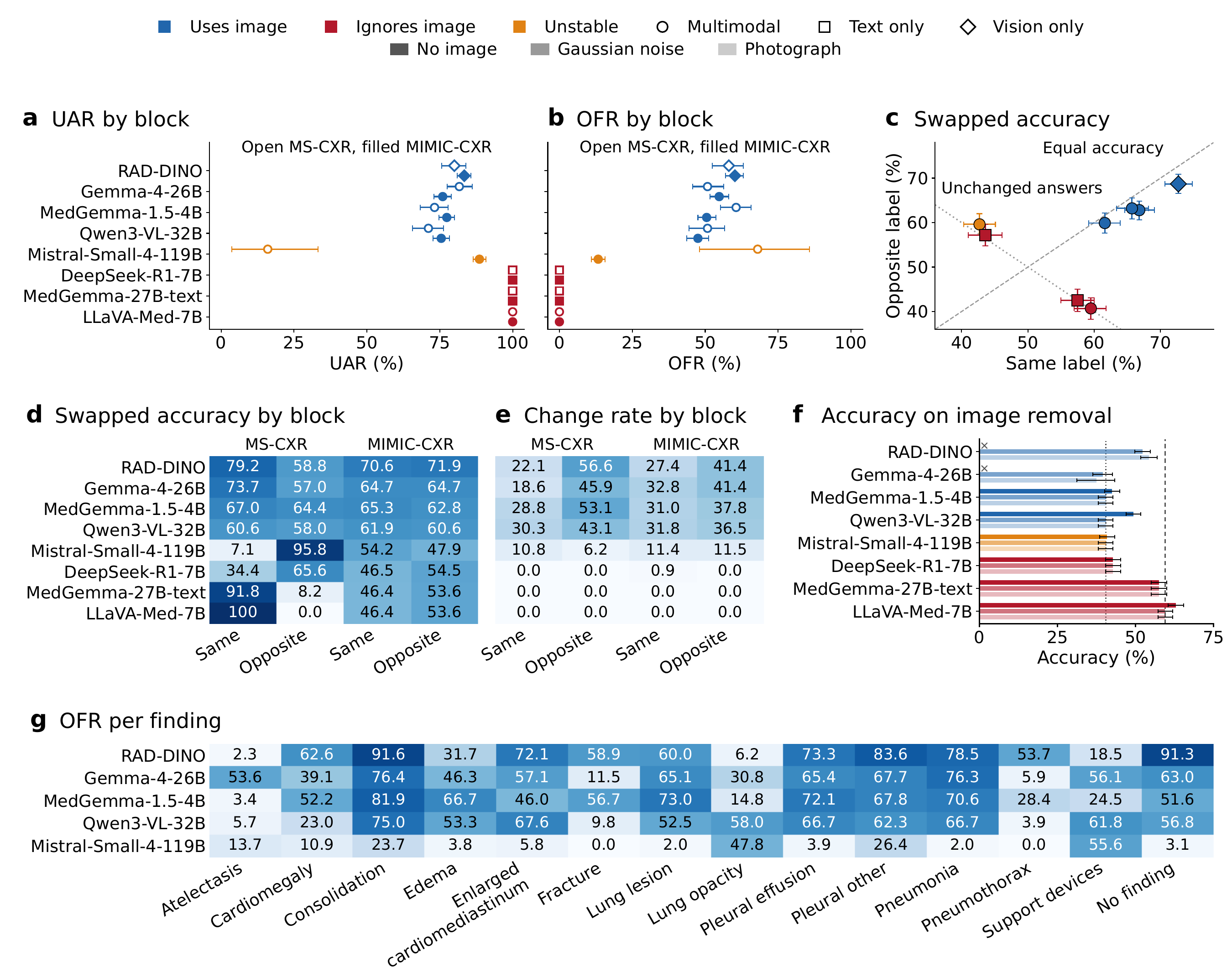}
\caption{Swap metrics per block and per finding, and accuracy under image removal, on the finding-presence cases of the MIMIC question set ($n$ = 1{,}852 questions from 1{,}499 patients). Fill color encodes the behavioral category (blue, uses image; red, ignores image; orange, unstable), and marker shape encodes the modality (circle, multimodal; square, text only; diamond, vision only). A marker, a bar, or a heatmap cell is the value measured on its cases. The error bar of a marker or a bar, drawn in \textbf{a}, \textbf{b}, \textbf{c}, and \textbf{f} only, is the 95\% confidence interval, the 2.5th to the 97.5th percentile of 1{,}000 patient-cluster bootstrap resamples of that value. Case counts per panel: 25 to 907 (\textbf{a}, \textbf{b}), 1{,}708 to 1{,}852 (\textbf{c}), 409 to 1{,}400 (\textbf{d}, \textbf{e}), 253 to 1{,}852 (\textbf{f}), 49 to 163 (\textbf{g}). The systems are ordered by the swap specificity premium, as in Fig.~\ref{fig:categories}b. \textbf{a}, UAR on the MS-CXR block (open, 452 questions) and the MIMIC-CXR block (filled, 1{,}400 questions), on the correct-on-original answers. \textbf{b}, OFR on the same blocks. \textbf{c}, Accuracy against the label of the swapped radiograph under the same-label and the opposite-label swap, on all answered questions. The dashed line marks equal accuracy. The dotted line marks the accuracies of a system that keeps every original answer. \textbf{d}, The accuracy of \textbf{c} on each block. \textbf{e}, Share of all answered questions whose answer changes under each swap, on each block. \textbf{f}, Accuracy with no image, with Gaussian noise, and with a photograph. The dashed line marks the accuracy of a constant Yes answer. The dotted line marks the accuracy of a constant No answer. A cross marks a missing no-image value. \textbf{g}, OFR on the correct-on-original answers of each finding. The systems that ignore the image have an OFR of 0.0 on every finding and are not shown. OFR, opposite-label flip rate; UAR, unrelated-image answer rate.}
\label{sfig:swaps}
\end{figure}

\begin{figure}[p]
\centering
\includegraphics[width=\textwidth]{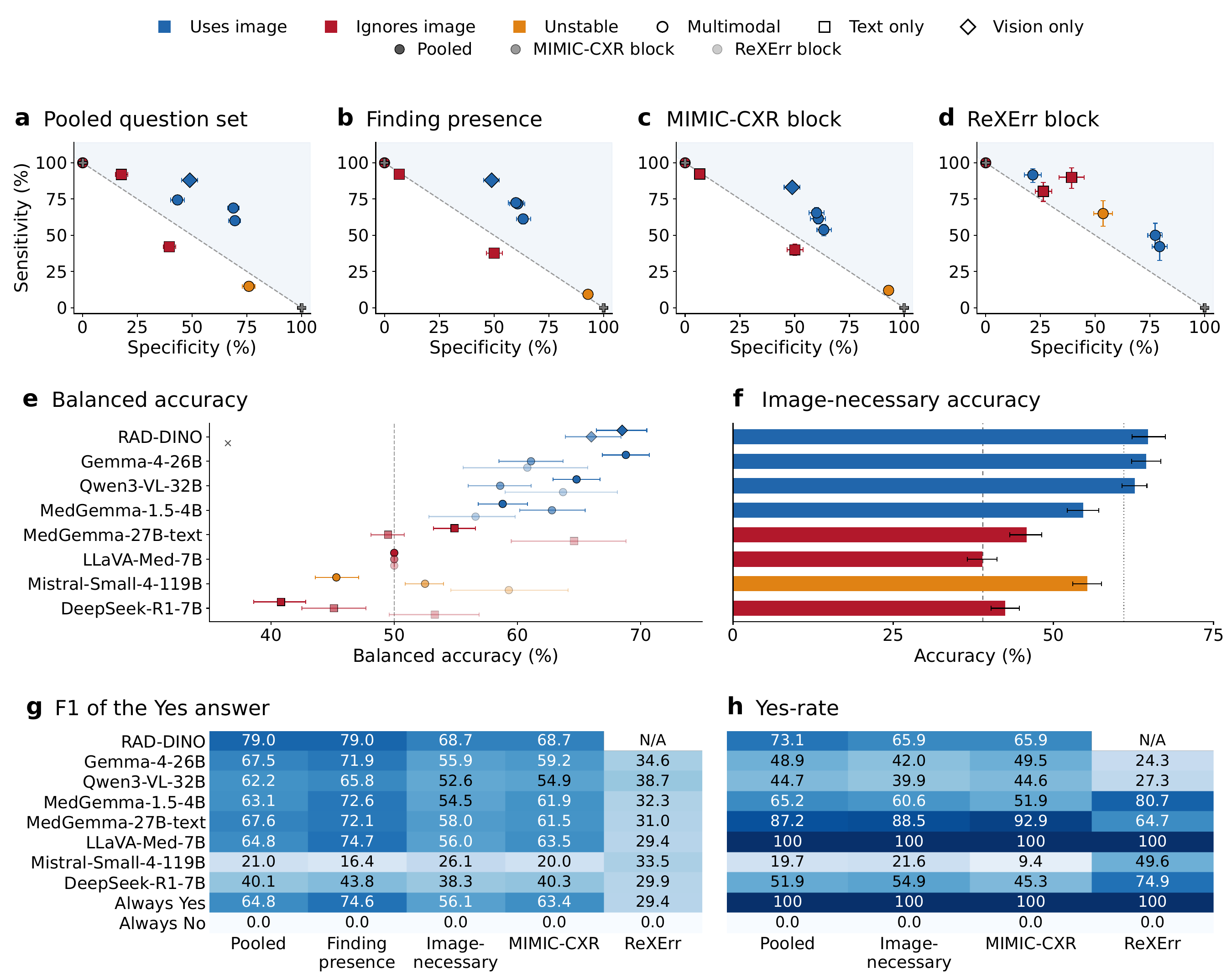}
\caption{The accuracy family of all systems on the MIMIC question set ($n$ = 2{,}548 questions from 1{,}608 patients), for the metrics and question sets absent from Fig.~\ref{fig:accuracy}. Fill color encodes the behavioral category (blue, uses image; red, ignores image; orange, unstable), gray marks the constant answers, and marker shape encodes the modality (circle, multimodal; square, text only; diamond, vision only). A marker, a bar, or a heatmap cell is the value measured on its cases. The error bar of a marker or a bar, drawn in \textbf{a} to \textbf{f} only, is the 95\% confidence interval, the 2.5th to the 97.5th percentile of 1{,}000 patient-cluster bootstrap resamples of that value. Case counts per panel: 750 to 1{,}326 (\textbf{a}), 700 to 1{,}102 (\textbf{b}), 598 to 750 (\textbf{c}), 60 to 576 (\textbf{d}), 445 to 2{,}548 (\textbf{e}, \textbf{g}, \textbf{h}), 1{,}400 to 1{,}976 (\textbf{f}). The systems are ordered by pooled accuracy, as in Fig.~\ref{fig:accuracy}a. The vision-only reference answers only the finding-presence questions. N/A marks its missing ReXErr values in \textbf{g} and \textbf{h}. The MS-CXR block is not drawn, because every question in it has the answer Yes. \textbf{a}--\textbf{d}, Sensitivity against specificity on the pooled question set, the finding-presence questions, the MIMIC-CXR block, and the ReXErr block, with the constant answers as plus markers at the corners. The dashed line marks a balanced accuracy of 50. The shading marks balanced accuracies above 50. \textbf{e}, Balanced accuracy on the pooled question set (dark), the MIMIC-CXR block (medium), and the ReXErr block (light). The dashed line marks 50, the balanced accuracy of a constant answer. A cross marks the missing ReXErr value of the vision-only reference. \textbf{f}, Accuracy on the image-necessary subset. The dashed line marks the accuracy of a constant Yes answer. The dotted line marks the accuracy of a constant No answer. \textbf{g}, F1 of the Yes answer on each question set. A constant No answer never answers Yes and has an F1 of 0. \textbf{h}, Share of the answers that are Yes.}
\label{sfig:accuracy}
\end{figure}

\begin{figure}[p]
\centering
\includegraphics[width=\textwidth]{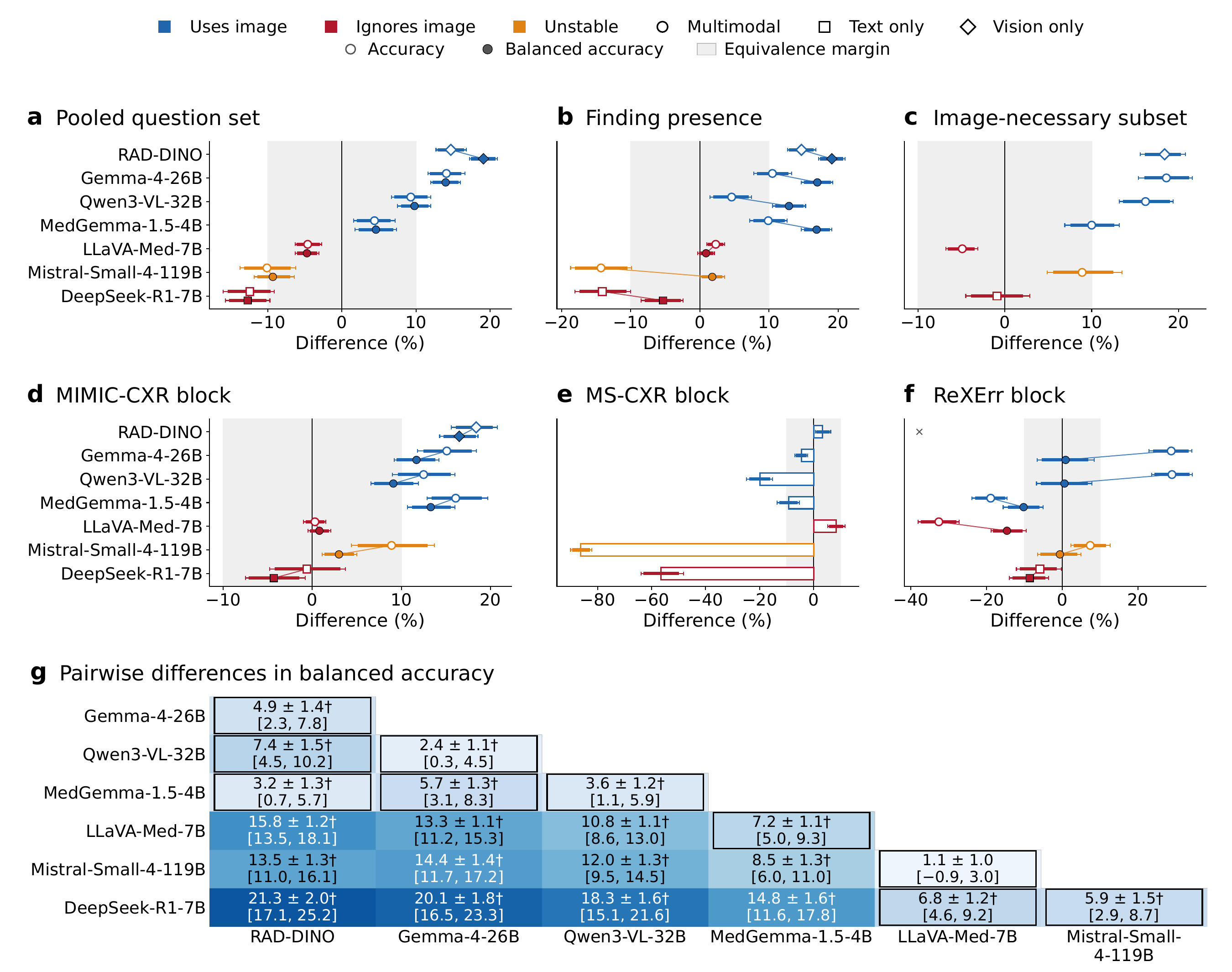}
\caption{Paired differences between each system and the text-only control MedGemma-27B-text, and between the other systems, on the MIMIC question set ($n$ = 2{,}548 questions from 1{,}608 patients). Fill color encodes the behavioral category (blue, uses image; red, ignores image; orange, unstable), and marker shape encodes the modality (circle, multimodal; square, text only; diamond, vision only). Each difference is the system minus the control, on the questions answered by both systems. An open marker or bar is the difference in accuracy, and a filled marker is the difference in balanced accuracy. The thin bar is the 95\% confidence interval, the 2.5th to the 97.5th percentile of 1{,}000 patient-cluster bootstrap resamples of the difference. The thick bar is the 90\% interval of the two one-sided tests. Equivalence is established where it lies inside the shaded margin of $\pm$10\%. Every FDR-corrected $p$-value is shown in Supplementary Fig.~\ref{sfig:vs_deepseek}g. Case counts per panel: 1{,}852 to 2{,}297 (\textbf{a}), 1{,}708 to 1{,}852 (\textbf{b}), 1{,}400 to 1{,}744 (\textbf{c}), 1{,}298 to 1{,}400 (\textbf{d}), 410 to 452 (\textbf{e}), 437 to 445 (\textbf{f}), 1{,}298 to 1{,}976 (\textbf{g}). The systems are ordered by pooled accuracy, as in Fig.~\ref{fig:accuracy}a. \textbf{a}--\textbf{f}, Difference from the control on each question set. Balanced accuracy is drawn in Fig.~\ref{fig:accuracy}e for the image-necessary subset and is undefined on the MS-CXR block, whose questions all have the answer Yes. A cross marks the missing ReXErr value of the vision-only reference. \textbf{g}, Difference in balanced accuracy on the image-necessary subset between every two systems other than the control, the column minus the row. Each cell contains the observed difference $\pm$ the standard deviation of the bootstrap resamples, with the 95\% confidence interval below. A dagger marks $p_{\mathrm{FDR}} < 0.05$ in the paired bootstrap test, corrected within the family of 28 pairs. A black frame marks equivalence within $\pm$10\%. FDR, false discovery rate.}
\label{sfig:vs_medgemma}
\end{figure}

\begin{figure}[p]
\centering
\includegraphics[width=\textwidth]{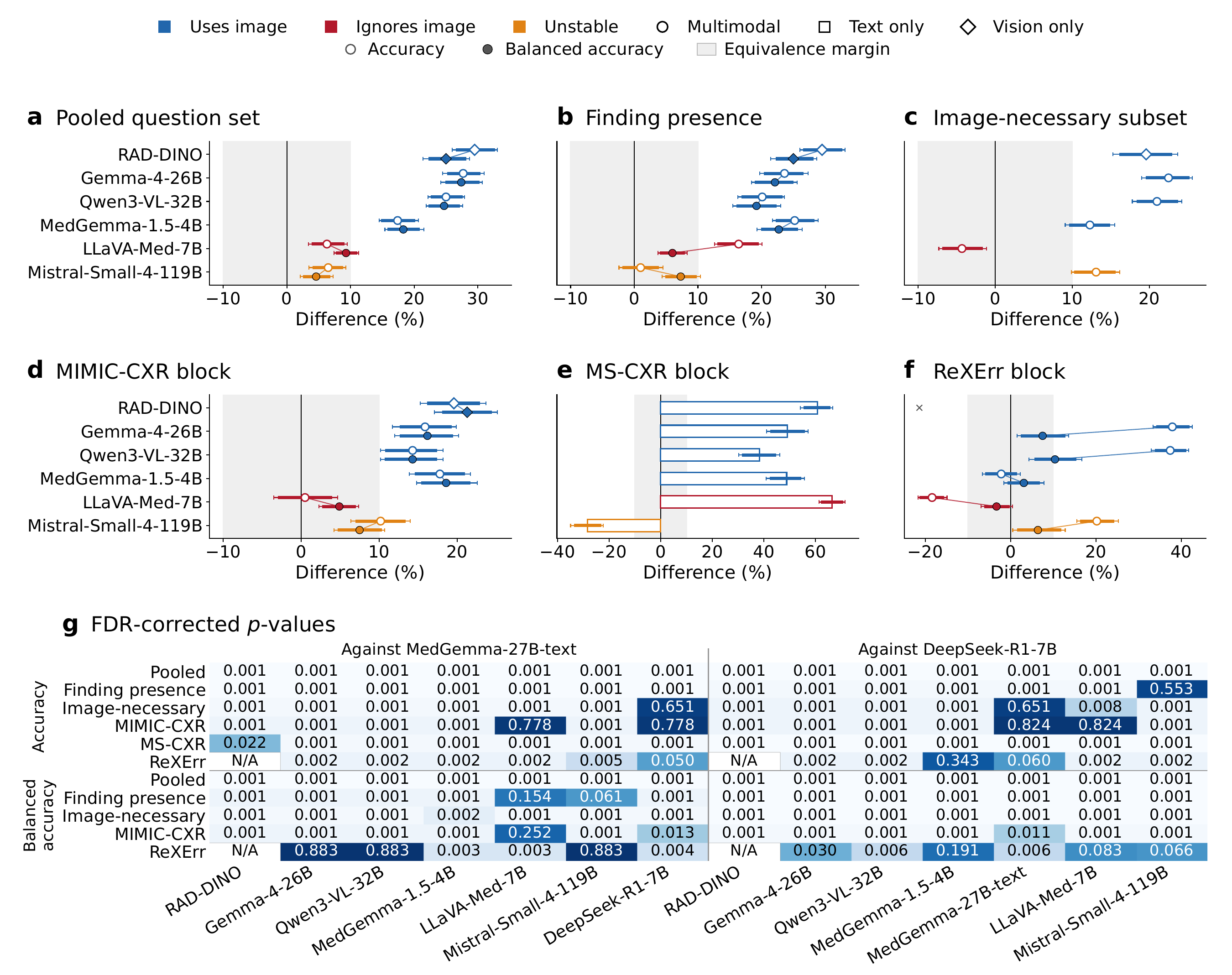}
\caption{Paired differences between each system and the text-only control DeepSeek-R1-7B, and the FDR-corrected $p$-value of every comparison with a control, on the MIMIC question set ($n$ = 2{,}548 questions from 1{,}608 patients). Fill color encodes the behavioral category (blue, uses image; red, ignores image; orange, unstable), and marker shape encodes the modality (circle, multimodal; square, text only; diamond, vision only). Each difference is the system minus the control, on the questions answered by both systems. An open marker or bar is the difference in accuracy, and a filled marker is the difference in balanced accuracy. The thin bar is the 95\% confidence interval, the 2.5th to the 97.5th percentile of 1{,}000 patient-cluster bootstrap resamples of the difference. The thick bar is the 90\% interval of the two one-sided tests. Equivalence is established where it lies inside the shaded margin of $\pm$10\%. Case counts per panel: 1{,}708 to 2{,}392 (\textbf{a}), 1{,}659 to 1{,}708 (\textbf{b}), 1{,}298 to 1{,}862 (\textbf{c}), 1{,}270 to 1{,}298 (\textbf{d}), 384 to 410 (\textbf{e}), 676 to 684 (\textbf{f}). The systems are ordered by pooled accuracy, as in Fig.~\ref{fig:accuracy}a. The two controls are compared in Supplementary Fig.~\ref{sfig:vs_medgemma}a--f (DeepSeek-R1-7B minus MedGemma-27B-text). \textbf{a}--\textbf{f}, Difference from the control on each question set. Balanced accuracy on the image-necessary subset is drawn in Supplementary Fig.~\ref{sfig:vs_medgemma}g. Balanced accuracy is undefined on the MS-CXR block, whose questions all have the answer Yes. A cross marks the missing ReXErr value of the vision-only reference. \textbf{g}, FDR-corrected $p$-value of the paired bootstrap test for every comparison with MedGemma-27B-text (left) and with DeepSeek-R1-7B (right), corrected within the seven comparisons with one control on one question set and one metric. A darker cell marks a larger $p$-value. N/A marks the vision-only reference on the ReXErr block. FDR, false discovery rate.}
\label{sfig:vs_deepseek}
\end{figure}

\begin{figure}[p]
\centering
\includegraphics[width=\textwidth]{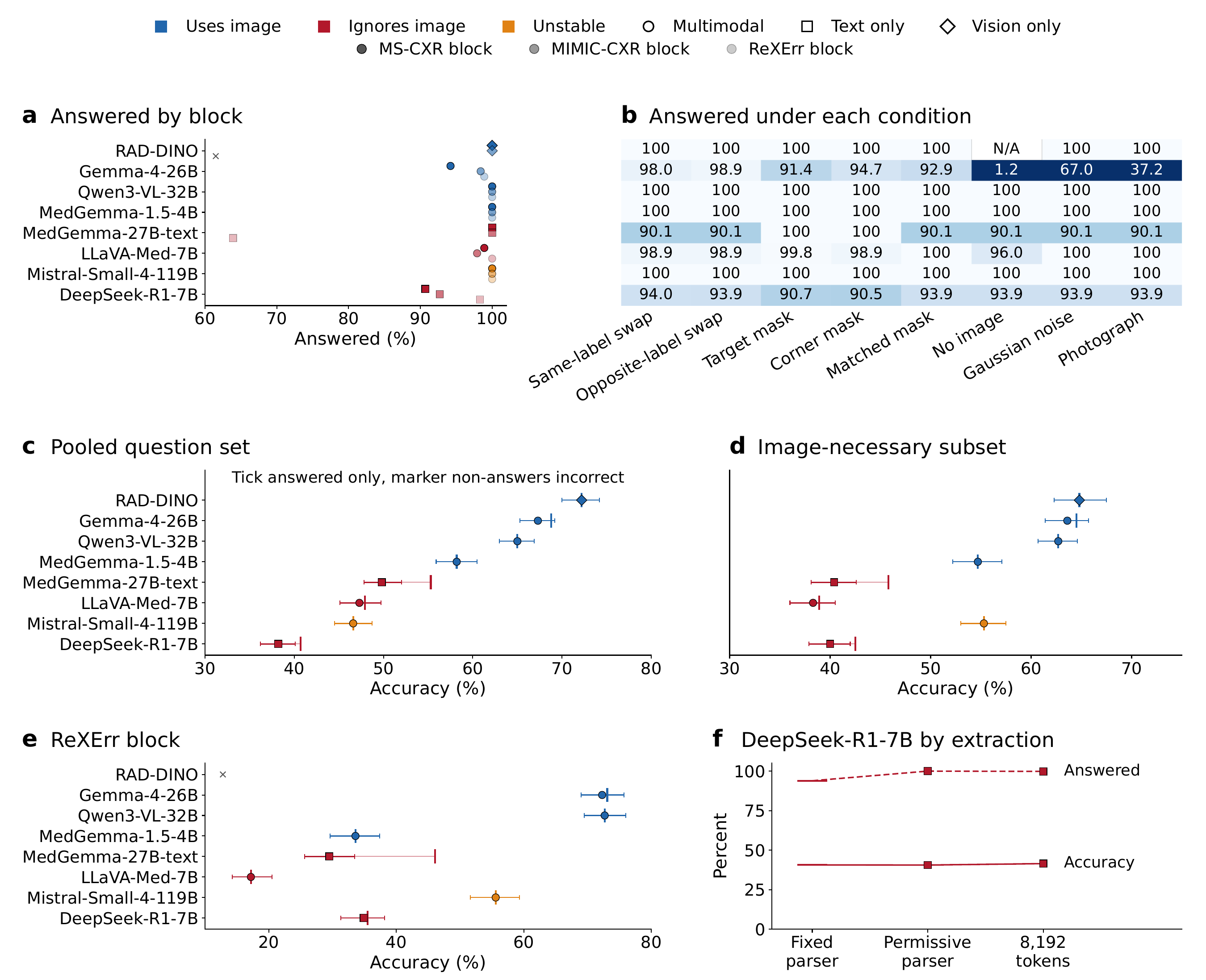}
\caption{Non-answers of every system and their effect on accuracy, on the MIMIC question set ($n$ = 2{,}548 questions from 1{,}608 patients). Fill color encodes the behavioral category (blue, uses image; red, ignores image; orange, unstable), and marker shape encodes the modality (circle, multimodal; square, text only; diamond, vision only). A non-answer is an abstention, a truncation, an empty output, or an unparsed output (Supplementary Note~\ref{snote:parser}). In \textbf{c} to \textbf{f}, a tick marks the main analysis, which excludes non-answers, and a marker is the sensitivity analysis. Its error bar is the 95\% confidence interval, the 2.5th to the 97.5th percentile of 1{,}000 patient-cluster bootstrap resamples of that value. Case counts per panel: 452 to 1{,}400 (\textbf{a}), 452 to 2{,}548 (\textbf{b}), 1{,}852 to 2{,}548 (\textbf{c}), 1{,}400 to 1{,}976 (\textbf{d}), 696 (\textbf{e}), 2{,}542 to 2{,}548 (\textbf{f}). The systems are ordered as in Fig.~\ref{fig:accuracy}a. \textbf{a}, Share of each block answered Yes or No under the original condition, with the block encoded by the shade. A cross marks the vision-only reference, which answers no ReXErr question. \textbf{b}, Share answered under each other condition, over the cases on which the condition exists. Without a call, a text-only control keeps the share of its original call on all questions. The shade grows with the share left unanswered. N/A marks the vision-only reference without an image. \textbf{c}--\textbf{e}, Accuracy with every non-answer counted as incorrect, on the pooled question set, the image-necessary subset, and the ReXErr block. The tick is the accuracy on the answered cases alone, drawn in Fig.~\ref{fig:accuracy}a,b and Supplementary Fig.~\ref{sfig:accuracy}f. \textbf{f}, Answered share and accuracy of DeepSeek-R1-7B under the fixed parser, under a permissive parser that accepts the first Yes or No anywhere in the output, and after a rerun of its truncated outputs with a budget of 8{,}192 tokens.}
\label{sfig:nonanswers}
\end{figure}

\begin{figure}[p]
\centering
\includegraphics[width=\textwidth]{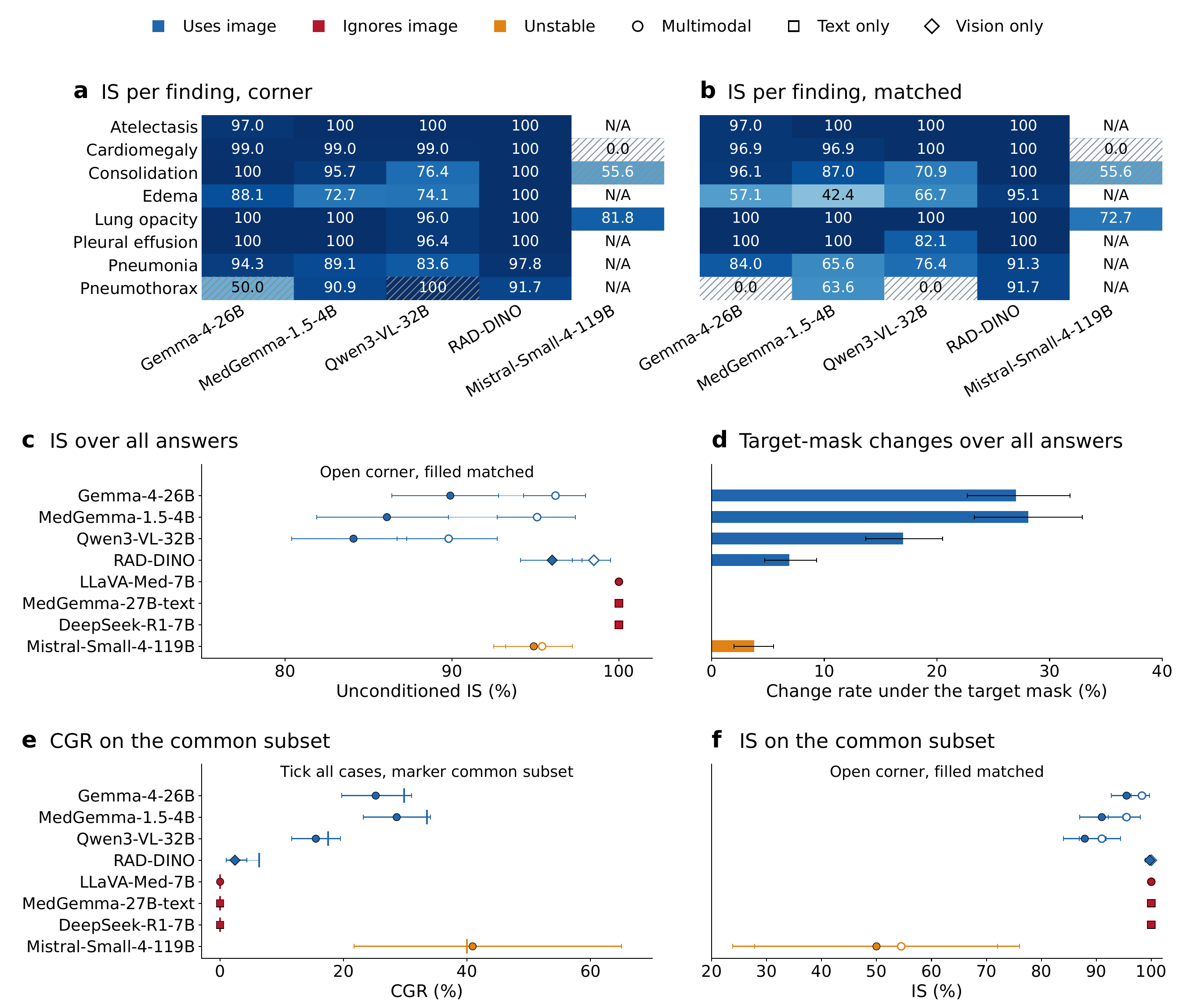}
\caption{Localization metrics absent from Fig.~\ref{fig:localization}, on the MS-CXR block of the MIMIC question set ($n$ = 452 cases with a radiologist-marked box, 321 patients). Fill color encodes the behavioral category (blue, uses image; red, ignores image; orange, unstable), and marker shape encodes the modality (circle, multimodal; square, text only; diamond, vision only). In \textbf{c} and \textbf{f}, open markers are the corner placement and filled markers are the matched placement of the irrelevant mask. A marker, a bar, or a heatmap cell is the value measured on its cases. The error bar of a marker or a bar, drawn in \textbf{c} to \textbf{f} only, is the 95\% confidence interval, the 2.5th to the 97.5th percentile of 1{,}000 patient-cluster bootstrap resamples of that value. A hatch marks a cell of fewer than 10 cases. Case counts per panel: 1 to 100 (\textbf{a}, \textbf{b}), 409 to 452 (\textbf{c}), 404 to 452 (\textbf{d}), 22 to 290 (\textbf{e}, \textbf{f}). The systems are ordered as in Fig.~\ref{fig:localization}a. \textbf{a}, \textbf{b}, IS per finding under the corner and the matched placement, for the image users and the unstable system, with the findings in the rows as in Fig.~\ref{fig:localization}f. N/A marks a finding with no correct answer. \textbf{c}, IS over all answered cases, correct or not. \textbf{d}, Share of all answered cases whose answer changes under the target mask. \textbf{e}, CGR on the 290 cases answered correctly by every image user, with a tick at the CGR on all cases, drawn in Fig.~\ref{fig:localization}a. The counts of the other systems are their correct answers among these cases. \textbf{f}, IS on the same cases. CGR, causal grounding rate; IS, irrelevant-mask stability.}
\label{sfig:localization}
\end{figure}

\begin{figure}[p]
\centering
\includegraphics[width=\textwidth]{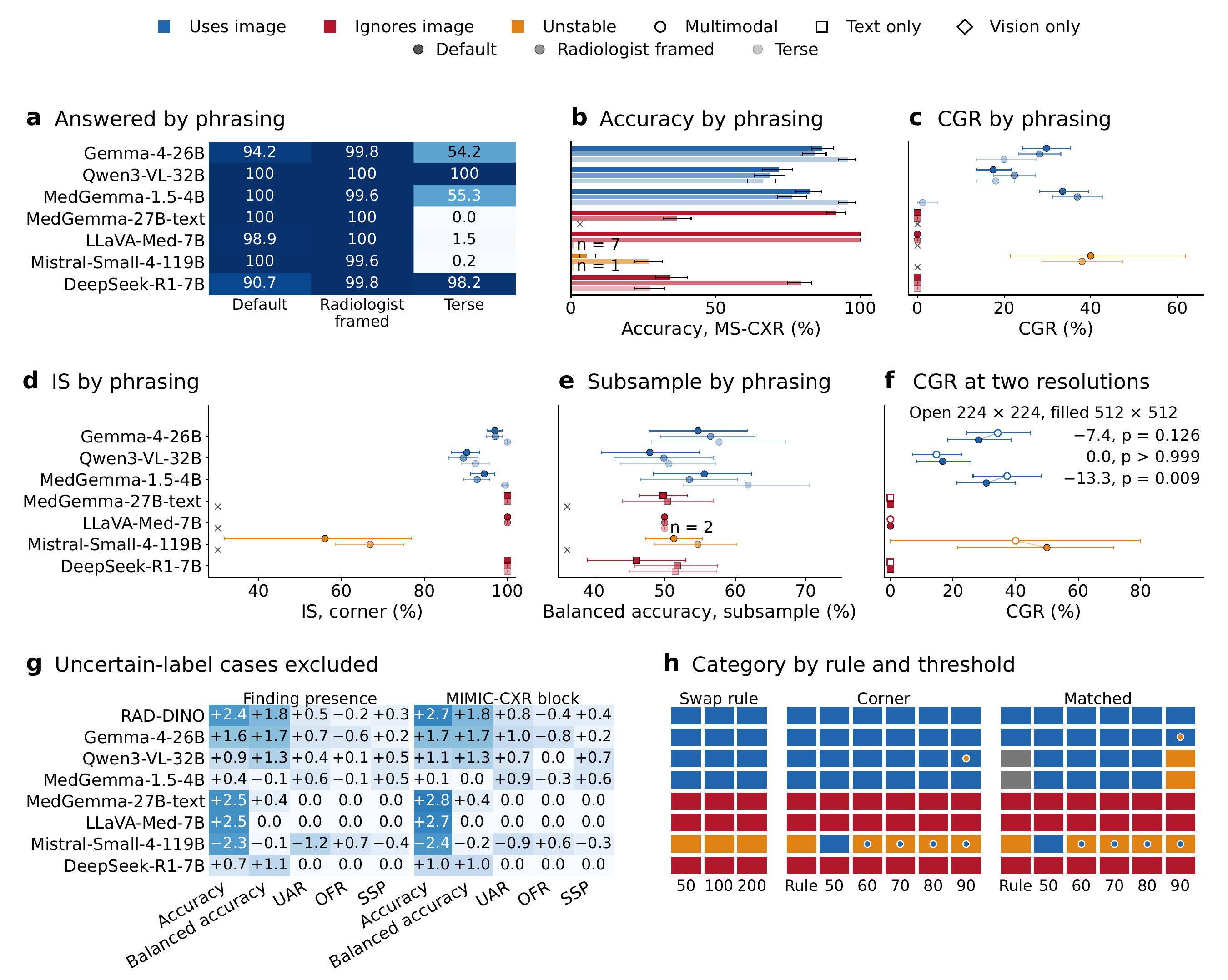}
\caption{Sensitivity analyses of the audit on the MIMIC question set ($n$ = 2{,}548 questions from 1{,}608 patients). Fill color encodes the behavioral category (blue, uses image; red, ignores image; orange, unstable; gray, unassigned), and marker shape encodes the modality (circle, multimodal; square, text only; diamond, vision only). In \textbf{b} to \textbf{e}, the shade encodes the phrasing (Supplementary Note~\ref{snote:prompts}). A marker, a bar, or a heatmap cell is the value measured on its cases. The error bar of a marker or a bar is the 95\% confidence interval, the 2.5th to the 97.5th percentile of 1{,}000 patient-cluster bootstrap resamples of that value. A value computed from fewer than 10 cases carries its case count, and its bar is hatched. A tick marks a bar of zero, and a cross marks a missing value. Case counts per panel: 452 (\textbf{a}), 1 to 452 (\textbf{b}), 25 to 452 (\textbf{c}, \textbf{d}), 2 to 200 (\textbf{e}), 5 to 99 (\textbf{f}), 590 to 1{,}774 (\textbf{g}). The systems are ordered by pooled accuracy, as in Fig.~\ref{fig:accuracy}a, and \textbf{a} to \textbf{f} omit the vision-only reference. \textbf{a}--\textbf{d}, Answered share, accuracy, CGR, and IS under the corner placement, on the 452 MS-CXR cases. \textbf{e}, Balanced accuracy on a 200-case subsample of the MIMIC-CXR block. \textbf{f}, CGR on 100 MS-CXR cases at $224 \times 224$ and $512 \times 512$ pixels, with the paired difference and its two-sided paired bootstrap $p$-value. \textbf{g}, Change of each metric, shaded by its size, when the 78 finding-absent cases with an uncertain label are excluded. \textbf{h}, Category under the swap rule at minimum counts of 50, 100, and 200 and under the alternative rule at its defined thresholds (Rule) or a common threshold of 50 to 90, per placement. A dot in the color of the new category marks a change under the unconditioned IS. CGR, causal grounding rate; IS, irrelevant-mask stability; OFR, opposite-label flip rate; SSP, swap specificity premium; UAR, unrelated-image answer rate.}
\label{sfig:robustness}
\end{figure}

\clearpage

\begin{table}[t]
\centering
\caption{Glossary of the terms of this audit, grouped by analysis set, intervention, metric, behavioral category, and reporting convention. Every rate is a percentage computed on the 2{,}548 questions of the MIMIC question set or on the subset named in its definition. The percent sign is omitted throughout. The evaluated systems are listed in Supplementary Table~\ref{stab:registry} and the prompt texts in Supplementary Note~\ref{snote:prompts}.}
\label{stab:glossary}
\setlength{\tabcolsep}{4pt}
\renewcommand{\arraystretch}{1.10}
\scriptsize
\begin{tabular}{@{}>{\raggedright\arraybackslash}p{0.26\textwidth}>{\raggedright\arraybackslash}p{0.66\textwidth}@{}}
\toprule
Term & Definition \\
\midrule
\multicolumn{2}{@{}l}{\textbf{Question set and analysis sets}} \\
MIMIC question set & The 2{,}548 yes-or-no questions of the primary question set, assembled from MS-CXR phrase-grounding boxes, MIMIC-CXR labels, and ReXErr-v1 report sentences. \\
CheXpert question set & The 1{,}380 finding-presence questions drawn from a held-out fifth of the CheXpert training set, on which the transfer of the categories is tested. \\
Finding-presence question & A question asking whether a named finding is present in the radiograph. The MS-CXR and MIMIC-CXR blocks together contain 1{,}852 of them. \\
Image-necessary subset & The 1{,}976 questions whose correct answer depends on the radiograph, meaning the MIMIC-CXR block with the ReXErr image-dependent errors and error-free controls. \\
Informative cases & The correct-on-original answers evaluated under a given intervention, which are the denominator of UAR, OFR, and CGR. \\
\midrule
\multicolumn{2}{@{}l}{\textbf{Interventions on the image}} \\
Same-label swap & Another patient's radiograph with the same label for the queried finding, presented in place of the original. \\
Opposite-label swap & Another patient's radiograph with the opposite label for the queried finding, presented in place of the original. \\
Target mask & The radiologist-marked region of the queried finding, occluded by a black rectangle. \\
Corner irrelevant mask & A region of the same size as the marked region, occluded at the image corner farthest from it. \\
Matched irrelevant mask & A region of the same size, occluded at the marked region mirrored across the midline, shifted along the same side of the image where a mirrored placement would overlap the marked region, or at the image corner where neither fits. \\
Image removal & The conditions in which no radiograph is presented, meaning no image, one fixed Gaussian-noise image, and one fixed photograph. \\
\midrule
\multicolumn{2}{@{}l}{\textbf{Behavioral metrics}} \\
Unrelated-image answer rate (UAR) & The share of correct answers unchanged under the same-label swap. \\
Opposite-label flip rate (OFR) & The share of correct answers changed under the opposite-label swap. \\
Swap specificity premium (SSP) & OFR minus (100 minus UAR). It is positive when answers change with the label of the radiograph and not only with its identity. \\
Causal grounding rate (CGR) & The share of correct affirmative answers changed under the target mask. \\
Irrelevant-mask stability (IS) & The share of correct answers unchanged under an irrelevant mask, reported under both placements. \\
Grounding specificity premium (GSP) & CGR minus (100 minus IS). It is positive when occluding the marked region changes more answers than occluding a region of the same size elsewhere. \\
\midrule
\multicolumn{2}{@{}l}{\textbf{Behavioral categories}} \\
Uses image & A system whose SSP is positive with a 95\% confidence interval above zero. \\
Ignores image & A system whose OFR is 0 and whose UAR is 100, each on at least 100 informative cases. \\
Unstable & Any system assigned by neither rule above. \\
\midrule
\multicolumn{2}{@{}l}{\textbf{Reporting}} \\
Accuracy & The share of answered questions answered correctly. Balanced accuracy is the mean of sensitivity and specificity, on which a constant answer scores 50. \\
Yes-rate & The share of answered questions answered Yes. \\
Non-answer & An output with no Yes or No, meaning an abstention, an empty output, a truncated output, or an unparsed output, each defined in Supplementary Note~\ref{snote:parser}. Non-answers are excluded from every rate and reported separately. \\
Patient-cluster bootstrap & The 1{,}000 resamples of patients at seed 0 from which the standard deviation and the percentile 95\% confidence interval of every per-system rate and every paired difference are computed. \\
Two one-sided tests & The equivalence test of a paired difference in accuracy or balanced accuracy against a margin of 10\%. \\
Benjamini-Hochberg false discovery rate (FDR) & The multiplicity correction applied within each family of tests. A comparison is significant when its corrected value is below 0.05. \\
\bottomrule
\end{tabular}
\end{table}

\begin{table}[t]
\centering
\caption{Summary of the findings of the audit, in the order in which they are reported. Values are taken unchanged from the main text and from Tables~\ref{tab:master} to~\ref{tab:subgroups}. Rates are percentages on the 2{,}548 questions of the question set or on the subset named in the row, and a bracketed interval is a 95\% confidence interval. CGR, causal grounding rate; UAR, unrelated-image answer rate; OFR, opposite-label flip rate; SSP, swap specificity premium; GSP, grounding specificity premium; IS, irrelevant-mask stability; AP, anteroposterior; PA, posteroanterior; FDR, false discovery rate.}
\label{stab:findings}
\setlength{\tabcolsep}{4pt}
\renewcommand{\arraystretch}{1.0}
\scriptsize
\begin{tabular}{@{}>{\raggedright\arraybackslash}p{0.16\textwidth}>{\raggedright\arraybackslash}p{0.18\textwidth}>{\raggedright\arraybackslash}p{0.33\textwidth}>{\raggedright\arraybackslash}p{0.25\textwidth}@{}}
\toprule
Analysis & Comparison & Result & Interpretation \\
\midrule
Behavioral categories & Both swaps, on the correct-on-original answers & OFR 0.0 and UAR 100.0 for three systems. OFR 48.4 to 59.5 and SSP 22.7 to 41.8 (intervals above zero) for four systems. OFR 15.0 and SSP 1.4 [$-1.0$, 3.9] for one system & LLaVA-Med-7B and the two text-only controls ignore the image, four systems use it, and Mistral-Small-4-119B changes its answers without following the label \\
\midrule
Image removal & No image, Gaussian noise, and a photograph compared with the original radiograph & Yes-rate 0.0 under noise and the photograph for every multimodal system except LLaVA-Med-7B. Correct answers kept 25.1 to 41.3 for the multimodal image users, 87.1 for Mistral-Small-4-119B, and 100.0 for LLaVA-Med-7B & Without a radiograph, the multimodal systems other than LLaVA-Med-7B answer mostly No or decline to answer \\
\midrule
Pooled accuracy & All systems and the constant answers, on the pooled question set & Text-only control 55.3, three systems above it at 58.2 to 68.8, constant Yes at 48.0, and constant No at 52.0 & A system that receives no image outscores two multimodal systems \\
\midrule
Accuracy by source block & The text-only control by source block & Higher than Gemma-4-26B by 4.5 on MS-CXR (accuracy 91.8) and lower than the constant No answer on MIMIC-CXR (46.4 vs 53.6). Yes-rate 92.6 on the finding-presence questions & The block in which every finding is present raises the pooled score of the control \\
\midrule
Image-necessary questions & Balanced accuracy on the 1{,}976 questions that need the image & Balanced accuracy 57.4 to 66.0 for the image users and 53.0 for the control. Paired advantage 4.6 to 16.5 (all $p_{\mathrm{FDR}} \leq 0.002$) & Every image user exceeds the control. For two of them, the advantage is below 10 by the equivalence test \\
\midrule
Localization & Target mask on the 452 MS-CXR cases & CGR 6.3 to 33.5 for the image users and 0.0 for the ignores-image systems & Occluding the marked region changes a minority of correct affirmative answers \\
\midrule
Mask placement & Corner vs anatomically matched irrelevant mask & IS 90.2 to 99.1 at the corner and 84.7 to 97.2 at the matched position. GSP is positive under both placements and smaller by 1.9 to 9.7 at the matched placement & Occluding the marked region changes more answers than occluding an equal region elsewhere \\
\midrule
Findings, views, and patients & CGR and OFR by finding, view, sex, and age band & CGR 0.0 to 4.0 on lung opacity and up to 69.7 on edema. CGR is higher on PA than on AP radiographs for every image user, significantly for Gemma-4-26B (72.1 vs 21.0, $p_{\mathrm{FDR}} = 0.004$) & Localization varies across findings. Views are compared without adjustment \\
\midrule
Transfer to CheXpert & The category rule applied unchanged to 1{,}380 CheXpert questions & Every system receives the same category as on the MIMIC question set. The five systems whose answers change keep their order in OFR & The categories are the same on a second dataset from another institution \\
\midrule
Confidence & Confidence on grounded vs ungrounded correct answers & Mean confidence is lower on grounded than on ungrounded correct answers for every image user, 97.9 vs 99.7 for Gemma-4-26B. Expected calibration error 0.139 to 0.505 & Confidence is not higher for an answer that depends on the image \\
\midrule
Radiologist comparison & Three radiologists, two of them on 200 balanced cases, and the systems & Reader accuracy 86.0 and 82.0, system accuracy 50.0 to 73.0 (all $p_{\mathrm{FDR}} \leq 0.005$). Reader CGR 38.4 and 13.3, CGR of the multimodal image users 20.5 to 27.7 & No system is as accurate as either reader. The CGR of the multimodal image users lies between those of the readers \\
\bottomrule
\end{tabular}
\end{table}

\begin{table}[t]
\centering
\caption{Composition of the MIMIC question set by source, finding, label, view, and sex ($n$ = 2{,}548 cases from 1{,}608 patients). Yes and No count the cases whose correct answer is Yes and No. For a finding-presence question, these are the finding-present and finding-absent cases. For a ReXErr question, they are the error-free and the erroneous sentences. Fallback counts the finding-absent cases with an uncertain label, drawn because fewer than 50 patients of the test split have the finding marked absent (all 50 edema and 28 pleural-other cases). The 100 normal studies are asked whether any acute abnormality is present, and their correct answer is No. PA and AP count posteroanterior and anteroposterior acquisitions, and F and M count female and male patients.}
\label{stab:composition}
\setlength{\tabcolsep}{4pt}
\renewcommand{\arraystretch}{1.05}
\footnotesize
\begin{tabular}{@{}p{0.34\textwidth}rrrrrrr@{}}
\toprule
Source and finding & Yes & No & Fallback & PA & AP & F & M \\
\midrule
\multicolumn{8}{@{}l}{\textbf{MS-CXR (phrase-grounded, with target boxes)}} \\
Atelectasis & 35 & 0 & 0 & 4 & 31 & 18 & 17 \\
Cardiomegaly & 100 & 0 & 0 & 26 & 74 & 48 & 52 \\
Consolidation & 76 & 0 & 0 & 12 & 64 & 28 & 48 \\
Edema & 43 & 0 & 0 & 8 & 35 & 13 & 30 \\
Lung opacity & 30 & 0 & 0 & 4 & 26 & 11 & 19 \\
Pleural effusion & 34 & 0 & 0 & 4 & 30 & 11 & 23 \\
Pneumonia & 97 & 0 & 0 & 17 & 80 & 43 & 54 \\
Pneumothorax & 37 & 0 & 0 & 11 & 26 & 16 & 21 \\
\textbf{Subtotal} & 452 & 0 & 0 & 86 & 366 & 188 & 264 \\
\midrule
\multicolumn{8}{@{}l}{\textbf{MIMIC-CXR (globally labeled, no target boxes)}} \\
Atelectasis & 50 & 50 & 0 & 26 & 74 & 50 & 50 \\
Cardiomegaly & 50 & 50 & 0 & 38 & 62 & 47 & 53 \\
Consolidation & 50 & 50 & 0 & 39 & 61 & 61 & 39 \\
Edema & 50 & 50 & 50 & 10 & 90 & 45 & 55 \\
Enlarged cardiomediastinum & 50 & 50 & 0 & 37 & 63 & 47 & 53 \\
Fracture & 50 & 50 & 0 & 51 & 49 & 59 & 41 \\
Lung lesion & 50 & 50 & 0 & 58 & 42 & 47 & 53 \\
Lung opacity & 50 & 50 & 0 & 40 & 60 & 49 & 51 \\
Pleural effusion & 50 & 50 & 0 & 39 & 61 & 55 & 45 \\
Pleural other & 50 & 50 & 28 & 62 & 38 & 47 & 53 \\
Pneumonia & 50 & 50 & 0 & 42 & 58 & 46 & 54 \\
Pneumothorax & 50 & 50 & 0 & 23 & 77 & 39 & 61 \\
Support devices & 50 & 50 & 0 & 28 & 72 & 38 & 62 \\
No finding & 0 & 100 & 0 & 60 & 40 & 48 & 52 \\
\textbf{Subtotal} & 650 & 750 & 78 & 553 & 847 & 678 & 722 \\
\midrule
\multicolumn{8}{@{}l}{\textbf{ReXErr-v1 (report-sentence questions over MIMIC-CXR images)}} \\
Image-dependent errors & 0 & 456 & 0 & 130 & 326 & 217 & 239 \\
Text-only errors & 0 & 120 & 0 & 32 & 88 & 53 & 67 \\
No-error controls & 120 & 0 & 0 & 44 & 76 & 65 & 55 \\
\textbf{Subtotal} & 120 & 576 & 0 & 206 & 490 & 335 & 361 \\
\midrule
\textbf{Question set} & 1222 & 1326 & 78 & 845 & 1703 & 1201 & 1347 \\
\bottomrule
\end{tabular}
\end{table}

\begin{table}[t]
\centering
\caption{Registry of the evaluated systems. Parameters are the nominal sizes of the releases in billions, with the active count given for mixture-of-experts models. Multimodal denotes text-and-image input, text only denotes a language model without image input, and vision only denotes a frozen image encoder with one logistic-regression head per finding. MedGemma-27B-text is the multimodal MedGemma 27B release, called with the question and no image as a text-only control. Every system is open weight, and Release is the month in which its checkpoint was made public.}
\label{stab:registry}
\setlength{\tabcolsep}{4pt}
\renewcommand{\arraystretch}{1.15}
\footnotesize
\begin{tabular}{@{}p{0.17\textwidth}p{0.12\textwidth}p{0.33\textwidth}p{0.14\textwidth}p{0.12\textwidth}@{}}
\toprule
Model & Parameters (billion) & Category & Developer & Release \\
\midrule
Gemma-4-26B & 26 (4 active) & Multimodal, general purpose, mixture of experts & Google DeepMind & April 2026 \\
Qwen3-VL-32B & 32 & Multimodal, general purpose & Alibaba (Qwen) & October 2025 \\
Mistral-Small-4-119B & 119 (6.5 active) & Multimodal, general purpose, optional reasoning, mixture of experts & Mistral AI & March 2026 \\
MedGemma-1.5-4B & 4 & Multimodal, medical specialist & Google DeepMind & January 2026 \\
LLaVA-Med-7B & 7 & Multimodal, medical specialist & Microsoft & May 2024 \\
MedGemma-27B-text & 27 & Multimodal release called without an image, medical specialist & Google DeepMind & July 2025 \\
DeepSeek-R1-7B & 7 & Text only, reasoning (distilled) & DeepSeek & January 2025 \\
RAD-DINO & 0.09 & Vision only, image encoder with logistic heads & Microsoft & May 2024 \\
\bottomrule
\end{tabular}
\end{table}

\begin{table}[t]
\centering
\caption{Irrelevant-mask stability under the matched placement by placement class on the MS-CXR block ($n$ = 452 cases). The matched mask is the target box mirrored across the vertical midline (mirrored, 322 cases). Where the mirrored box overlaps the target, it is shifted along the same side of the image to the nearest non-overlapping position (shifted, 119 cases). A box too large for a mirrored or a shifted position is masked at the corner (corner fallback, 11 cases). Each value is the share of correct-on-original answers unchanged under the mask, as the mean over its cases $\pm$ the standard deviation of 1{,}000 patient-cluster bootstrap resamples, with the percentile 95\% confidence interval and the case count $n$.}
\label{stab:is_placement}
\setlength{\tabcolsep}{4pt}
\renewcommand{\arraystretch}{1.05}
\scriptsize
\begin{tabular}{@{}p{0.2\textwidth}p{0.23\textwidth}p{0.23\textwidth}p{0.23\textwidth}@{}}
\toprule
Model & Mirrored & Shifted & Corner fallback \\
\midrule
Gemma-4-26B & \begin{tabular}[t]{@{}l@{}}88.8 $\pm$ 2.2 [84.3, 92.9]\\ {\scriptsize $n=242$}\end{tabular} & \begin{tabular}[t]{@{}l@{}}95.5 $\pm$ 2.0 [91.1, 99.1]\\ {\scriptsize $n=110$}\end{tabular} & \begin{tabular}[t]{@{}l@{}}72.7 $\pm$ 13.3\\ {\scriptsize [45.5, 100.0]}\\ {\scriptsize $n=11$}\end{tabular} \\
Qwen3-VL-32B & \begin{tabular}[t]{@{}l@{}}81.6 $\pm$ 2.7 [76.5, 86.7]\\ {\scriptsize $n=207$}\end{tabular} & \begin{tabular}[t]{@{}l@{}}97.2 $\pm$ 1.6\\ {\scriptsize [93.5, 100.0]}\\ {\scriptsize $n=108$}\end{tabular} & \begin{tabular}[t]{@{}l@{}}70.0 $\pm$ 14.4\\ {\scriptsize [40.0, 100.0]}\\ {\scriptsize $n=10$}\end{tabular} \\
Mistral-Small-4-119B & \begin{tabular}[t]{@{}l@{}}72.2 $\pm$ 11.0\\ {\scriptsize [50.0, 90.0]}\\ {\scriptsize $n=18$}\end{tabular} & \begin{tabular}[t]{@{}l@{}}0.0 $\pm$ 0.0 [0.0, 0.0]\\ {\scriptsize $n=6$}\end{tabular} & \begin{tabular}[t]{@{}l@{}}0.0 $\pm$ 0.0 [0.0, 0.0]\\ {\scriptsize $n=1$}\end{tabular} \\
MedGemma-1.5-4B & \begin{tabular}[t]{@{}l@{}}81.1 $\pm$ 3.0 [75.0, 86.7]\\ {\scriptsize $n=254$}\end{tabular} & \begin{tabular}[t]{@{}l@{}}94.5 $\pm$ 2.2 [90.1, 98.2]\\ {\scriptsize $n=109$}\end{tabular} & \begin{tabular}[t]{@{}l@{}}70.0 $\pm$ 14.4\\ {\scriptsize [40.0, 90.3]}\\ {\scriptsize $n=10$}\end{tabular} \\
LLaVA-Med-7B & \begin{tabular}[t]{@{}l@{}}100.0 $\pm$ 0.0\\ {\scriptsize [100.0, 100.0]}\\ {\scriptsize $n=322$}\end{tabular} & \begin{tabular}[t]{@{}l@{}}100.0 $\pm$ 0.0\\ {\scriptsize [100.0, 100.0]}\\ {\scriptsize $n=114$}\end{tabular} & \begin{tabular}[t]{@{}l@{}}100.0 $\pm$ 0.0\\ {\scriptsize [100.0, 100.0]}\\ {\scriptsize $n=11$}\end{tabular} \\
MedGemma-27B-text & \begin{tabular}[t]{@{}l@{}}100.0 $\pm$ 0.0\\ {\scriptsize [100.0, 100.0]}\\ {\scriptsize $n=289$}\end{tabular} & \begin{tabular}[t]{@{}l@{}}100.0 $\pm$ 0.0\\ {\scriptsize [100.0, 100.0]}\\ {\scriptsize $n=115$}\end{tabular} & \begin{tabular}[t]{@{}l@{}}100.0 $\pm$ 0.0\\ {\scriptsize [100.0, 100.0]}\\ {\scriptsize $n=11$}\end{tabular} \\
DeepSeek-R1-7B & \begin{tabular}[t]{@{}l@{}}100.0 $\pm$ 0.0\\ {\scriptsize [100.0, 100.0]}\\ {\scriptsize $n=38$}\end{tabular} & \begin{tabular}[t]{@{}l@{}}100.0 $\pm$ 0.0\\ {\scriptsize [100.0, 100.0]}\\ {\scriptsize $n=96$}\end{tabular} & \begin{tabular}[t]{@{}l@{}}100.0 $\pm$ 0.0\\ {\scriptsize [100.0, 100.0]}\\ {\scriptsize $n=7$}\end{tabular} \\
RAD-DINO & \begin{tabular}[t]{@{}l@{}}96.7 $\pm$ 1.0 [94.5, 98.6]\\ {\scriptsize $n=302$}\end{tabular} & \begin{tabular}[t]{@{}l@{}}98.3 $\pm$ 1.1\\ {\scriptsize [95.7, 100.0]}\\ {\scriptsize $n=117$}\end{tabular} & \begin{tabular}[t]{@{}l@{}}100.0 $\pm$ 0.0\\ {\scriptsize [100.0, 100.0]}\\ {\scriptsize $n=11$}\end{tabular} \\
\bottomrule
\end{tabular}
\end{table}

\begin{table}[t]
\centering
\caption{The causal grounding rate per finding on the MS-CXR block ($n$ = 452 cases), for the image users and the unstable system. Each value is the share of correct affirmative answers changed under the target mask, as the mean over its cases $\pm$ the standard deviation of 1{,}000 patient-cluster bootstrap resamples, with the percentile 95\% interval and the case count $n$. A cell of fewer than 10 cases is not interpreted, and N/A marks a finding with no evaluable case. The systems that ignore the image have a grounding rate of 0.0 on every finding and are omitted.}
\label{stab:cgr_finding}
\setlength{\tabcolsep}{2pt}
\renewcommand{\arraystretch}{1.05}
\scriptsize
\begin{tabular}{@{}p{0.14\textwidth}p{0.15\textwidth}p{0.15\textwidth}p{0.15\textwidth}p{0.15\textwidth}p{0.15\textwidth}@{}}
\toprule
Finding & Gemma-4-26B & Qwen3-VL-32B & MedGemma-1.5-4B & RAD-DINO & Mistral-Small-4-119B \\
\midrule
Atelectasis & \begin{tabular}[t]{@{}l@{}}27.3 $\pm$ 8.2\\ {\scriptsize [12.8, 44.8]}\\ {\scriptsize $n=33$}\end{tabular} & \begin{tabular}[t]{@{}l@{}}0.0 $\pm$ 0.0\\ {\scriptsize [0.0, 0.0]}\\ {\scriptsize $n=35$}\end{tabular} & \begin{tabular}[t]{@{}l@{}}0.0 $\pm$ 0.0\\ {\scriptsize [0.0, 0.0]}\\ {\scriptsize $n=35$}\end{tabular} & \begin{tabular}[t]{@{}l@{}}0.0 $\pm$ 0.0\\ {\scriptsize [0.0, 0.0]}\\ {\scriptsize $n=35$}\end{tabular} & N/A \\
Cardiomegaly & \begin{tabular}[t]{@{}l@{}}48.0 $\pm$ 5.1\\ {\scriptsize [37.9, 57.4]}\\ {\scriptsize $n=98$}\end{tabular} & \begin{tabular}[t]{@{}l@{}}3.0 $\pm$ 1.7\\ {\scriptsize [0.0, 6.9]}\\ {\scriptsize $n=99$}\end{tabular} & \begin{tabular}[t]{@{}l@{}}35.1 $\pm$ 5.0\\ {\scriptsize [25.5, 44.8]}\\ {\scriptsize $n=97$}\end{tabular} & \begin{tabular}[t]{@{}l@{}}1.0 $\pm$ 1.0\\ {\scriptsize [0.0, 3.1]}\\ {\scriptsize $n=100$}\end{tabular} & \begin{tabular}[t]{@{}l@{}}100.0 $\pm$ 0.0\\ {\scriptsize [100.0, 100.0]}\\ {\scriptsize $n=5$}\end{tabular} \\
Consolidation & \begin{tabular}[t]{@{}l@{}}9.2 $\pm$ 3.3\\ {\scriptsize [3.4, 16.0]}\\ {\scriptsize $n=76$}\end{tabular} & \begin{tabular}[t]{@{}l@{}}32.7 $\pm$ 6.1\\ {\scriptsize [20.8, 44.6]}\\ {\scriptsize $n=55$}\end{tabular} & \begin{tabular}[t]{@{}l@{}}37.7 $\pm$ 6.5\\ {\scriptsize [25.0, 50.8]}\\ {\scriptsize $n=69$}\end{tabular} & \begin{tabular}[t]{@{}l@{}}1.4 $\pm$ 1.3\\ {\scriptsize [0.0, 4.3]}\\ {\scriptsize $n=74$}\end{tabular} & \begin{tabular}[t]{@{}l@{}}44.4 $\pm$ 14.5\\ {\scriptsize [20.0, 75.0]}\\ {\scriptsize $n=9$}\end{tabular} \\
Edema & \begin{tabular}[t]{@{}l@{}}50.0 $\pm$ 7.6\\ {\scriptsize [35.9, 64.3]}\\ {\scriptsize $n=42$}\end{tabular} & \begin{tabular}[t]{@{}l@{}}29.6 $\pm$ 8.8\\ {\scriptsize [14.3, 48.1]}\\ {\scriptsize $n=27$}\end{tabular} & \begin{tabular}[t]{@{}l@{}}69.7 $\pm$ 9.7\\ {\scriptsize [50.0, 85.7]}\\ {\scriptsize $n=33$}\end{tabular} & \begin{tabular}[t]{@{}l@{}}4.9 $\pm$ 3.3\\ {\scriptsize [0.0, 12.5]}\\ {\scriptsize $n=41$}\end{tabular} & N/A \\
Lung opacity & \begin{tabular}[t]{@{}l@{}}0.0 $\pm$ 0.0\\ {\scriptsize [0.0, 0.0]}\\ {\scriptsize $n=28$}\end{tabular} & \begin{tabular}[t]{@{}l@{}}4.0 $\pm$ 3.7\\ {\scriptsize [0.0, 12.0]}\\ {\scriptsize $n=25$}\end{tabular} & \begin{tabular}[t]{@{}l@{}}0.0 $\pm$ 0.0\\ {\scriptsize [0.0, 0.0]}\\ {\scriptsize $n=30$}\end{tabular} & \begin{tabular}[t]{@{}l@{}}0.0 $\pm$ 0.0\\ {\scriptsize [0.0, 0.0]}\\ {\scriptsize $n=30$}\end{tabular} & \begin{tabular}[t]{@{}l@{}}9.1 $\pm$ 9.4\\ {\scriptsize [0.0, 33.3]}\\ {\scriptsize $n=11$}\end{tabular} \\
Pleural effusion & \begin{tabular}[t]{@{}l@{}}5.9 $\pm$ 4.1\\ {\scriptsize [0.0, 15.6]}\\ {\scriptsize $n=34$}\end{tabular} & \begin{tabular}[t]{@{}l@{}}25.0 $\pm$ 7.3\\ {\scriptsize [11.1, 40.7]}\\ {\scriptsize $n=28$}\end{tabular} & \begin{tabular}[t]{@{}l@{}}8.8 $\pm$ 5.3\\ {\scriptsize [0.0, 21.9]}\\ {\scriptsize $n=34$}\end{tabular} & \begin{tabular}[t]{@{}l@{}}2.9 $\pm$ 3.0\\ {\scriptsize [0.0, 10.0]}\\ {\scriptsize $n=34$}\end{tabular} & N/A \\
Pneumonia & \begin{tabular}[t]{@{}l@{}}46.2 $\pm$ 6.8\\ {\scriptsize [32.5, 60.0]}\\ {\scriptsize $n=39$}\end{tabular} & \begin{tabular}[t]{@{}l@{}}34.5 $\pm$ 5.6\\ {\scriptsize [23.6, 46.3]}\\ {\scriptsize $n=55$}\end{tabular} & \begin{tabular}[t]{@{}l@{}}53.1 $\pm$ 6.8\\ {\scriptsize [40.0, 66.7]}\\ {\scriptsize $n=64$}\end{tabular} & \begin{tabular}[t]{@{}l@{}}17.4 $\pm$ 4.1\\ {\scriptsize [9.7, 25.8]}\\ {\scriptsize $n=92$}\end{tabular} & N/A \\
Pneumothorax & \begin{tabular}[t]{@{}l@{}}50.0 $\pm$ 35.8\\ {\scriptsize [0.0, 100.0]}\\ {\scriptsize $n=2$}\end{tabular} & \begin{tabular}[t]{@{}l@{}}100.0 $\pm$ 0.0\\ {\scriptsize [100.0, 100.0]}\\ {\scriptsize $n=1$}\end{tabular} & \begin{tabular}[t]{@{}l@{}}45.5 $\pm$ 17.2\\ {\scriptsize [10.0, 80.0]}\\ {\scriptsize $n=11$}\end{tabular} & \begin{tabular}[t]{@{}l@{}}25.0 $\pm$ 8.8\\ {\scriptsize [8.7, 42.3]}\\ {\scriptsize $n=24$}\end{tabular} & N/A \\
\bottomrule
\end{tabular}
\end{table}

\begin{table}[t]
\centering
\caption{Sex-disaggregated accuracy and intervention metrics per system on the MIMIC question set ($n$ = 2{,}548 cases). F and M are female and male patients. The denominators are all cases for accuracy, the correct MS-CXR answers for CGR, all correct-on-original answers for UAR, and the correct-on-original finding-presence answers for OFR. No $p$-value is given where every case of both groups has the same outcome. Each rate is the mean over its cases $\pm$ the standard deviation of 1{,}000 patient-cluster bootstrap resamples, with the 95\% interval and the case count $n$. The $p$-value is the FDR-corrected permutation test of the difference between the sexes within each model's family. CGR, causal grounding rate; FDR, false discovery rate; OFR, opposite-label flip rate; UAR, unrelated-image answer rate.}
\label{stab:sex}
\setlength{\tabcolsep}{2pt}
\renewcommand{\arraystretch}{1.0}
\scriptsize
\begin{tabular}{@{}p{0.12\textwidth}p{0.195\textwidth}p{0.195\textwidth}p{0.195\textwidth}p{0.195\textwidth}@{}}
\toprule
Model & Accuracy & CGR & UAR & OFR \\
\midrule
Gemma-4-26B & \begin{tabular}[t]{@{}l@{}}F 67.9 $\pm$ 1.4\\ {\scriptsize [65.1, 70.6], $n=1{,}170$}\\ M 69.6 $\pm$ 1.4\\ {\scriptsize [67.0, 72.1], $n=1{,}321$}\end{tabular} & \begin{tabular}[t]{@{}l@{}}F 36.7 $\pm$ 4.6\\ {\scriptsize [28.4, 46.3], $n=139$}\\ M 25.4 $\pm$ 3.3\\ {\scriptsize [19.5, 32.0], $n=213$}\\ $p_{\mathrm{FDR}}$ = 0.040\end{tabular} & \begin{tabular}[t]{@{}l@{}}F 79.8 $\pm$ 1.6\\ {\scriptsize [76.7, 82.7], $n=783$}\\ M 82.2 $\pm$ 1.4\\ {\scriptsize [79.7, 84.9], $n=906$}\\ $p_{\mathrm{FDR}}$ = 0.284\end{tabular} & \begin{tabular}[t]{@{}l@{}}F 54.0 $\pm$ 2.1\\ {\scriptsize [49.7, 58.1], $n=554$}\\ M 53.3 $\pm$ 2.0\\ {\scriptsize [49.5, 57.1], $n=646$}\\ $p_{\mathrm{FDR}}$ = 0.803\end{tabular} \\
Qwen3-VL-32B & \begin{tabular}[t]{@{}l@{}}F 64.9 $\pm$ 1.5\\ {\scriptsize [61.7, 67.7], $n=1{,}201$}\\ M 65.1 $\pm$ 1.4\\ {\scriptsize [62.3, 67.6], $n=1{,}347$}\end{tabular} & \begin{tabular}[t]{@{}l@{}}F 17.2 $\pm$ 3.2\\ {\scriptsize [11.0, 23.6], $n=134$}\\ M 17.8 $\pm$ 2.7\\ {\scriptsize [12.9, 23.4], $n=191$}\\ $p_{\mathrm{FDR}}$ = 0.877\end{tabular} & \begin{tabular}[t]{@{}l@{}}F 79.8 $\pm$ 1.5\\ {\scriptsize [76.8, 82.6], $n=779$}\\ M 77.1 $\pm$ 1.4\\ {\scriptsize [74.4, 80.0], $n=877$}\\ $p_{\mathrm{FDR}}$ = 0.301\end{tabular} & \begin{tabular}[t]{@{}l@{}}F 44.6 $\pm$ 2.2\\ {\scriptsize [40.3, 48.8], $n=534$}\\ M 51.8 $\pm$ 2.1\\ {\scriptsize [47.8, 55.9], $n=616$}\\ $p_{\mathrm{FDR}}$ = 0.180\end{tabular} \\
Mistral-Small-4-119B & \begin{tabular}[t]{@{}l@{}}F 47.5 $\pm$ 1.6\\ {\scriptsize [44.2, 50.5], $n=1{,}201$}\\ M 45.8 $\pm$ 1.5\\ {\scriptsize [43.0, 48.6], $n=1{,}347$}\end{tabular} & \begin{tabular}[t]{@{}l@{}}F 50.0 $\pm$ 17.6\\ {\scriptsize [20.0, 85.7], $n=8$}\\ M 35.3 $\pm$ 12.2\\ {\scriptsize [12.5, 60.0], $n=17$}\\ $p_{\mathrm{FDR}}$ = 0.863\end{tabular} & \begin{tabular}[t]{@{}l@{}}F 87.0 $\pm$ 1.4\\ {\scriptsize [84.2, 89.7], $n=570$}\\ M 84.4 $\pm$ 1.5\\ {\scriptsize [81.5, 87.3], $n=617$}\\ $p_{\mathrm{FDR}}$ = 0.486\end{tabular} & \begin{tabular}[t]{@{}l@{}}F 15.0 $\pm$ 1.8\\ {\scriptsize [11.3, 18.5], $n=380$}\\ M 15.0 $\pm$ 1.8\\ {\scriptsize [11.5, 18.7], $n=420$}\\ $p_{\mathrm{FDR}} > 0.999$\end{tabular} \\
MedGemma-1.5-4B & \begin{tabular}[t]{@{}l@{}}F 59.4 $\pm$ 1.7\\ {\scriptsize [56.1, 62.9], $n=1{,}201$}\\ M 57.2 $\pm$ 1.6\\ {\scriptsize [54.1, 60.3], $n=1{,}347$}\end{tabular} & \begin{tabular}[t]{@{}l@{}}F 39.5 $\pm$ 4.8\\ {\scriptsize [29.9, 48.8], $n=157$}\\ M 29.2 $\pm$ 3.5\\ {\scriptsize [22.7, 36.2], $n=216$}\\ $p_{\mathrm{FDR}}$ = 0.256\end{tabular} & \begin{tabular}[t]{@{}l@{}}F 75.5 $\pm$ 1.7\\ {\scriptsize [72.2, 78.7], $n=713$}\\ M 78.8 $\pm$ 1.5\\ {\scriptsize [76.0, 81.7], $n=770$}\\ $p_{\mathrm{FDR}}$ = 0.297\end{tabular} & \begin{tabular}[t]{@{}l@{}}F 55.9 $\pm$ 2.0\\ {\scriptsize [51.9, 59.9], $n=599$}\\ M 51.2 $\pm$ 1.9\\ {\scriptsize [47.1, 54.9], $n=650$}\\ $p_{\mathrm{FDR}}$ = 0.297\end{tabular} \\
LLaVA-Med-7B & \begin{tabular}[t]{@{}l@{}}F 47.1 $\pm$ 1.8\\ {\scriptsize [43.4, 50.7], $n=1{,}180$}\\ M 48.6 $\pm$ 1.7\\ {\scriptsize [45.6, 52.2], $n=1{,}333$}\end{tabular} & \begin{tabular}[t]{@{}l@{}}F 0.0 $\pm$ 0.0\\ {\scriptsize [0.0, 0.0], $n=184$}\\ M 0.0 $\pm$ 0.0\\ {\scriptsize [0.0, 0.0], $n=263$}\end{tabular} & \begin{tabular}[t]{@{}l@{}}F 100.0 $\pm$ 0.0\\ {\scriptsize [100.0, 100.0], $n=547$}\\ M 100.0 $\pm$ 0.0\\ {\scriptsize [100.0, 100.0], $n=641$}\end{tabular} & \begin{tabular}[t]{@{}l@{}}F 0.0 $\pm$ 0.0\\ {\scriptsize [0.0, 0.0], $n=486$}\\ M 0.0 $\pm$ 0.0\\ {\scriptsize [0.0, 0.0], $n=584$}\end{tabular} \\
MedGemma-27B-text & \begin{tabular}[t]{@{}l@{}}F 54.7 $\pm$ 1.7\\ {\scriptsize [51.4, 58.1], $n=1{,}081$}\\ M 55.8 $\pm$ 1.5\\ {\scriptsize [52.8, 58.6], $n=1{,}216$}\end{tabular} & \begin{tabular}[t]{@{}l@{}}F 0.0 $\pm$ 0.0\\ {\scriptsize [0.0, 0.0], $n=172$}\\ M 0.0 $\pm$ 0.0\\ {\scriptsize [0.0, 0.0], $n=243$}\end{tabular} & \begin{tabular}[t]{@{}l@{}}F 100.0 $\pm$ 0.0\\ {\scriptsize [100.0, 100.0], $n=591$}\\ M 100.0 $\pm$ 0.0\\ {\scriptsize [100.0, 100.0], $n=679$}\end{tabular} & \begin{tabular}[t]{@{}l@{}}F 0.0 $\pm$ 0.0\\ {\scriptsize [0.0, 0.0], $n=487$}\\ M 0.0 $\pm$ 0.0\\ {\scriptsize [0.0, 0.0], $n=578$}\end{tabular} \\
DeepSeek-R1-7B & \begin{tabular}[t]{@{}l@{}}F 43.8 $\pm$ 1.5\\ {\scriptsize [40.9, 46.8], $n=1{,}135$}\\ M 37.9 $\pm$ 1.4\\ {\scriptsize [35.3, 40.8], $n=1{,}257$}\end{tabular} & \begin{tabular}[t]{@{}l@{}}F 0.0 $\pm$ 0.0\\ {\scriptsize [0.0, 0.0], $n=64$}\\ M 0.0 $\pm$ 0.0\\ {\scriptsize [0.0, 0.0], $n=77$}\end{tabular} & \begin{tabular}[t]{@{}l@{}}F 100.0 $\pm$ 0.0\\ {\scriptsize [100.0, 100.0], $n=497$}\\ M 100.0 $\pm$ 0.0\\ {\scriptsize [100.0, 100.0], $n=477$}\end{tabular} & \begin{tabular}[t]{@{}l@{}}F 0.0 $\pm$ 0.0\\ {\scriptsize [0.0, 0.0], $n=369$}\\ M 0.0 $\pm$ 0.0\\ {\scriptsize [0.0, 0.0], $n=362$}\end{tabular} \\
RAD-DINO & \begin{tabular}[t]{@{}l@{}}F 71.5 $\pm$ 1.6\\ {\scriptsize [68.4, 74.5], $n=866$}\\ M 72.8 $\pm$ 1.5\\ {\scriptsize [69.7, 75.7], $n=986$}\end{tabular} & \begin{tabular}[t]{@{}l@{}}F 7.9 $\pm$ 2.1\\ {\scriptsize [4.0, 12.2], $n=178$}\\ M 5.2 $\pm$ 1.4\\ {\scriptsize [2.4, 8.2], $n=252$}\\ $p_{\mathrm{FDR}}$ = 0.362\end{tabular} & \begin{tabular}[t]{@{}l@{}}F 80.5 $\pm$ 1.7\\ {\scriptsize [77.0, 83.8], $n=619$}\\ M 83.8 $\pm$ 1.4\\ {\scriptsize [81.2, 86.6], $n=718$}\\ $p_{\mathrm{FDR}}$ = 0.190\end{tabular} & \begin{tabular}[t]{@{}l@{}}F 63.0 $\pm$ 2.1\\ {\scriptsize [58.7, 67.0], $n=619$}\\ M 56.5 $\pm$ 1.9\\ {\scriptsize [53.0, 60.3], $n=718$}\\ $p_{\mathrm{FDR}}$ = 0.047\end{tabular} \\
\bottomrule
\end{tabular}
\end{table}

\begin{table}[t]
\centering
\caption{Sex-disaggregated accuracy and intervention metrics per system on the CheXpert question set ($n$ = 1{,}380 cases). F and M are female and male patients. The denominators are all cases for accuracy and the correct-on-original answers for UAR and OFR. CGR is N/A because CheXpert has no boxes. No $p$-value is given where every case of both groups has the same outcome. Each rate is the mean over its cases $\pm$ the standard deviation of 1{,}000 patient-cluster bootstrap resamples, with the 95\% interval and the case count $n$. The $p$-value is the FDR-corrected permutation test of the difference between the sexes within each model's family. CGR, causal grounding rate; FDR, false discovery rate; N/A, not applicable; OFR, opposite-label flip rate; UAR, unrelated-image answer rate.}
\label{stab:sex_chexpert}
\setlength{\tabcolsep}{2pt}
\renewcommand{\arraystretch}{1.0}
\scriptsize
\begin{tabular}{@{}p{0.12\textwidth}p{0.195\textwidth}p{0.195\textwidth}p{0.195\textwidth}p{0.195\textwidth}@{}}
\toprule
Model & Accuracy & CGR & UAR & OFR \\
\midrule
Gemma-4-26B & \begin{tabular}[t]{@{}l@{}}F 70.5 $\pm$ 1.9\\ {\scriptsize [66.7, 74.1], $n=567$}\\ M 66.6 $\pm$ 1.7\\ {\scriptsize [63.5, 69.8], $n=788$}\end{tabular} & N/A & \begin{tabular}[t]{@{}l@{}}F 78.7 $\pm$ 2.1\\ {\scriptsize [74.6, 82.7], $n=395$}\\ M 73.2 $\pm$ 1.8\\ {\scriptsize [69.6, 76.7], $n=515$}\\ $p_{\mathrm{FDR}}$ = 0.228\end{tabular} & \begin{tabular}[t]{@{}l@{}}F 59.5 $\pm$ 2.5\\ {\scriptsize [54.8, 64.5], $n=395$}\\ M 59.6 $\pm$ 2.3\\ {\scriptsize [55.1, 64.1], $n=510$}\\ $p_{\mathrm{FDR}} > 0.999$\end{tabular} \\
Qwen3-VL-32B & \begin{tabular}[t]{@{}l@{}}F 61.7 $\pm$ 2.0\\ {\scriptsize [57.8, 65.3], $n=582$}\\ M 61.4 $\pm$ 1.7\\ {\scriptsize [57.9, 64.5], $n=798$}\end{tabular} & N/A & \begin{tabular}[t]{@{}l@{}}F 75.5 $\pm$ 2.2\\ {\scriptsize [70.9, 79.8], $n=359$}\\ M 77.6 $\pm$ 1.9\\ {\scriptsize [73.7, 81.3], $n=490$}\\ $p_{\mathrm{FDR}}$ = 0.716\end{tabular} & \begin{tabular}[t]{@{}l@{}}F 50.7 $\pm$ 2.5\\ {\scriptsize [46.1, 55.8], $n=359$}\\ M 48.8 $\pm$ 2.2\\ {\scriptsize [44.7, 53.1], $n=490$}\\ $p_{\mathrm{FDR}}$ = 0.716\end{tabular} \\
Mistral-Small-4-119B & \begin{tabular}[t]{@{}l@{}}F 52.6 $\pm$ 2.1\\ {\scriptsize [48.5, 56.6], $n=582$}\\ M 53.8 $\pm$ 1.8\\ {\scriptsize [50.3, 57.3], $n=798$}\end{tabular} & N/A & \begin{tabular}[t]{@{}l@{}}F 92.5 $\pm$ 1.6\\ {\scriptsize [89.4, 95.7], $n=306$}\\ M 91.8 $\pm$ 1.3\\ {\scriptsize [89.2, 94.1], $n=429$}\\ $p_{\mathrm{FDR}}$ = 0.780\end{tabular} & \begin{tabular}[t]{@{}l@{}}F 10.5 $\pm$ 1.7\\ {\scriptsize [7.3, 13.8], $n=306$}\\ M 7.5 $\pm$ 1.2\\ {\scriptsize [5.2, 9.9], $n=429$}\\ $p_{\mathrm{FDR}}$ = 0.227\end{tabular} \\
MedGemma-1.5-4B & \begin{tabular}[t]{@{}l@{}}F 66.8 $\pm$ 2.0\\ {\scriptsize [62.9, 70.3], $n=582$}\\ M 65.3 $\pm$ 1.7\\ {\scriptsize [62.0, 68.6], $n=798$}\end{tabular} & N/A & \begin{tabular}[t]{@{}l@{}}F 77.6 $\pm$ 2.1\\ {\scriptsize [73.4, 81.8], $n=389$}\\ M 80.6 $\pm$ 1.7\\ {\scriptsize [77.4, 84.1], $n=521$}\\ $p_{\mathrm{FDR}}$ = 0.569\end{tabular} & \begin{tabular}[t]{@{}l@{}}F 52.4 $\pm$ 2.4\\ {\scriptsize [47.6, 57.0], $n=389$}\\ M 50.7 $\pm$ 2.2\\ {\scriptsize [46.5, 55.0], $n=521$}\\ $p_{\mathrm{FDR}}$ = 0.777\end{tabular} \\
LLaVA-Med-7B & \begin{tabular}[t]{@{}l@{}}F 48.9 $\pm$ 2.1\\ {\scriptsize [44.8, 52.9], $n=579$}\\ M 46.0 $\pm$ 1.8\\ {\scriptsize [42.7, 49.6], $n=795$}\end{tabular} & N/A & \begin{tabular}[t]{@{}l@{}}F 100.0 $\pm$ 0.0\\ {\scriptsize [100.0, 100.0], $n=283$}\\ M 100.0 $\pm$ 0.0\\ {\scriptsize [100.0, 100.0], $n=365$}\end{tabular} & \begin{tabular}[t]{@{}l@{}}F 0.0 $\pm$ 0.0\\ {\scriptsize [0.0, 0.0], $n=280$}\\ M 0.0 $\pm$ 0.0\\ {\scriptsize [0.0, 0.0], $n=366$}\end{tabular} \\
MedGemma-27B-text & \begin{tabular}[t]{@{}l@{}}F 48.3 $\pm$ 2.1\\ {\scriptsize [44.2, 52.7], $n=582$}\\ M 46.2 $\pm$ 1.9\\ {\scriptsize [42.8, 50.2], $n=798$}\end{tabular} & N/A & \begin{tabular}[t]{@{}l@{}}F 100.0 $\pm$ 0.0\\ {\scriptsize [100.0, 100.0], $n=281$}\\ M 100.0 $\pm$ 0.0\\ {\scriptsize [100.0, 100.0], $n=369$}\end{tabular} & \begin{tabular}[t]{@{}l@{}}F 0.0 $\pm$ 0.0\\ {\scriptsize [0.0, 0.0], $n=281$}\\ M 0.0 $\pm$ 0.0\\ {\scriptsize [0.0, 0.0], $n=369$}\end{tabular} \\
DeepSeek-R1-7B & \begin{tabular}[t]{@{}l@{}}F 46.5 $\pm$ 2.1\\ {\scriptsize [42.4, 50.7], $n=527$}\\ M 47.7 $\pm$ 1.8\\ {\scriptsize [44.2, 50.9], $n=753$}\end{tabular} & N/A & \begin{tabular}[t]{@{}l@{}}F 100.0 $\pm$ 0.0\\ {\scriptsize [100.0, 100.0], $n=245$}\\ M 100.0 $\pm$ 0.0\\ {\scriptsize [100.0, 100.0], $n=359$}\end{tabular} & \begin{tabular}[t]{@{}l@{}}F 0.0 $\pm$ 0.0\\ {\scriptsize [0.0, 0.0], $n=245$}\\ M 0.0 $\pm$ 0.0\\ {\scriptsize [0.0, 0.0], $n=359$}\end{tabular} \\
RAD-DINO & \begin{tabular}[t]{@{}l@{}}F 69.4 $\pm$ 1.9\\ {\scriptsize [65.5, 72.9], $n=582$}\\ M 66.9 $\pm$ 1.8\\ {\scriptsize [63.4, 70.5], $n=798$}\end{tabular} & N/A & \begin{tabular}[t]{@{}l@{}}F 83.7 $\pm$ 1.7\\ {\scriptsize [80.0, 86.9], $n=404$}\\ M 82.4 $\pm$ 1.6\\ {\scriptsize [79.4, 85.4], $n=534$}\\ $p_{\mathrm{FDR}}$ = 0.784\end{tabular} & \begin{tabular}[t]{@{}l@{}}F 68.6 $\pm$ 2.3\\ {\scriptsize [64.1, 73.3], $n=404$}\\ M 68.5 $\pm$ 2.0\\ {\scriptsize [64.5, 72.5], $n=534$}\\ $p_{\mathrm{FDR}} > 0.999$\end{tabular} \\
\bottomrule
\end{tabular}
\end{table}

\begin{table}[t]
\centering
\caption{Main metrics on the CheXpert question set ($n$ = 1{,}380 cases). Category is assigned by the swap rule, applied unchanged. Accuracy and balanced accuracy are computed on the pooled CheXpert cases, all of which are finding-presence questions. The unrelated-image answer rate, the opposite-label flip rate, and the swap specificity premium are computed as on MIMIC. The heads of the vision-only reference are fit on the training and validation images of the CheXpert list. CheXpert is therefore in distribution for this reference and for no other system. Each value is the mean over its cases $\pm$ the standard deviation of 1{,}000 patient-cluster bootstrap resamples, with the percentile 95\% confidence interval and the case count $n$. UAR, unrelated-image answer rate; OFR, opposite-label flip rate; SSP, swap specificity premium. N/A, not applicable.}
\label{stab:chexpert}
\setlength{\tabcolsep}{3pt}
\renewcommand{\arraystretch}{1.05}
\scriptsize
\begin{tabular}{@{}p{0.14\textwidth}p{0.08\textwidth}p{0.13\textwidth}p{0.13\textwidth}p{0.13\textwidth}p{0.13\textwidth}p{0.13\textwidth}@{}}
\toprule
Model & Category & Accuracy & Balanced accuracy & UAR & OFR & SSP \\
\midrule
\multicolumn{7}{@{}l}{\textbf{General-purpose multimodal}} \\
Gemma-4-26B & Uses image & \begin{tabular}[t]{@{}l@{}}68.3 $\pm$ 1.3\\ {\scriptsize [65.8, 70.8]}\\ {\scriptsize $n=1{,}355$}\end{tabular} & \begin{tabular}[t]{@{}l@{}}67.9 $\pm$ 1.3\\ {\scriptsize [65.5, 70.4]}\\ {\scriptsize $n=1{,}355$}\end{tabular} & \begin{tabular}[t]{@{}l@{}}75.6 $\pm$ 1.4\\ {\scriptsize [72.8, 78.3]}\\ {\scriptsize $n=910$}\end{tabular} & \begin{tabular}[t]{@{}l@{}}59.6 $\pm$ 1.5\\ {\scriptsize [56.7, 62.5]}\\ {\scriptsize $n=905$}\end{tabular} & \begin{tabular}[t]{@{}l@{}}$+34.7$ $\pm$ 2.1\\ {\scriptsize [+30.8, +38.7]}\\ {\scriptsize $n=895$}\end{tabular} \\
Qwen3-VL-32B & Uses image & \begin{tabular}[t]{@{}l@{}}61.5 $\pm$ 1.3\\ {\scriptsize [58.9, 63.9]}\\ {\scriptsize $n=1{,}380$}\end{tabular} & \begin{tabular}[t]{@{}l@{}}60.6 $\pm$ 1.2\\ {\scriptsize [58.2, 63.0]}\\ {\scriptsize $n=1{,}380$}\end{tabular} & \begin{tabular}[t]{@{}l@{}}76.7 $\pm$ 1.4\\ {\scriptsize [73.9, 79.4]}\\ {\scriptsize $n=849$}\end{tabular} & \begin{tabular}[t]{@{}l@{}}49.6 $\pm$ 1.7\\ {\scriptsize [46.4, 52.9]}\\ {\scriptsize $n=849$}\end{tabular} & \begin{tabular}[t]{@{}l@{}}$+26.3$ $\pm$ 1.9\\ {\scriptsize [+22.4, +29.9]}\\ {\scriptsize $n=849$}\end{tabular} \\
Mistral-Small-4-119B & Unstable & \begin{tabular}[t]{@{}l@{}}53.3 $\pm$ 1.3\\ {\scriptsize [50.5, 55.9]}\\ {\scriptsize $n=1{,}380$}\end{tabular} & \begin{tabular}[t]{@{}l@{}}50.8 $\pm$ 0.7\\ {\scriptsize [49.5, 52.0]}\\ {\scriptsize $n=1{,}380$}\end{tabular} & \begin{tabular}[t]{@{}l@{}}92.1 $\pm$ 1.0\\ {\scriptsize [90.1, 94.0]}\\ {\scriptsize $n=735$}\end{tabular} & \begin{tabular}[t]{@{}l@{}}8.7 $\pm$ 1.0\\ {\scriptsize [6.8, 10.7]}\\ {\scriptsize $n=735$}\end{tabular} & \begin{tabular}[t]{@{}l@{}}$+0.8$ $\pm$ 1.0\\ {\scriptsize [-1.2, +2.9]}\\ {\scriptsize $n=735$}\end{tabular} \\
\midrule
\multicolumn{7}{@{}l}{\textbf{Medical multimodal}} \\
MedGemma-1.5-4B & Uses image & \begin{tabular}[t]{@{}l@{}}65.9 $\pm$ 1.3\\ {\scriptsize [63.5, 68.5]}\\ {\scriptsize $n=1{,}380$}\end{tabular} & \begin{tabular}[t]{@{}l@{}}66.0 $\pm$ 1.3\\ {\scriptsize [63.5, 68.5]}\\ {\scriptsize $n=1{,}380$}\end{tabular} & \begin{tabular}[t]{@{}l@{}}79.3 $\pm$ 1.4\\ {\scriptsize [76.8, 82.0]}\\ {\scriptsize $n=910$}\end{tabular} & \begin{tabular}[t]{@{}l@{}}51.4 $\pm$ 1.7\\ {\scriptsize [48.2, 54.7]}\\ {\scriptsize $n=910$}\end{tabular} & \begin{tabular}[t]{@{}l@{}}$+30.8$ $\pm$ 1.9\\ {\scriptsize [+27.2, +34.5]}\\ {\scriptsize $n=910$}\end{tabular} \\
LLaVA-Med-7B & Ignores image & \begin{tabular}[t]{@{}l@{}}47.2 $\pm$ 1.3\\ {\scriptsize [44.6, 49.8]}\\ {\scriptsize $n=1{,}374$}\end{tabular} & \begin{tabular}[t]{@{}l@{}}50.0 $\pm$ 0.0\\ {\scriptsize [50.0, 50.0]}\\ {\scriptsize $n=1{,}374$}\end{tabular} & \begin{tabular}[t]{@{}l@{}}100.0 $\pm$ 0.0\\ {\scriptsize [100.0, 100.0]}\\ {\scriptsize $n=648$}\end{tabular} & \begin{tabular}[t]{@{}l@{}}0.0 $\pm$ 0.0\\ {\scriptsize [0.0, 0.0]}\\ {\scriptsize $n=646$}\end{tabular} & \begin{tabular}[t]{@{}l@{}}$+0.0$ $\pm$ 0.0\\ {\scriptsize [+0.0, +0.0]}\\ {\scriptsize $n=645$}\end{tabular} \\
\midrule
\multicolumn{7}{@{}l}{\textbf{Text-only controls}} \\
MedGemma-27B-text & Ignores image & \begin{tabular}[t]{@{}l@{}}47.1 $\pm$ 1.4\\ {\scriptsize [44.5, 49.7]}\\ {\scriptsize $n=1{,}380$}\end{tabular} & \begin{tabular}[t]{@{}l@{}}49.6 $\pm$ 0.7\\ {\scriptsize [48.3, 51.0]}\\ {\scriptsize $n=1{,}380$}\end{tabular} & \begin{tabular}[t]{@{}l@{}}100.0 $\pm$ 0.0\\ {\scriptsize [100.0, 100.0]}\\ {\scriptsize $n=650$}\end{tabular} & \begin{tabular}[t]{@{}l@{}}0.0 $\pm$ 0.0\\ {\scriptsize [0.0, 0.0]}\\ {\scriptsize $n=650$}\end{tabular} & \begin{tabular}[t]{@{}l@{}}$+0.0$ $\pm$ 0.0\\ {\scriptsize [+0.0, +0.0]}\\ {\scriptsize $n=650$}\end{tabular} \\
DeepSeek-R1-7B & Ignores image & \begin{tabular}[t]{@{}l@{}}47.2 $\pm$ 1.3\\ {\scriptsize [44.7, 50.1]}\\ {\scriptsize $n=1{,}280$}\end{tabular} & \begin{tabular}[t]{@{}l@{}}46.9 $\pm$ 1.3\\ {\scriptsize [44.4, 49.8]}\\ {\scriptsize $n=1{,}280$}\end{tabular} & \begin{tabular}[t]{@{}l@{}}100.0 $\pm$ 0.0\\ {\scriptsize [100.0, 100.0]}\\ {\scriptsize $n=604$}\end{tabular} & \begin{tabular}[t]{@{}l@{}}0.0 $\pm$ 0.0\\ {\scriptsize [0.0, 0.0]}\\ {\scriptsize $n=604$}\end{tabular} & \begin{tabular}[t]{@{}l@{}}$+0.0$ $\pm$ 0.0\\ {\scriptsize [+0.0, +0.0]}\\ {\scriptsize $n=604$}\end{tabular} \\
\midrule
\multicolumn{7}{@{}l}{\textbf{Vision-only reference}} \\
RAD-DINO & Uses image & \begin{tabular}[t]{@{}l@{}}68.0 $\pm$ 1.3\\ {\scriptsize [65.5, 70.4]}\\ {\scriptsize $n=1{,}380$}\end{tabular} & \begin{tabular}[t]{@{}l@{}}68.7 $\pm$ 1.2\\ {\scriptsize [66.4, 71.0]}\\ {\scriptsize $n=1{,}380$}\end{tabular} & \begin{tabular}[t]{@{}l@{}}82.9 $\pm$ 1.2\\ {\scriptsize [80.6, 85.1]}\\ {\scriptsize $n=938$}\end{tabular} & \begin{tabular}[t]{@{}l@{}}68.6 $\pm$ 1.5\\ {\scriptsize [65.6, 71.3]}\\ {\scriptsize $n=938$}\end{tabular} & \begin{tabular}[t]{@{}l@{}}$+51.5$ $\pm$ 1.7\\ {\scriptsize [+48.1, +54.8]}\\ {\scriptsize $n=938$}\end{tabular} \\
\midrule
\multicolumn{7}{@{}l}{\textbf{Constant references}} \\
Always Yes & -- & \begin{tabular}[t]{@{}l@{}}47.1 $\pm$ 1.4\\ {\scriptsize [44.4, 49.7]}\\ {\scriptsize $n=1{,}380$}\end{tabular} & \begin{tabular}[t]{@{}l@{}}50.0 $\pm$ 0.0\\ {\scriptsize [50.0, 50.0]}\\ {\scriptsize $n=1{,}380$}\end{tabular} & N/A & N/A & N/A \\
Always No & -- & \begin{tabular}[t]{@{}l@{}}52.9 $\pm$ 1.4\\ {\scriptsize [50.3, 55.6]}\\ {\scriptsize $n=1{,}380$}\end{tabular} & \begin{tabular}[t]{@{}l@{}}50.0 $\pm$ 0.0\\ {\scriptsize [50.0, 50.0]}\\ {\scriptsize $n=1{,}380$}\end{tabular} & N/A & N/A & N/A \\
\bottomrule
\end{tabular}
\end{table}

\begin{table}[t]
\centering
\caption{Confidence by decision regime and calibration on the MIMIC question set ($n$ = 2{,}548 cases), for every system except DeepSeek-R1-7B, which has no confidence. Confidence is the probability of the given answer. Grounded correct comprises the correct MS-CXR answers that change under the target mask, ungrounded correct comprises those that stay unchanged, and incorrect comprises the incorrect answers on the pooled question set, each given as the mean $\pm$ SD and the count. AUROC is that of the affirmative probability for the Yes label, $\pm$ the SD of 1{,}000 patient-cluster bootstrap resamples, with the percentile 95\% interval. The Brier score is computed on the affirmative probability, and the expected calibration error (ECE, 10 equal-width bins) on the confidence. AUROC, area under the receiver operating characteristic curve; SD, standard deviation. N/A, not applicable.}
\label{stab:confidence}
\setlength{\tabcolsep}{2pt}
\renewcommand{\arraystretch}{1.05}
\scriptsize
\begin{tabular}{@{}p{0.15\textwidth}p{0.15\textwidth}p{0.15\textwidth}p{0.15\textwidth}p{0.15\textwidth}p{0.06\textwidth}p{0.06\textwidth}@{}}
\toprule
Model & Grounded correct & Ungrounded correct & Incorrect & AUROC & Brier & ECE \\
\midrule
Gemma-4-26B & \begin{tabular}[t]{@{}l@{}}97.9 $\pm$ 5.4\\ {\scriptsize $n=105$}\end{tabular} & \begin{tabular}[t]{@{}l@{}}99.7 $\pm$ 2.1\\ {\scriptsize $n=247$}\end{tabular} & \begin{tabular}[t]{@{}l@{}}94.0 $\pm$ 11.5\\ {\scriptsize $n=777$}\end{tabular} & \begin{tabular}[t]{@{}l@{}}74.8 $\pm$ 1.1\\ {\scriptsize [72.6, 76.7]}\end{tabular} & 0.285 & 0.276 \\
Qwen3-VL-32B & \begin{tabular}[t]{@{}l@{}}82.3 $\pm$ 15.4\\ {\scriptsize $n=57$}\end{tabular} & \begin{tabular}[t]{@{}l@{}}93.5 $\pm$ 12.1\\ {\scriptsize $n=268$}\end{tabular} & \begin{tabular}[t]{@{}l@{}}87.1 $\pm$ 15.1\\ {\scriptsize $n=892$}\end{tabular} & \begin{tabular}[t]{@{}l@{}}70.8 $\pm$ 1.1\\ {\scriptsize [68.6, 72.8]}\end{tabular} & 0.288 & 0.254 \\
Mistral-Small-4-119B & \begin{tabular}[t]{@{}l@{}}64.5 $\pm$ 10.7\\ {\scriptsize $n=10$}\end{tabular} & \begin{tabular}[t]{@{}l@{}}70.3 $\pm$ 8.1\\ {\scriptsize $n=15$}\end{tabular} & \begin{tabular}[t]{@{}l@{}}83.0 $\pm$ 13.5\\ {\scriptsize $n=1{,}361$}\end{tabular} & \begin{tabular}[t]{@{}l@{}}46.2 $\pm$ 1.3\\ {\scriptsize [43.6, 48.8]}\end{tabular} & 0.399 & 0.366 \\
MedGemma-1.5-4B & \begin{tabular}[t]{@{}l@{}}95.0 $\pm$ 11.3\\ {\scriptsize $n=125$}\end{tabular} & \begin{tabular}[t]{@{}l@{}}99.7 $\pm$ 1.3\\ {\scriptsize $n=248$}\end{tabular} & \begin{tabular}[t]{@{}l@{}}94.0 $\pm$ 11.5\\ {\scriptsize $n=1{,}065$}\end{tabular} & \begin{tabular}[t]{@{}l@{}}67.6 $\pm$ 1.2\\ {\scriptsize [65.3, 69.8]}\end{tabular} & 0.380 & 0.372 \\
LLaVA-Med-7B & N/A & \begin{tabular}[t]{@{}l@{}}99.0 $\pm$ 0.7\\ {\scriptsize $n=447$}\end{tabular} & \begin{tabular}[t]{@{}l@{}}98.0 $\pm$ 2.0\\ {\scriptsize $n=1{,}309$}\end{tabular} & \begin{tabular}[t]{@{}l@{}}62.6 $\pm$ 1.2\\ {\scriptsize [60.0, 65.0]}\end{tabular} & 0.501 & 0.505 \\
MedGemma-27B-text & N/A & \begin{tabular}[t]{@{}l@{}}92.6 $\pm$ 6.7\\ {\scriptsize $n=415$}\end{tabular} & \begin{tabular}[t]{@{}l@{}}90.2 $\pm$ 11.5\\ {\scriptsize $n=1{,}027$}\end{tabular} & \begin{tabular}[t]{@{}l@{}}58.8 $\pm$ 1.3\\ {\scriptsize [56.2, 61.3]}\end{tabular} & 0.376 & 0.366 \\
RAD-DINO & \begin{tabular}[t]{@{}l@{}}68.7 $\pm$ 14.8\\ {\scriptsize $n=27$}\end{tabular} & \begin{tabular}[t]{@{}l@{}}91.2 $\pm$ 11.3\\ {\scriptsize $n=403$}\end{tabular} & \begin{tabular}[t]{@{}l@{}}80.7 $\pm$ 15.6\\ {\scriptsize $n=515$}\end{tabular} & \begin{tabular}[t]{@{}l@{}}74.6 $\pm$ 1.2\\ {\scriptsize [72.4, 77.0]}\end{tabular} & 0.215 & 0.139 \\
\bottomrule
\end{tabular}
\end{table}

\begin{table}[t]
\centering
\caption{Agreement between the readers on the balanced set ($n$ = 200 cases) and the difficulty-stratified set ($n$ = 80), between each reader and the report-derived label, and between the raters of 120 boxes and of the 22 taxonomy cases. Percent agreement is the share of displays, boxes, or cases with the same answer, rating, or category, as the mean over its cases $\pm$ the SD of 1{,}000 patient-cluster bootstrap resamples, with the percentile 95\% interval. Cohen's $\kappa$, or Fleiss' $\kappa$ for three readers, is the chance-corrected agreement, and the weighted $\kappa$ is the quadratic-weighted agreement on the 1-to-5 confidence rating. Each $\kappa$ is given with its one-sided permutation $p$-value against no agreement beyond chance (1{,}000 permutations, seed 0) and its count. SD, standard deviation. N/A, not applicable.}
\label{stab:reader_agreement}
\setlength{\tabcolsep}{3pt}
\renewcommand{\arraystretch}{0.95}
\scriptsize
\begin{tabular}{@{}>{\raggedright\arraybackslash}p{0.27\textwidth}p{0.30\textwidth}p{0.18\textwidth}p{0.18\textwidth}@{}}
\toprule
Readers or raters & Percent agreement & $\kappa$ & Weighted $\kappa$ \\
\midrule
\multicolumn{4}{@{}l}{\textbf{Balanced set, L.A. and T.T.N.}} \\
Original displays & \begin{tabular}[t]{@{}l@{}}81.0 $\pm$ 2.8 [75.4, 86.1]\\ {\scriptsize $n=200$}\end{tabular} & \begin{tabular}[t]{@{}l@{}}0.620\\ {\scriptsize $p < 0.001$, $n=200$}\end{tabular} & \begin{tabular}[t]{@{}l@{}}0.218\\ {\scriptsize $p < 0.001$, $n=200$}\end{tabular} \\
Target-mask displays & \begin{tabular}[t]{@{}l@{}}69.0 $\pm$ 4.6 [60.0, 78.0]\\ {\scriptsize $n=100$}\end{tabular} & \begin{tabular}[t]{@{}l@{}}0.327\\ {\scriptsize $p < 0.001$, $n=100$}\end{tabular} & \begin{tabular}[t]{@{}l@{}}0.592\\ {\scriptsize $p < 0.001$, $n=100$}\end{tabular} \\
Corner-mask displays & \begin{tabular}[t]{@{}l@{}}85.0 $\pm$ 3.6 [77.1, 91.7]\\ {\scriptsize $n=100$}\end{tabular} & \begin{tabular}[t]{@{}l@{}}0.265\\ {\scriptsize $p$ = 0.016, $n=100$}\end{tabular} & \begin{tabular}[t]{@{}l@{}}0.080\\ {\scriptsize $p$ = 0.133, $n=100$}\end{tabular} \\
All masked displays & \begin{tabular}[t]{@{}l@{}}77.0 $\pm$ 3.3 [70.2, 82.8]\\ {\scriptsize $n=200$}\end{tabular} & \begin{tabular}[t]{@{}l@{}}0.344\\ {\scriptsize $p < 0.001$, $n=200$}\end{tabular} & \begin{tabular}[t]{@{}l@{}}0.529\\ {\scriptsize $p < 0.001$, $n=200$}\end{tabular} \\
\midrule
\multicolumn{4}{@{}l}{\textbf{Balanced set, reader and report-derived label}} \\
T.T.N. & \begin{tabular}[t]{@{}l@{}}86.0 $\pm$ 2.5 [81.0, 91.0]\\ {\scriptsize $n=200$}\end{tabular} & \begin{tabular}[t]{@{}l@{}}0.720\\ {\scriptsize $p < 0.001$, $n=200$}\end{tabular} & N/A \\
L.A. & \begin{tabular}[t]{@{}l@{}}82.0 $\pm$ 2.7 [76.5, 87.2]\\ {\scriptsize $n=200$}\end{tabular} & \begin{tabular}[t]{@{}l@{}}0.640\\ {\scriptsize $p < 0.001$, $n=200$}\end{tabular} & N/A \\
T.T.N., without uncertain-label negatives & \begin{tabular}[t]{@{}l@{}}86.1 $\pm$ 2.5 [81.2, 91.1]\\ {\scriptsize $n=187$}\end{tabular} & \begin{tabular}[t]{@{}l@{}}0.721\\ {\scriptsize $p < 0.001$, $n=187$}\end{tabular} & N/A \\
L.A., without uncertain-label negatives & \begin{tabular}[t]{@{}l@{}}81.3 $\pm$ 2.9 [75.4, 86.6]\\ {\scriptsize $n=187$}\end{tabular} & \begin{tabular}[t]{@{}l@{}}0.620\\ {\scriptsize $p < 0.001$, $n=187$}\end{tabular} & N/A \\
\midrule
\multicolumn{4}{@{}l}{\textbf{Difficulty-stratified set, original displays}} \\
S.Z. and T.T.N. & \begin{tabular}[t]{@{}l@{}}80.0 $\pm$ 4.9 [70.0, 89.2]\\ {\scriptsize $n=80$}\end{tabular} & \begin{tabular}[t]{@{}l@{}}0.431\\ {\scriptsize $p < 0.001$, $n=80$}\end{tabular} & \begin{tabular}[t]{@{}l@{}}0.306\\ {\scriptsize $p < 0.001$, $n=80$}\end{tabular} \\
L.A. and S.Z. & \begin{tabular}[t]{@{}l@{}}65.0 $\pm$ 6.1 [52.6, 76.9]\\ {\scriptsize $n=80$}\end{tabular} & \begin{tabular}[t]{@{}l@{}}0.214\\ {\scriptsize $p$ = 0.034, $n=80$}\end{tabular} & \begin{tabular}[t]{@{}l@{}}0.137\\ {\scriptsize $p$ = 0.066, $n=80$}\end{tabular} \\
L.A. and T.T.N. & \begin{tabular}[t]{@{}l@{}}65.0 $\pm$ 6.0 [53.5, 76.3]\\ {\scriptsize $n=80$}\end{tabular} & \begin{tabular}[t]{@{}l@{}}0.237\\ {\scriptsize $p$ = 0.023, $n=80$}\end{tabular} & \begin{tabular}[t]{@{}l@{}}0.333\\ {\scriptsize $p$ = 0.004, $n=80$}\end{tabular} \\
Three readers (Fleiss) & \begin{tabular}[t]{@{}l@{}}55.0 $\pm$ 6.2 [42.7, 67.5]\\ {\scriptsize $n=80$}\end{tabular} & \begin{tabular}[t]{@{}l@{}}0.268\\ {\scriptsize $p < 0.001$, $n=80$}\end{tabular} & N/A \\
Three readers, masked & \begin{tabular}[t]{@{}l@{}}51.3 $\pm$ 5.0 [41.5, 61.0]\\ {\scriptsize $n=160$}\end{tabular} & \begin{tabular}[t]{@{}l@{}}0.291\\ {\scriptsize $p < 0.001$, $n=160$}\end{tabular} & N/A \\
\midrule
\multicolumn{4}{@{}l}{\textbf{Box ratings, S.Z. and T.T.N., 120 boxes}} \\
Four rating levels & \begin{tabular}[t]{@{}l@{}}70.8 $\pm$ 4.5 [61.8, 79.7]\\ {\scriptsize $n=120$}\end{tabular} & \begin{tabular}[t]{@{}l@{}}0.533\\ {\scriptsize $p < 0.001$, $n=120$}\end{tabular} & N/A \\
Accurate against the rest & N/A & \begin{tabular}[t]{@{}l@{}}0.580\\ {\scriptsize $p < 0.001$, $n=120$}\end{tabular} & N/A \\
\midrule
\multicolumn{4}{@{}l}{\textbf{Failure taxonomy, L.A. and T.T.N., 22 cases}} \\
Five categories & \begin{tabular}[t]{@{}l@{}}31.8 $\pm$ 10.1\\ {\scriptsize [13.6, 50.1]}\\ {\scriptsize $n=22$}\end{tabular} & \begin{tabular}[t]{@{}l@{}}0.078\\ {\scriptsize $p$ = 0.342, $n=22$}\end{tabular} & \begin{tabular}[t]{@{}l@{}}0.299\\ {\scriptsize $p$ = 0.091, $n=22$}\end{tabular} \\
\bottomrule
\end{tabular}
\end{table}

\begin{table}[t]
\centering
\caption{Accuracy and balanced accuracy of all systems against the two readers on the original displays of the balanced set ($n$ = 200 cases, 100 of them finding present). The first row lists each reader's value as the mean over its cases $\pm$ the SD of 1{,}000 patient-cluster bootstrap resamples, with the percentile 95\% interval and the case count. Every other row is the paired difference, system minus reader, on the cases answered by both the system and the reader. It is given as the observed difference $\pm$ the SD of its paired bootstrap resamples, with the 95\% interval, the FDR-corrected $p$-value of the paired bootstrap test, and the shared case count, and with the 90\% interval of the two one-sided tests at a margin of $\pm$10. FDR, false discovery rate; SD, standard deviation.}
\label{stab:reader_balanced}
\setlength{\tabcolsep}{2pt}
\renewcommand{\arraystretch}{0.95}
\scriptsize
\begin{tabular}{@{}p{0.155\textwidth}p{0.19\textwidth}p{0.19\textwidth}p{0.19\textwidth}p{0.19\textwidth}@{}}
\toprule
System & Accuracy, T.T.N. & Accuracy, L.A. & Balanced accuracy, T.T.N. & Balanced accuracy, L.A. \\
\midrule
\textbf{Reader value} & \begin{tabular}[t]{@{}l@{}}86.0 $\pm$ 2.5\\ {\scriptsize [81.0, 91.0]}\\ {\scriptsize $n=200$}\end{tabular} & \begin{tabular}[t]{@{}l@{}}82.0 $\pm$ 2.7\\ {\scriptsize [76.5, 87.2]}\\ {\scriptsize $n=200$}\end{tabular} & \begin{tabular}[t]{@{}l@{}}86.0 $\pm$ 2.5\\ {\scriptsize [80.9, 90.9]}\\ {\scriptsize $n=200$}\end{tabular} & \begin{tabular}[t]{@{}l@{}}82.0 $\pm$ 2.6\\ {\scriptsize [76.9, 87.0]}\\ {\scriptsize $n=200$}\end{tabular} \\
\midrule
\multicolumn{5}{@{}l}{\textbf{System minus reader}} \\
Gemma-4-26B & \begin{tabular}[t]{@{}l@{}}$-16.1$ $\pm$ 3.9\\ {\scriptsize [-23.7, -9.3]}\\ {\scriptsize $p_{\mathrm{FDR}}$ = 0.001, $n=199$}\\ {\scriptsize 90\% [-22.5, -10.0]}\end{tabular} & \begin{tabular}[t]{@{}l@{}}$-12.6$ $\pm$ 3.3\\ {\scriptsize [-18.9, -6.1]}\\ {\scriptsize $p_{\mathrm{FDR}}$ = 0.001, $n=199$}\\ {\scriptsize 90\% [-18.0, -7.2]}\end{tabular} & \begin{tabular}[t]{@{}l@{}}$-16.0$ $\pm$ 3.7\\ {\scriptsize [-23.4, -9.3]}\\ {\scriptsize $p_{\mathrm{FDR}}$ = 0.001, $n=199$}\\ {\scriptsize 90\% [-22.5, -10.4]}\end{tabular} & \begin{tabular}[t]{@{}l@{}}$-12.5$ $\pm$ 3.3\\ {\scriptsize [-18.9, -6.2]}\\ {\scriptsize $p_{\mathrm{FDR}}$ = 0.001, $n=199$}\\ {\scriptsize 90\% [-18.0, -7.3]}\end{tabular} \\
Qwen3-VL-32B & \begin{tabular}[t]{@{}l@{}}$-21.0$ $\pm$ 3.9\\ {\scriptsize [-28.4, -13.6]}\\ {\scriptsize $p_{\mathrm{FDR}}$ = 0.001, $n=200$}\\ {\scriptsize 90\% [-27.3, -14.9]}\end{tabular} & \begin{tabular}[t]{@{}l@{}}$-17.0$ $\pm$ 3.6\\ {\scriptsize [-24.5, -10.0]}\\ {\scriptsize $p_{\mathrm{FDR}}$ = 0.001, $n=200$}\\ {\scriptsize 90\% [-23.0, -11.3]}\end{tabular} & \begin{tabular}[t]{@{}l@{}}$-21.0$ $\pm$ 3.9\\ {\scriptsize [-28.5, -13.8]}\\ {\scriptsize $p_{\mathrm{FDR}}$ = 0.001, $n=200$}\\ {\scriptsize 90\% [-27.3, -14.9]}\end{tabular} & \begin{tabular}[t]{@{}l@{}}$-17.0$ $\pm$ 3.6\\ {\scriptsize [-24.4, -10.0]}\\ {\scriptsize $p_{\mathrm{FDR}}$ = 0.001, $n=200$}\\ {\scriptsize 90\% [-23.0, -11.3]}\end{tabular} \\
Mistral-Small-4-119B & \begin{tabular}[t]{@{}l@{}}$-34.5$ $\pm$ 4.8\\ {\scriptsize [-44.6, -25.3]}\\ {\scriptsize $p_{\mathrm{FDR}}$ = 0.001, $n=200$}\\ {\scriptsize 90\% [-42.2, -26.6]}\end{tabular} & \begin{tabular}[t]{@{}l@{}}$-30.5$ $\pm$ 5.0\\ {\scriptsize [-40.4, -20.7]}\\ {\scriptsize $p_{\mathrm{FDR}}$ = 0.001, $n=200$}\\ {\scriptsize 90\% [-38.7, -22.6]}\end{tabular} & \begin{tabular}[t]{@{}l@{}}$-34.5$ $\pm$ 3.5\\ {\scriptsize [-41.2, -27.3]}\\ {\scriptsize $p_{\mathrm{FDR}}$ = 0.001, $n=200$}\\ {\scriptsize 90\% [-40.5, -28.5]}\end{tabular} & \begin{tabular}[t]{@{}l@{}}$-30.5$ $\pm$ 3.2\\ {\scriptsize [-36.8, -24.0]}\\ {\scriptsize $p_{\mathrm{FDR}}$ = 0.001, $n=200$}\\ {\scriptsize 90\% [-35.8, -25.2]}\end{tabular} \\
MedGemma-1.5-4B & \begin{tabular}[t]{@{}l@{}}$-14.5$ $\pm$ 3.8\\ {\scriptsize [-21.6, -6.9]}\\ {\scriptsize $p_{\mathrm{FDR}}$ = 0.001, $n=200$}\\ {\scriptsize 90\% [-20.8, -8.4]}\end{tabular} & \begin{tabular}[t]{@{}l@{}}$-10.5$ $\pm$ 2.7\\ {\scriptsize [-16.0, -5.4]}\\ {\scriptsize $p_{\mathrm{FDR}}$ = 0.001, $n=200$}\\ {\scriptsize 90\% [-15.3, -6.0]}\end{tabular} & \begin{tabular}[t]{@{}l@{}}$-14.5$ $\pm$ 3.8\\ {\scriptsize [-21.4, -6.8]}\\ {\scriptsize $p_{\mathrm{FDR}}$ = 0.001, $n=200$}\\ {\scriptsize 90\% [-20.7, -8.2]}\end{tabular} & \begin{tabular}[t]{@{}l@{}}$-10.5$ $\pm$ 2.7\\ {\scriptsize [-16.0, -5.4]}\\ {\scriptsize $p_{\mathrm{FDR}}$ = 0.001, $n=200$}\\ {\scriptsize 90\% [-15.3, -6.1]}\end{tabular} \\
LLaVA-Med-7B & \begin{tabular}[t]{@{}l@{}}$-35.0$ $\pm$ 4.5\\ {\scriptsize [-43.8, -26.9]}\\ {\scriptsize $p_{\mathrm{FDR}}$ = 0.001, $n=197$}\\ {\scriptsize 90\% [-42.5, -27.8]}\end{tabular} & \begin{tabular}[t]{@{}l@{}}$-31.0$ $\pm$ 4.3\\ {\scriptsize [-39.6, -23.0]}\\ {\scriptsize $p_{\mathrm{FDR}}$ = 0.001, $n=197$}\\ {\scriptsize 90\% [-38.2, -23.9]}\end{tabular} & \begin{tabular}[t]{@{}l@{}}$-35.8$ $\pm$ 2.6\\ {\scriptsize [-40.6, -30.5]}\\ {\scriptsize $p_{\mathrm{FDR}}$ = 0.001, $n=197$}\\ {\scriptsize 90\% [-39.8, -31.4]}\end{tabular} & \begin{tabular}[t]{@{}l@{}}$-31.6$ $\pm$ 2.9\\ {\scriptsize [-36.7, -25.7]}\\ {\scriptsize $p_{\mathrm{FDR}}$ = 0.001, $n=197$}\\ {\scriptsize 90\% [-36.1, -26.4]}\end{tabular} \\
MedGemma-27B-text & \begin{tabular}[t]{@{}l@{}}$-36.0$ $\pm$ 4.4\\ {\scriptsize [-44.5, -27.6]}\\ {\scriptsize $p_{\mathrm{FDR}}$ = 0.001, $n=200$}\\ {\scriptsize 90\% [-43.3, -28.7]}\end{tabular} & \begin{tabular}[t]{@{}l@{}}$-32.0$ $\pm$ 3.9\\ {\scriptsize [-39.5, -24.0]}\\ {\scriptsize $p_{\mathrm{FDR}}$ = 0.001, $n=200$}\\ {\scriptsize 90\% [-38.2, -25.9]}\end{tabular} & \begin{tabular}[t]{@{}l@{}}$-36.0$ $\pm$ 3.4\\ {\scriptsize [-42.5, -29.2]}\\ {\scriptsize $p_{\mathrm{FDR}}$ = 0.001, $n=200$}\\ {\scriptsize 90\% [-41.4, -30.5]}\end{tabular} & \begin{tabular}[t]{@{}l@{}}$-32.0$ $\pm$ 3.2\\ {\scriptsize [-38.0, -25.7]}\\ {\scriptsize $p_{\mathrm{FDR}}$ = 0.001, $n=200$}\\ {\scriptsize 90\% [-37.1, -26.9]}\end{tabular} \\
DeepSeek-R1-7B & \begin{tabular}[t]{@{}l@{}}$-35.6$ $\pm$ 4.7\\ {\scriptsize [-44.6, -26.6]}\\ {\scriptsize $p_{\mathrm{FDR}}$ = 0.001, $n=174$}\\ {\scriptsize 90\% [-43.4, -28.0]}\end{tabular} & \begin{tabular}[t]{@{}l@{}}$-29.9$ $\pm$ 5.8\\ {\scriptsize [-40.5, -18.5]}\\ {\scriptsize $p_{\mathrm{FDR}}$ = 0.001, $n=174$}\\ {\scriptsize 90\% [-39.2, -20.6]}\end{tabular} & \begin{tabular}[t]{@{}l@{}}$-35.6$ $\pm$ 4.3\\ {\scriptsize [-44.0, -27.2]}\\ {\scriptsize $p_{\mathrm{FDR}}$ = 0.001, $n=174$}\\ {\scriptsize 90\% [-42.3, -28.5]}\end{tabular} & \begin{tabular}[t]{@{}l@{}}$-29.9$ $\pm$ 5.1\\ {\scriptsize [-39.5, -19.6]}\\ {\scriptsize $p_{\mathrm{FDR}}$ = 0.001, $n=174$}\\ {\scriptsize 90\% [-38.1, -21.1]}\end{tabular} \\
RAD-DINO & \begin{tabular}[t]{@{}l@{}}$-13.0$ $\pm$ 3.7\\ {\scriptsize [-20.2, -6.1]}\\ {\scriptsize $p_{\mathrm{FDR}}$ = 0.001, $n=200$}\\ {\scriptsize 90\% [-19.2, -7.0]}\end{tabular} & \begin{tabular}[t]{@{}l@{}}$-9.0$ $\pm$ 3.1\\ {\scriptsize [-15.2, -3.4]}\\ {\scriptsize $p_{\mathrm{FDR}}$ = 0.005, $n=200$}\\ {\scriptsize 90\% [-14.2, -4.1]}\end{tabular} & \begin{tabular}[t]{@{}l@{}}$-13.0$ $\pm$ 3.4\\ {\scriptsize [-19.4, -6.6]}\\ {\scriptsize $p_{\mathrm{FDR}}$ = 0.001, $n=200$}\\ {\scriptsize 90\% [-18.7, -7.3]}\end{tabular} & \begin{tabular}[t]{@{}l@{}}$-9.0$ $\pm$ 3.0\\ {\scriptsize [-15.1, -3.3]}\\ {\scriptsize $p_{\mathrm{FDR}}$ = 0.005, $n=200$}\\ {\scriptsize 90\% [-14.0, -4.4]}\end{tabular} \\
\bottomrule
\end{tabular}
\end{table}

\begin{table}[t]
\centering
\caption{Sensitivity and specificity of all systems against the two readers on the balanced set ($n$ = 100 finding-present and 100 finding-absent cases). The first row lists each reader's value as the mean over its cases $\pm$ the SD of 1{,}000 patient-cluster bootstrap resamples, with the percentile 95\% interval and the case count. Every other row is the paired difference, system minus reader, on the cases answered by both the system and the reader. It is given as the observed difference $\pm$ the SD of its paired bootstrap resamples, with the 95\% interval, the FDR-corrected $p$-value of the paired bootstrap test, and the shared case count. FDR, false discovery rate; SD, standard deviation.}
\label{stab:reader_senspec}
\setlength{\tabcolsep}{2pt}
\renewcommand{\arraystretch}{0.95}
\scriptsize
\begin{tabular}{@{}p{0.155\textwidth}p{0.19\textwidth}p{0.19\textwidth}p{0.19\textwidth}p{0.19\textwidth}@{}}
\toprule
System & Sensitivity, T.T.N. & Sensitivity, L.A. & Specificity, T.T.N. & Specificity, L.A. \\
\midrule
\textbf{Reader value} & \begin{tabular}[t]{@{}l@{}}86.0 $\pm$ 3.4\\ {\scriptsize [79.0, 92.2]}\\ {\scriptsize $n=100$}\end{tabular} & \begin{tabular}[t]{@{}l@{}}90.0 $\pm$ 3.0\\ {\scriptsize [83.7, 95.2]}\\ {\scriptsize $n=100$}\end{tabular} & \begin{tabular}[t]{@{}l@{}}86.0 $\pm$ 3.6\\ {\scriptsize [78.6, 92.9]}\\ {\scriptsize $n=100$}\end{tabular} & \begin{tabular}[t]{@{}l@{}}74.0 $\pm$ 4.6\\ {\scriptsize [64.4, 82.0]}\\ {\scriptsize $n=100$}\end{tabular} \\
\midrule
\multicolumn{5}{@{}l}{\textbf{System minus reader}} \\
Gemma-4-26B & \begin{tabular}[t]{@{}l@{}}$+0.0$ $\pm$ 5.1\\ {\scriptsize [-9.9, +9.8]}\\ {\scriptsize $p_{\mathrm{FDR}} > 0.999$, $n=99$}\end{tabular} & \begin{tabular}[t]{@{}l@{}}$-5.1$ $\pm$ 4.2\\ {\scriptsize [-13.3, +3.1]}\\ {\scriptsize $p_{\mathrm{FDR}}$ = 0.303, $n=99$}\end{tabular} & \begin{tabular}[t]{@{}l@{}}$-32.0$ $\pm$ 5.1\\ {\scriptsize [-42.0, -22.2]}\\ {\scriptsize $p_{\mathrm{FDR}}$ = 0.001, $n=100$}\end{tabular} & \begin{tabular}[t]{@{}l@{}}$-20.0$ $\pm$ 5.3\\ {\scriptsize [-30.0, -10.0]}\\ {\scriptsize $p_{\mathrm{FDR}}$ = 0.001, $n=100$}\end{tabular} \\
Qwen3-VL-32B & \begin{tabular}[t]{@{}l@{}}$-13.0$ $\pm$ 5.7\\ {\scriptsize [-24.8, -2.0]}\\ {\scriptsize $p_{\mathrm{FDR}}$ = 0.056, $n=100$}\end{tabular} & \begin{tabular}[t]{@{}l@{}}$-17.0$ $\pm$ 5.6\\ {\scriptsize [-27.9, -6.2]}\\ {\scriptsize $p_{\mathrm{FDR}}$ = 0.003, $n=100$}\end{tabular} & \begin{tabular}[t]{@{}l@{}}$-29.0$ $\pm$ 5.1\\ {\scriptsize [-39.4, -19.8]}\\ {\scriptsize $p_{\mathrm{FDR}}$ = 0.001, $n=100$}\end{tabular} & \begin{tabular}[t]{@{}l@{}}$-17.0$ $\pm$ 4.8\\ {\scriptsize [-27.3, -8.1]}\\ {\scriptsize $p_{\mathrm{FDR}}$ = 0.001, $n=100$}\end{tabular} \\
Mistral-Small-4-119B & \begin{tabular}[t]{@{}l@{}}$-75.0$ $\pm$ 5.4\\ {\scriptsize [-85.3, -65.0]}\\ {\scriptsize $p_{\mathrm{FDR}}$ = 0.003, $n=100$}\end{tabular} & \begin{tabular}[t]{@{}l@{}}$-79.0$ $\pm$ 4.3\\ {\scriptsize [-87.1, -70.0]}\\ {\scriptsize $p_{\mathrm{FDR}}$ = 0.003, $n=100$}\end{tabular} & \begin{tabular}[t]{@{}l@{}}$+6.0$ $\pm$ 4.6\\ {\scriptsize [-3.0, +14.9]}\\ {\scriptsize $p_{\mathrm{FDR}}$ = 0.197, $n=100$}\end{tabular} & \begin{tabular}[t]{@{}l@{}}$+18.0$ $\pm$ 5.0\\ {\scriptsize [+8.1, +28.4]}\\ {\scriptsize $p_{\mathrm{FDR}}$ = 0.001, $n=100$}\end{tabular} \\
MedGemma-1.5-4B & \begin{tabular}[t]{@{}l@{}}$-3.0$ $\pm$ 5.4\\ {\scriptsize [-14.1, +7.2]}\\ {\scriptsize $p_{\mathrm{FDR}}$ = 0.815, $n=100$}\end{tabular} & \begin{tabular}[t]{@{}l@{}}$-7.0$ $\pm$ 4.4\\ {\scriptsize [-16.2, +1.0]}\\ {\scriptsize $p_{\mathrm{FDR}}$ = 0.193, $n=100$}\end{tabular} & \begin{tabular}[t]{@{}l@{}}$-26.0$ $\pm$ 5.0\\ {\scriptsize [-36.4, -17.0]}\\ {\scriptsize $p_{\mathrm{FDR}}$ = 0.001, $n=100$}\end{tabular} & \begin{tabular}[t]{@{}l@{}}$-14.0$ $\pm$ 3.8\\ {\scriptsize [-21.2, -7.0]}\\ {\scriptsize $p_{\mathrm{FDR}}$ = 0.003, $n=100$}\end{tabular} \\
LLaVA-Med-7B & \begin{tabular}[t]{@{}l@{}}$+14.0$ $\pm$ 3.4\\ {\scriptsize [+7.8, +21.0]}\\ {\scriptsize $p_{\mathrm{FDR}}$ = 0.003, $n=100$}\end{tabular} & \begin{tabular}[t]{@{}l@{}}$+10.0$ $\pm$ 3.0\\ {\scriptsize [+4.8, +16.3]}\\ {\scriptsize $p_{\mathrm{FDR}}$ = 0.006, $n=100$}\end{tabular} & \begin{tabular}[t]{@{}l@{}}$-85.6$ $\pm$ 3.8\\ {\scriptsize [-92.7, -77.3]}\\ {\scriptsize $p_{\mathrm{FDR}}$ = 0.001, $n=97$}\end{tabular} & \begin{tabular}[t]{@{}l@{}}$-73.2$ $\pm$ 4.7\\ {\scriptsize [-82.3, -63.5]}\\ {\scriptsize $p_{\mathrm{FDR}}$ = 0.001, $n=97$}\end{tabular} \\
MedGemma-27B-text & \begin{tabular}[t]{@{}l@{}}$+2.0$ $\pm$ 4.9\\ {\scriptsize [-7.7, +11.2]}\\ {\scriptsize $p_{\mathrm{FDR}}$ = 0.839, $n=100$}\end{tabular} & \begin{tabular}[t]{@{}l@{}}$-2.0$ $\pm$ 4.2\\ {\scriptsize [-10.1, +6.1]}\\ {\scriptsize $p_{\mathrm{FDR}}$ = 0.679, $n=100$}\end{tabular} & \begin{tabular}[t]{@{}l@{}}$-74.0$ $\pm$ 4.4\\ {\scriptsize [-82.8, -65.0]}\\ {\scriptsize $p_{\mathrm{FDR}}$ = 0.001, $n=100$}\end{tabular} & \begin{tabular}[t]{@{}l@{}}$-62.0$ $\pm$ 5.0\\ {\scriptsize [-72.0, -52.0]}\\ {\scriptsize $p_{\mathrm{FDR}}$ = 0.001, $n=100$}\end{tabular} \\
DeepSeek-R1-7B & \begin{tabular}[t]{@{}l@{}}$-57.5$ $\pm$ 6.2\\ {\scriptsize [-69.2, -44.6]}\\ {\scriptsize $p_{\mathrm{FDR}}$ = 0.003, $n=87$}\end{tabular} & \begin{tabular}[t]{@{}l@{}}$-60.9$ $\pm$ 6.6\\ {\scriptsize [-72.7, -46.6]}\\ {\scriptsize $p_{\mathrm{FDR}}$ = 0.003, $n=87$}\end{tabular} & \begin{tabular}[t]{@{}l@{}}$-13.8$ $\pm$ 6.4\\ {\scriptsize [-26.4, -1.1]}\\ {\scriptsize $p_{\mathrm{FDR}}$ = 0.054, $n=87$}\end{tabular} & \begin{tabular}[t]{@{}l@{}}$+1.1$ $\pm$ 8.0\\ {\scriptsize [-13.8, +16.1]}\\ {\scriptsize $p_{\mathrm{FDR}}$ = 0.939, $n=87$}\end{tabular} \\
RAD-DINO & \begin{tabular}[t]{@{}l@{}}$+9.0$ $\pm$ 4.0\\ {\scriptsize [+1.0, +16.5]}\\ {\scriptsize $p_{\mathrm{FDR}}$ = 0.048, $n=100$}\end{tabular} & \begin{tabular}[t]{@{}l@{}}$+5.0$ $\pm$ 3.3\\ {\scriptsize [-1.0, +11.8]}\\ {\scriptsize $p_{\mathrm{FDR}}$ = 0.193, $n=100$}\end{tabular} & \begin{tabular}[t]{@{}l@{}}$-35.0$ $\pm$ 5.1\\ {\scriptsize [-45.5, -26.0]}\\ {\scriptsize $p_{\mathrm{FDR}}$ = 0.001, $n=100$}\end{tabular} & \begin{tabular}[t]{@{}l@{}}$-23.0$ $\pm$ 5.0\\ {\scriptsize [-33.3, -13.7]}\\ {\scriptsize $p_{\mathrm{FDR}}$ = 0.001, $n=100$}\end{tabular} \\
\bottomrule
\end{tabular}
\end{table}

\begin{table}[t]
\centering
\caption{Localization of all systems against the two readers on the positives of the balanced set ($n$ = 100). L.A. did not read the displays with the matched mask. The first row lists each reader's value as the mean over its cases $\pm$ the SD of 1{,}000 patient-cluster bootstrap resamples, with the percentile 95\% interval and the case count. Every other row is the paired difference, system minus reader, on the positives answered correctly by both the system and the reader. It is given as the observed difference $\pm$ the SD of its paired bootstrap resamples, with the 95\% interval, the FDR-corrected $p$-value of the paired bootstrap test, and the shared case count. CGR, causal grounding rate; FDR, false discovery rate; IS, irrelevant-mask stability; SD, standard deviation. N/A, not applicable.}
\label{stab:reader_masks}
\setlength{\tabcolsep}{2pt}
\renewcommand{\arraystretch}{0.95}
\scriptsize
\begin{tabular}{@{}p{0.175\textwidth}p{0.155\textwidth}p{0.155\textwidth}p{0.155\textwidth}p{0.155\textwidth}p{0.155\textwidth}@{}}
\toprule
System & CGR, T.T.N. & CGR, L.A. & IS (corner), T.T.N. & IS (corner), L.A. & IS (matched), T.T.N. \\
\midrule
\textbf{Reader value} & \begin{tabular}[t]{@{}l@{}}38.4 $\pm$ 5.2\\ {\scriptsize [28.4, 48.8]}\\ {\scriptsize $n=86$}\end{tabular} & \begin{tabular}[t]{@{}l@{}}13.3 $\pm$ 3.6\\ {\scriptsize [6.7, 21.2]}\\ {\scriptsize $n=90$}\end{tabular} & \begin{tabular}[t]{@{}l@{}}95.3 $\pm$ 2.3\\ {\scriptsize [90.4, 98.9]}\\ {\scriptsize $n=86$}\end{tabular} & \begin{tabular}[t]{@{}l@{}}100.0 $\pm$ 0.0\\ {\scriptsize [100.0, 100.0]}\\ {\scriptsize $n=90$}\end{tabular} & \begin{tabular}[t]{@{}l@{}}90.7 $\pm$ 3.2\\ {\scriptsize [84.1, 96.5]}\\ {\scriptsize $n=86$}\end{tabular} \\
\midrule
\multicolumn{6}{@{}l}{\textbf{System minus reader}} \\
Gemma-4-26B & \begin{tabular}[t]{@{}l@{}}$-11.3$ $\pm$ 6.0\\ {\scriptsize [-23.3, +0.0]}\\ {\scriptsize $p_{\mathrm{FDR}}$ = 0.104}\\ {\scriptsize $n=71$}\end{tabular} & \begin{tabular}[t]{@{}l@{}}$+15.4$ $\pm$ 5.1\\ {\scriptsize [+5.2, +25.3]}\\ {\scriptsize $p_{\mathrm{FDR}}$ = 0.011}\\ {\scriptsize $n=78$}\end{tabular} & \begin{tabular}[t]{@{}l@{}}$-1.4$ $\pm$ 3.5\\ {\scriptsize [-8.6, +5.3]}\\ {\scriptsize $p_{\mathrm{FDR}}$ = 0.784}\\ {\scriptsize $n=72$}\end{tabular} & \begin{tabular}[t]{@{}l@{}}$-5.1$ $\pm$ 2.4\\ {\scriptsize [-10.3, -1.2]}\\ {\scriptsize $p_{\mathrm{FDR}}$ = 0.192}\\ {\scriptsize $n=79$}\end{tabular} & \begin{tabular}[t]{@{}l@{}}$-4.3$ $\pm$ 4.2\\ {\scriptsize [-12.5, +4.2]}\\ {\scriptsize $p_{\mathrm{FDR}}$ = 0.413}\\ {\scriptsize $n=70$}\end{tabular} \\
Qwen3-VL-32B & \begin{tabular}[t]{@{}l@{}}$-9.4$ $\pm$ 8.6\\ {\scriptsize [-26.6, +7.9]}\\ {\scriptsize $p_{\mathrm{FDR}}$ = 0.369}\\ {\scriptsize $n=64$}\end{tabular} & \begin{tabular}[t]{@{}l@{}}$+16.2$ $\pm$ 5.3\\ {\scriptsize [+6.1, +26.9]}\\ {\scriptsize $p_{\mathrm{FDR}}$ = 0.006}\\ {\scriptsize $n=68$}\end{tabular} & \begin{tabular}[t]{@{}l@{}}$-1.6$ $\pm$ 4.2\\ {\scriptsize [-9.4, +6.7]}\\ {\scriptsize $p_{\mathrm{FDR}}$ = 0.784}\\ {\scriptsize $n=64$}\end{tabular} & \begin{tabular}[t]{@{}l@{}}$-4.4$ $\pm$ 2.4\\ {\scriptsize [-9.7, +0.0]}\\ {\scriptsize $p_{\mathrm{FDR}}$ = 0.204}\\ {\scriptsize $n=68$}\end{tabular} & \begin{tabular}[t]{@{}l@{}}$-6.3$ $\pm$ 4.8\\ {\scriptsize [-15.6, +3.3]}\\ {\scriptsize $p_{\mathrm{FDR}}$ = 0.312}\\ {\scriptsize $n=64$}\end{tabular} \\
Mistral-Small-4-119B & N/A & \begin{tabular}[t]{@{}l@{}}$+18.2$ $\pm$ 10.6\\ {\scriptsize [+0.0, +40.0]}\\ {\scriptsize $p_{\mathrm{FDR}}$ = 0.154}\\ {\scriptsize $n=11$}\end{tabular} & N/A & \begin{tabular}[t]{@{}l@{}}$-27.3$ $\pm$ 14.5\\ {\scriptsize [-55.7, +0.0]}\\ {\scriptsize $p_{\mathrm{FDR}}$ = 0.204}\\ {\scriptsize $n=11$}\end{tabular} & N/A \\
MedGemma-1.5-4B & \begin{tabular}[t]{@{}l@{}}$-4.2$ $\pm$ 6.7\\ {\scriptsize [-16.9, +8.8]}\\ {\scriptsize $p_{\mathrm{FDR}}$ = 0.563}\\ {\scriptsize $n=72$}\end{tabular} & \begin{tabular}[t]{@{}l@{}}$+20.3$ $\pm$ 5.7\\ {\scriptsize [+9.1, +31.1]}\\ {\scriptsize $p_{\mathrm{FDR}}$ = 0.004}\\ {\scriptsize $n=79$}\end{tabular} & \begin{tabular}[t]{@{}l@{}}$-1.4$ $\pm$ 3.0\\ {\scriptsize [-7.2, +4.3]}\\ {\scriptsize $p_{\mathrm{FDR}}$ = 0.784}\\ {\scriptsize $n=72$}\end{tabular} & \begin{tabular}[t]{@{}l@{}}$-5.1$ $\pm$ 2.4\\ {\scriptsize [-10.8, -1.3]}\\ {\scriptsize $p_{\mathrm{FDR}}$ = 0.192}\\ {\scriptsize $n=79$}\end{tabular} & \begin{tabular}[t]{@{}l@{}}$-4.2$ $\pm$ 4.7\\ {\scriptsize [-13.2, +5.3]}\\ {\scriptsize $p_{\mathrm{FDR}}$ = 0.426}\\ {\scriptsize $n=72$}\end{tabular} \\
LLaVA-Med-7B & \begin{tabular}[t]{@{}l@{}}$-38.4$ $\pm$ 5.2\\ {\scriptsize [-48.8, -28.4]}\\ {\scriptsize $p_{\mathrm{FDR}}$ = 0.002}\\ {\scriptsize $n=86$}\end{tabular} & \begin{tabular}[t]{@{}l@{}}$-13.3$ $\pm$ 3.6\\ {\scriptsize [-21.2, -6.7]}\\ {\scriptsize $p_{\mathrm{FDR}}$ = 0.004}\\ {\scriptsize $n=90$}\end{tabular} & \begin{tabular}[t]{@{}l@{}}$+4.7$ $\pm$ 2.3\\ {\scriptsize [+1.1, +9.5]}\\ {\scriptsize $p_{\mathrm{FDR}}$ = 0.233}\\ {\scriptsize $n=85$}\end{tabular} & \begin{tabular}[t]{@{}l@{}}$+0.0$ $\pm$ 0.0\\ {\scriptsize [+0.0, +0.0]}\\ {\scriptsize $p_{\mathrm{FDR}} > 0.999$}\\ {\scriptsize $n=89$}\end{tabular} & \begin{tabular}[t]{@{}l@{}}$+9.3$ $\pm$ 3.2\\ {\scriptsize [+3.5, +15.9]}\\ {\scriptsize $p_{\mathrm{FDR}}$ = 0.028}\\ {\scriptsize $n=86$}\end{tabular} \\
MedGemma-27B-text & \begin{tabular}[t]{@{}l@{}}$-34.2$ $\pm$ 5.6\\ {\scriptsize [-44.7, -23.9]}\\ {\scriptsize $p_{\mathrm{FDR}}$ = 0.002}\\ {\scriptsize $n=76$}\end{tabular} & \begin{tabular}[t]{@{}l@{}}$-7.4$ $\pm$ 3.0\\ {\scriptsize [-13.8, -2.4]}\\ {\scriptsize $p_{\mathrm{FDR}}$ = 0.019}\\ {\scriptsize $n=81$}\end{tabular} & \begin{tabular}[t]{@{}l@{}}$+3.9$ $\pm$ 2.2\\ {\scriptsize [+0.0, +8.3]}\\ {\scriptsize $p_{\mathrm{FDR}}$ = 0.233}\\ {\scriptsize $n=76$}\end{tabular} & \begin{tabular}[t]{@{}l@{}}$+0.0$ $\pm$ 0.0\\ {\scriptsize [+0.0, +0.0]}\\ {\scriptsize $p_{\mathrm{FDR}} > 0.999$}\\ {\scriptsize $n=81$}\end{tabular} & \begin{tabular}[t]{@{}l@{}}$+7.9$ $\pm$ 3.0\\ {\scriptsize [+2.6, +14.5]}\\ {\scriptsize $p_{\mathrm{FDR}}$ = 0.059}\\ {\scriptsize $n=76$}\end{tabular} \\
DeepSeek-R1-7B & \begin{tabular}[t]{@{}l@{}}$-60.9$ $\pm$ 10.9\\ {\scriptsize [-81.8, -39.1]}\\ {\scriptsize $p_{\mathrm{FDR}}$ = 0.002}\\ {\scriptsize $n=23$}\end{tabular} & \begin{tabular}[t]{@{}l@{}}$-28.6$ $\pm$ 9.8\\ {\scriptsize [-50.0, -12.5]}\\ {\scriptsize $p_{\mathrm{FDR}}$ = 0.006}\\ {\scriptsize $n=21$}\end{tabular} & \begin{tabular}[t]{@{}l@{}}$+8.7$ $\pm$ 6.0\\ {\scriptsize [+0.0, +22.7]}\\ {\scriptsize $p_{\mathrm{FDR}}$ = 0.389}\\ {\scriptsize $n=23$}\end{tabular} & \begin{tabular}[t]{@{}l@{}}$+0.0$ $\pm$ 0.0\\ {\scriptsize [+0.0, +0.0]}\\ {\scriptsize $p_{\mathrm{FDR}} > 0.999$}\\ {\scriptsize $n=21$}\end{tabular} & \begin{tabular}[t]{@{}l@{}}$+8.7$ $\pm$ 6.0\\ {\scriptsize [+0.0, +22.7]}\\ {\scriptsize $p_{\mathrm{FDR}}$ = 0.312}\\ {\scriptsize $n=23$}\end{tabular} \\
RAD-DINO & \begin{tabular}[t]{@{}l@{}}$-30.5$ $\pm$ 5.3\\ {\scriptsize [-40.7, -20.5]}\\ {\scriptsize $p_{\mathrm{FDR}}$ = 0.002}\\ {\scriptsize $n=82$}\end{tabular} & \begin{tabular}[t]{@{}l@{}}$-6.9$ $\pm$ 3.8\\ {\scriptsize [-14.5, +0.0]}\\ {\scriptsize $p_{\mathrm{FDR}}$ = 0.079}\\ {\scriptsize $n=87$}\end{tabular} & \begin{tabular}[t]{@{}l@{}}$+3.7$ $\pm$ 2.1\\ {\scriptsize [+0.0, +7.8]}\\ {\scriptsize $p_{\mathrm{FDR}}$ = 0.233}\\ {\scriptsize $n=82$}\end{tabular} & \begin{tabular}[t]{@{}l@{}}$+0.0$ $\pm$ 0.0\\ {\scriptsize [+0.0, +0.0]}\\ {\scriptsize $p_{\mathrm{FDR}} > 0.999$}\\ {\scriptsize $n=87$}\end{tabular} & \begin{tabular}[t]{@{}l@{}}$+6.1$ $\pm$ 3.1\\ {\scriptsize [+0.0, +12.8]}\\ {\scriptsize $p_{\mathrm{FDR}}$ = 0.145}\\ {\scriptsize $n=82$}\end{tabular} \\
\bottomrule
\end{tabular}
\end{table}

\begin{table}[t]
\centering
\caption{The difficulty-stratified set ($n$ = 80 MS-CXR finding-present cases, 240 displays) for the three readers, their majority answer, and all systems. Every correct answer is Yes. So accuracy equals sensitivity. Accuracy is measured on the original displays, and CGR and IS (corner placement) are measured on the correct answers. Each is the mean over its cases $\pm$ the SD of 1{,}000 patient-cluster bootstrap resamples, with the percentile 95\% interval and the case count. The last column is the paired difference in accuracy, system minus majority answer, with the statistics of Supplementary Table~\ref{stab:reader_balanced}. CGR, causal grounding rate; FDR, false discovery rate; IS, irrelevant-mask stability; SD, standard deviation. N/A, not applicable.}
\label{stab:reader_submitted}
\setlength{\tabcolsep}{2pt}
\renewcommand{\arraystretch}{0.9}
\scriptsize
\begin{tabular}{@{}p{0.17\textwidth}p{0.17\textwidth}p{0.17\textwidth}p{0.17\textwidth}p{0.25\textwidth}@{}}
\toprule
Reader or system & Accuracy & CGR & IS (corner) & Accuracy minus majority \\
\midrule
\multicolumn{5}{@{}l}{\textbf{Readers}} \\
S.Z. & \begin{tabular}[t]{@{}l@{}}81.3 $\pm$ 4.7\\ {\scriptsize [71.2, 89.7]}\\ {\scriptsize $n=80$}\end{tabular} & \begin{tabular}[t]{@{}l@{}}23.1 $\pm$ 5.3\\ {\scriptsize [12.7, 33.8]}\\ {\scriptsize $n=65$}\end{tabular} & \begin{tabular}[t]{@{}l@{}}93.8 $\pm$ 2.7\\ {\scriptsize [88.2, 98.5]}\\ {\scriptsize $n=65$}\end{tabular} & N/A \\
L.A. & \begin{tabular}[t]{@{}l@{}}58.8 $\pm$ 6.2\\ {\scriptsize [46.7, 70.1]}\\ {\scriptsize $n=80$}\end{tabular} & \begin{tabular}[t]{@{}l@{}}0.0 $\pm$ 0.0\\ {\scriptsize [0.0, 0.0]}\\ {\scriptsize $n=47$}\end{tabular} & \begin{tabular}[t]{@{}l@{}}100.0 $\pm$ 0.0\\ {\scriptsize [100.0, 100.0]}\\ {\scriptsize $n=47$}\end{tabular} & N/A \\
T.T.N. & \begin{tabular}[t]{@{}l@{}}73.8 $\pm$ 5.6\\ {\scriptsize [62.6, 84.2]}\\ {\scriptsize $n=80$}\end{tabular} & \begin{tabular}[t]{@{}l@{}}33.9 $\pm$ 6.2\\ {\scriptsize [22.4, 45.9]}\\ {\scriptsize $n=59$}\end{tabular} & \begin{tabular}[t]{@{}l@{}}91.5 $\pm$ 4.0\\ {\scriptsize [82.8, 98.3]}\\ {\scriptsize $n=59$}\end{tabular} & N/A \\
Majority answer & \begin{tabular}[t]{@{}l@{}}76.3 $\pm$ 5.3\\ {\scriptsize [65.4, 85.7]}\\ {\scriptsize $n=80$}\end{tabular} & \begin{tabular}[t]{@{}l@{}}27.9 $\pm$ 5.9\\ {\scriptsize [16.7, 39.0]}\\ {\scriptsize $n=61$}\end{tabular} & \begin{tabular}[t]{@{}l@{}}93.4 $\pm$ 3.6\\ {\scriptsize [86.2, 100.0]}\\ {\scriptsize $n=61$}\end{tabular} & N/A \\
\midrule
\multicolumn{5}{@{}l}{\textbf{Systems}} \\
Gemma-4-26B & \begin{tabular}[t]{@{}l@{}}69.0 $\pm$ 6.0\\ {\scriptsize [57.1, 80.3]}\\ {\scriptsize $n=71$}\end{tabular} & \begin{tabular}[t]{@{}l@{}}29.8 $\pm$ 7.1\\ {\scriptsize [16.0, 43.5]}\\ {\scriptsize $n=47$}\end{tabular} & \begin{tabular}[t]{@{}l@{}}95.9 $\pm$ 2.9\\ {\scriptsize [89.4, 100.0]}\\ {\scriptsize $n=49$}\end{tabular} & \begin{tabular}[t]{@{}l@{}}$-11.3$ $\pm$ 5.7\\ {\scriptsize [-22.2, -1.4]}\\ {\scriptsize $p_{\mathrm{FDR}}$ = 0.061, $n=71$}\\ {\scriptsize 90\% [-20.3, -1.5]}\end{tabular} \\
Qwen3-VL-32B & \begin{tabular}[t]{@{}l@{}}45.0 $\pm$ 6.3\\ {\scriptsize [33.3, 58.4]}\\ {\scriptsize $n=80$}\end{tabular} & \begin{tabular}[t]{@{}l@{}}19.4 $\pm$ 6.6\\ {\scriptsize [8.1, 32.5]}\\ {\scriptsize $n=36$}\end{tabular} & \begin{tabular}[t]{@{}l@{}}83.3 $\pm$ 6.3\\ {\scriptsize [70.3, 94.6]}\\ {\scriptsize $n=36$}\end{tabular} & \begin{tabular}[t]{@{}l@{}}$-31.3$ $\pm$ 6.3\\ {\scriptsize [-42.9, -19.2]}\\ {\scriptsize $p_{\mathrm{FDR}}$ = 0.002, $n=80$}\\ {\scriptsize 90\% [-41.5, -20.8]}\end{tabular} \\
Mistral-Small-4-119B & \begin{tabular}[t]{@{}l@{}}5.0 $\pm$ 2.5\\ {\scriptsize [1.2, 10.3]}\\ {\scriptsize $n=80$}\end{tabular} & \begin{tabular}[t]{@{}l@{}}25.0 $\pm$ 21.7\\ {\scriptsize [0.0, 75.0]}\\ {\scriptsize $n=4$}\end{tabular} & \begin{tabular}[t]{@{}l@{}}50.0 $\pm$ 25.0\\ {\scriptsize [0.0, 100.0]}\\ {\scriptsize $n=4$}\end{tabular} & \begin{tabular}[t]{@{}l@{}}$-71.3$ $\pm$ 5.5\\ {\scriptsize [-81.0, -59.3]}\\ {\scriptsize $p_{\mathrm{FDR}}$ = 0.002, $n=80$}\\ {\scriptsize 90\% [-79.5, -61.7]}\end{tabular} \\
MedGemma-1.5-4B & \begin{tabular}[t]{@{}l@{}}66.3 $\pm$ 5.9\\ {\scriptsize [54.9, 77.9]}\\ {\scriptsize $n=80$}\end{tabular} & \begin{tabular}[t]{@{}l@{}}43.4 $\pm$ 7.1\\ {\scriptsize [30.2, 57.7]}\\ {\scriptsize $n=53$}\end{tabular} & \begin{tabular}[t]{@{}l@{}}92.5 $\pm$ 3.6\\ {\scriptsize [84.9, 98.2]}\\ {\scriptsize $n=53$}\end{tabular} & \begin{tabular}[t]{@{}l@{}}$-10.0$ $\pm$ 6.7\\ {\scriptsize [-22.5, +3.8]}\\ {\scriptsize $p_{\mathrm{FDR}}$ = 0.193, $n=80$}\\ {\scriptsize 90\% [-20.7, +1.3]}\end{tabular} \\
LLaVA-Med-7B & \begin{tabular}[t]{@{}l@{}}100.0 $\pm$ 0.0\\ {\scriptsize [100.0, 100.0]}\\ {\scriptsize $n=79$}\end{tabular} & \begin{tabular}[t]{@{}l@{}}0.0 $\pm$ 0.0\\ {\scriptsize [0.0, 0.0]}\\ {\scriptsize $n=79$}\end{tabular} & \begin{tabular}[t]{@{}l@{}}100.0 $\pm$ 0.0\\ {\scriptsize [100.0, 100.0]}\\ {\scriptsize $n=79$}\end{tabular} & \begin{tabular}[t]{@{}l@{}}$+24.1$ $\pm$ 5.3\\ {\scriptsize [+14.3, +34.2]}\\ {\scriptsize $p_{\mathrm{FDR}}$ = 0.002, $n=79$}\\ {\scriptsize 90\% [+15.8, +33.3]}\end{tabular} \\
MedGemma-27B-text & \begin{tabular}[t]{@{}l@{}}83.8 $\pm$ 4.7\\ {\scriptsize [74.0, 92.5]}\\ {\scriptsize $n=80$}\end{tabular} & \begin{tabular}[t]{@{}l@{}}0.0 $\pm$ 0.0\\ {\scriptsize [0.0, 0.0]}\\ {\scriptsize $n=67$}\end{tabular} & \begin{tabular}[t]{@{}l@{}}100.0 $\pm$ 0.0\\ {\scriptsize [100.0, 100.0]}\\ {\scriptsize $n=67$}\end{tabular} & \begin{tabular}[t]{@{}l@{}}$+7.5$ $\pm$ 7.1\\ {\scriptsize [-5.9, +21.1]}\\ {\scriptsize $p_{\mathrm{FDR}}$ = 0.316, $n=80$}\\ {\scriptsize 90\% [-3.6, +18.8]}\end{tabular} \\
DeepSeek-R1-7B & \begin{tabular}[t]{@{}l@{}}28.2 $\pm$ 5.9\\ {\scriptsize [17.1, 40.3]}\\ {\scriptsize $n=71$}\end{tabular} & \begin{tabular}[t]{@{}l@{}}0.0 $\pm$ 0.0\\ {\scriptsize [0.0, 0.0]}\\ {\scriptsize $n=20$}\end{tabular} & \begin{tabular}[t]{@{}l@{}}100.0 $\pm$ 0.0\\ {\scriptsize [100.0, 100.0]}\\ {\scriptsize $n=20$}\end{tabular} & \begin{tabular}[t]{@{}l@{}}$-49.3$ $\pm$ 7.5\\ {\scriptsize [-63.4, -33.3]}\\ {\scriptsize $p_{\mathrm{FDR}}$ = 0.002, $n=71$}\\ {\scriptsize 90\% [-61.1, -36.6]}\end{tabular} \\
RAD-DINO & \begin{tabular}[t]{@{}l@{}}93.8 $\pm$ 2.7\\ {\scriptsize [87.9, 98.7]}\\ {\scriptsize $n=80$}\end{tabular} & \begin{tabular}[t]{@{}l@{}}12.0 $\pm$ 3.9\\ {\scriptsize [4.9, 20.3]}\\ {\scriptsize $n=75$}\end{tabular} & \begin{tabular}[t]{@{}l@{}}98.7 $\pm$ 1.3\\ {\scriptsize [95.8, 100.0]}\\ {\scriptsize $n=75$}\end{tabular} & \begin{tabular}[t]{@{}l@{}}$+17.5$ $\pm$ 6.1\\ {\scriptsize [+6.2, +29.6]}\\ {\scriptsize $p_{\mathrm{FDR}}$ = 0.008, $n=80$}\\ {\scriptsize 90\% [+8.4, +27.8]}\end{tabular} \\
\bottomrule
\end{tabular}
\end{table}

\begin{table}[t]
\centering
\caption{Validity of the radiologist-marked MS-CXR boxes, rated by S.Z. and T.T.N. on a sample of 120 boxes, 15 per finding, and by T.T.N. on the 100 boxes of the balanced set. A box is rated accurate when it covers the primary location of the finding. Each cell is the count of boxes rated accurate out of the boxes rated and their share in percent. The interval of one finding is the Wilson 95\% interval. For all findings, the share is given $\pm$ the SD of 1{,}000 patient-cluster bootstrap resamples, with the percentile 95\% interval. The column headed Both raters is the number of boxes rated accurate by both raters. SD, standard deviation.}
\label{stab:box_validity}
\setlength{\tabcolsep}{2pt}
\renewcommand{\arraystretch}{1.05}
\scriptsize
\begin{tabular}{@{}p{0.16\textwidth}p{0.195\textwidth}p{0.195\textwidth}p{0.195\textwidth}p{0.195\textwidth}@{}}
\toprule
Finding & S.Z., 120 boxes & T.T.N., 120 boxes & Both raters, 120 boxes & T.T.N., 100 boxes \\
\midrule
Atelectasis & \begin{tabular}[t]{@{}l@{}}6 of 15, 40.0\\ {\scriptsize Wilson [19.8, 64.3]}\end{tabular} & \begin{tabular}[t]{@{}l@{}}8 of 15, 53.3\\ {\scriptsize Wilson [30.1, 75.2]}\end{tabular} & \begin{tabular}[t]{@{}l@{}}5 of 15, 33.3\\ {\scriptsize Wilson [15.2, 58.3]}\end{tabular} & \begin{tabular}[t]{@{}l@{}}6 of 13, 46.2\\ {\scriptsize Wilson [23.2, 70.9]}\end{tabular} \\
Cardiomegaly & \begin{tabular}[t]{@{}l@{}}15 of 15, 100.0\\ {\scriptsize Wilson [79.6, 100.0]}\end{tabular} & \begin{tabular}[t]{@{}l@{}}15 of 15, 100.0\\ {\scriptsize Wilson [79.6, 100.0]}\end{tabular} & \begin{tabular}[t]{@{}l@{}}15 of 15, 100.0\\ {\scriptsize Wilson [79.6, 100.0]}\end{tabular} & \begin{tabular}[t]{@{}l@{}}12 of 12, 100.0\\ {\scriptsize Wilson [75.8, 100.0]}\end{tabular} \\
Consolidation & \begin{tabular}[t]{@{}l@{}}4 of 15, 26.7\\ {\scriptsize Wilson [10.9, 52.0]}\end{tabular} & \begin{tabular}[t]{@{}l@{}}4 of 15, 26.7\\ {\scriptsize Wilson [10.9, 52.0]}\end{tabular} & \begin{tabular}[t]{@{}l@{}}3 of 15, 20.0\\ {\scriptsize Wilson [7.0, 45.2]}\end{tabular} & \begin{tabular}[t]{@{}l@{}}6 of 12, 50.0\\ {\scriptsize Wilson [25.4, 74.6]}\end{tabular} \\
Edema & \begin{tabular}[t]{@{}l@{}}1 of 15, 6.7\\ {\scriptsize Wilson [1.2, 29.8]}\end{tabular} & \begin{tabular}[t]{@{}l@{}}0 of 15, 0.0\\ {\scriptsize Wilson [0.0, 20.4]}\end{tabular} & \begin{tabular}[t]{@{}l@{}}0 of 15, 0.0\\ {\scriptsize Wilson [0.0, 20.4]}\end{tabular} & \begin{tabular}[t]{@{}l@{}}1 of 13, 7.7\\ {\scriptsize Wilson [1.4, 33.3]}\end{tabular} \\
Lung opacity & \begin{tabular}[t]{@{}l@{}}3 of 15, 20.0\\ {\scriptsize Wilson [7.0, 45.2]}\end{tabular} & \begin{tabular}[t]{@{}l@{}}1 of 15, 6.7\\ {\scriptsize Wilson [1.2, 29.8]}\end{tabular} & \begin{tabular}[t]{@{}l@{}}1 of 15, 6.7\\ {\scriptsize Wilson [1.2, 29.8]}\end{tabular} & \begin{tabular}[t]{@{}l@{}}7 of 13, 53.8\\ {\scriptsize Wilson [29.1, 76.8]}\end{tabular} \\
Pleural effusion & \begin{tabular}[t]{@{}l@{}}7 of 15, 46.7\\ {\scriptsize Wilson [24.8, 69.9]}\end{tabular} & \begin{tabular}[t]{@{}l@{}}6 of 15, 40.0\\ {\scriptsize Wilson [19.8, 64.3]}\end{tabular} & \begin{tabular}[t]{@{}l@{}}4 of 15, 26.7\\ {\scriptsize Wilson [10.9, 52.0]}\end{tabular} & \begin{tabular}[t]{@{}l@{}}3 of 12, 25.0\\ {\scriptsize Wilson [8.9, 53.2]}\end{tabular} \\
Pneumonia & \begin{tabular}[t]{@{}l@{}}9 of 15, 60.0\\ {\scriptsize Wilson [35.7, 80.2]}\end{tabular} & \begin{tabular}[t]{@{}l@{}}6 of 15, 40.0\\ {\scriptsize Wilson [19.8, 64.3]}\end{tabular} & \begin{tabular}[t]{@{}l@{}}5 of 15, 33.3\\ {\scriptsize Wilson [15.2, 58.3]}\end{tabular} & \begin{tabular}[t]{@{}l@{}}6 of 13, 46.2\\ {\scriptsize Wilson [23.2, 70.9]}\end{tabular} \\
Pneumothorax & \begin{tabular}[t]{@{}l@{}}13 of 15, 86.7\\ {\scriptsize Wilson [62.1, 96.3]}\end{tabular} & \begin{tabular}[t]{@{}l@{}}7 of 15, 46.7\\ {\scriptsize Wilson [24.8, 69.9]}\end{tabular} & \begin{tabular}[t]{@{}l@{}}7 of 15, 46.7\\ {\scriptsize Wilson [24.8, 69.9]}\end{tabular} & \begin{tabular}[t]{@{}l@{}}9 of 12, 75.0\\ {\scriptsize Wilson [46.8, 91.1]}\end{tabular} \\
\midrule
\textbf{All findings} & \begin{tabular}[t]{@{}l@{}}58 of 120, 48.3\\ {\scriptsize $\pm$ 4.5 [39.7, 57.4]}\end{tabular} & \begin{tabular}[t]{@{}l@{}}47 of 120, 39.2\\ {\scriptsize $\pm$ 4.7 [29.8, 48.7]}\end{tabular} & \begin{tabular}[t]{@{}l@{}}40 of 120, 33.3\\ {\scriptsize $\pm$ 4.5 [24.8, 42.0]}\end{tabular} & \begin{tabular}[t]{@{}l@{}}50 of 100, 50.0\\ {\scriptsize $\pm$ 4.8 [41.3, 59.6]}\end{tabular} \\
\bottomrule
\end{tabular}
\end{table}

\begin{table}[t]
\centering
\caption{Failure taxonomy of 22 finding-presence cases of the MIMIC-CXR block, drawn from the most confident quarter of the wrong answers of Gemma-4-26B (14 cases) and of the vision-only reference (eight cases) and classified by L.A. and by T.T.N. The vision-only reference analyzed here answers one of its eight cases correctly. Each category cell is the count of cases placed in the category by L.A. and by T.T.N., written L.A. / T.T.N. The last column is the mean 1-to-5 rating of how confidently a typical radiologist would answer the case, in the same order.}
\label{stab:taxonomy}
\setlength{\tabcolsep}{2pt}
\renewcommand{\arraystretch}{1.05}
\scriptsize
\begin{tabular}{@{}p{0.135\textwidth}p{0.055\textwidth}p{0.095\textwidth}p{0.11\textwidth}p{0.095\textwidth}p{0.09\textwidth}p{0.06\textwidth}p{0.115\textwidth}@{}}
\toprule
System & Cases & Ambiguous case & Plausible image confounder & Clear model failure & Poor image quality & Other & Mean confidence \\
\midrule
Gemma-4-26B & 14 & 8 / 8 & 4 / 1 & 2 / 1 & 0 / 2 & 0 / 2 & 2.93 / 2.14 \\
RAD-DINO & 8 & 2 / 1 & 4 / 1 & 1 / 5 & 1 / 0 & 0 / 1 & 3.25 / 3.25 \\
\midrule
\textbf{All} & 22 & 10 / 9 & 8 / 2 & 3 / 6 & 1 / 2 & 0 / 3 & 3.05 / 2.55 \\
\bottomrule
\end{tabular}
\end{table}

\begin{table}[t]
\centering
\caption{Composition of the CheXpert question set by finding, label, view, and sex ($n$ = 1{,}380 cases from 1{,}285 patients). Yes and No count the finding-present and finding-absent cases. The 100 normal studies are asked whether any acute abnormality is present, and their correct answer is No. PA and AP count posteroanterior and anteroposterior acquisitions, and F and M count female and male patients as recorded in the dataset's sex field.}
\label{stab:composition_chexpert}
\setlength{\tabcolsep}{4pt}
\renewcommand{\arraystretch}{1.05}
\footnotesize
\begin{tabular}{@{}p{0.34\textwidth}rrrrrr@{}}
\toprule
Finding & Yes & No & PA & AP & F & M \\
\midrule
Atelectasis & 50 & 50 & 35 & 65 & 41 & 59 \\
Cardiomegaly & 50 & 50 & 44 & 56 & 35 & 65 \\
Consolidation & 50 & 50 & 46 & 54 & 47 & 53 \\
Edema & 50 & 50 & 27 & 73 & 54 & 46 \\
Enlarged cardiomediastinum & 50 & 50 & 50 & 50 & 39 & 61 \\
Fracture & 50 & 50 & 38 & 62 & 40 & 60 \\
Lung lesion & 50 & 50 & 76 & 24 & 38 & 62 \\
Lung opacity & 50 & 50 & 42 & 58 & 39 & 61 \\
Pleural effusion & 50 & 50 & 39 & 61 & 52 & 48 \\
Pleural other & 50 & 30 & 57 & 23 & 35 & 45 \\
Pneumonia & 50 & 50 & 62 & 38 & 42 & 58 \\
Pneumothorax & 50 & 50 & 26 & 74 & 36 & 64 \\
Support devices & 50 & 50 & 25 & 75 & 44 & 56 \\
No finding & 0 & 100 & 52 & 48 & 40 & 60 \\
\midrule
\textbf{Question set} & 650 & 730 & 619 & 761 & 582 & 798 \\
\bottomrule
\end{tabular}
\end{table}

\clearpage

\begin{algorithm}[t]
\caption{Paired patient-cluster bootstrap with the shift-and-reflect $p$-value and the equivalence interval}
\label{alg:paired_bootstrap}
\begin{algorithmic}[1]
\Require Paired outcomes $\{(o^A_j, o^B_j)\}_{j=1}^{n}$ on the shared cases, the patient $c_j$ of every case, the resample count $B$, the seed, and the equivalence margin $m$
\Ensure The point difference $\hat{\delta}$, its 95\% and 90\% intervals, the two-sided $p$-value, and whether equivalence is established
\State $\hat{\delta} \gets \overline{o^A} - \overline{o^B}$
\State Initialize the random number generator with the seed; let $P$ be the set of distinct patients
\For{$b = 1, \dots, B$}
    \State Draw $|P|$ patients uniformly with replacement from $P$, and take every case of every drawn patient as often as the patient was drawn
    \State $\hat{\delta}_b \gets$ the mean of $o^A_j - o^B_j$ over the drawn cases
\EndFor
\State $L_{95}, U_{95} \gets$ the 2.5th and 97.5th percentiles of $\{\hat{\delta}_b\}$; $L_{90}, U_{90} \gets$ the 5th and 95th percentiles
\State $\tilde{\delta}_b \gets \hat{\delta}_b - \hat{\delta}$ for $b = 1, \dots, B$
\State $p \gets \max\!\left(\frac{1}{B},\; \frac{1}{B} \sum_{b=1}^B \mathbf{1}\!\{|\tilde{\delta}_b| \geq |\hat{\delta}|\}\right)$
\State equivalent $\gets \mathbf{1}\{L_{90} > -m \text{ and } U_{90} < m\}$
\State \Return $\hat{\delta}, [L_{95}, U_{95}], [L_{90}, U_{90}], p$, equivalent
\end{algorithmic}
\end{algorithm}

\end{document}